\documentclass{article} 
\usepackage{iclr2027_conference,times}

\usepackage{amsmath,amsfonts,bm}

\def\eqref#1{equation~\ref{#1}}
\def\Eqref#1{Equation~\ref{#1}}

\def\1{\bm{1}}

\DeclareMathAlphabet{\mathsfit}{\encodingdefault}{\sfdefault}{m}{sl}
\SetMathAlphabet{\mathsfit}{bold}{\encodingdefault}{\sfdefault}{bx}{n}

\usepackage{hyperref}
\hypersetup{hidelinks}
\usepackage{url}

\usepackage{amsmath}
\usepackage{amssymb}
\usepackage{amsfonts}
\usepackage{booktabs}
\usepackage{graphicx}
\usepackage{multirow}
\usepackage{adjustbox}
\usepackage{subcaption}
\usepackage{xspace}
\usepackage{enumitem}
\usepackage{algorithm}
\usepackage{algorithmic}
\usepackage{fancyvrb}
\usepackage{wrapfig}

\usepackage{pgf}
\usepackage[table]{xcolor}

\definecolor{abovegreen}{RGB}{118,176,83}
\definecolor{belowblue}{RGB}{68,141,192}
\definecolor{heatneutral}{gray}{0.94}

\newcommand{\setbase}[1]{\gdef\base{#1}}
\newcommand{\base}{}

\newcommand{\gcell}[1]{%
  \pgfmathsetmacro{\db}{(#1)-(\base)}%
  \pgfmathsetmacro{\ad}{abs(\db)}%
  \ifdim \db pt < 0pt
    \ifdim \ad pt > 2.70pt {\cellcolor{belowblue!70}#1}%
    \else\ifdim \ad pt > 1.70pt {\cellcolor{belowblue!55}#1}%
    \else\ifdim \ad pt > 0.70pt {\cellcolor{belowblue!40}#1}%
    \else {\cellcolor{belowblue!25}#1}%
    \fi\fi\fi
  \else
    \ifdim \ad pt > 0.50pt {\cellcolor{abovegreen!70}#1}%
    \else\ifdim \ad pt > 0.20pt {\cellcolor{abovegreen!55}#1}%
    \else\ifdim \ad pt > 0.10pt {\cellcolor{abovegreen!40}#1}%
    \else {\cellcolor{abovegreen!25}#1}%
    \fi\fi\fi
  \fi
}

\newcommand{\pcell}[1]{%
  \pgfmathsetmacro{\db}{(\base)-(#1)}%
  \pgfmathsetmacro{\ad}{abs(\db)}%
  \ifdim \db pt < 0pt
    \ifdim \ad pt > 0.50pt {\cellcolor{belowblue!70}#1}%
    \else\ifdim \ad pt > 0.20pt {\cellcolor{belowblue!55}#1}%
    \else\ifdim \ad pt > 0.10pt {\cellcolor{belowblue!40}#1}%
    \else {\cellcolor{belowblue!25}#1}%
    \fi\fi\fi
  \else
    \ifdim \ad pt > 0.50pt {\cellcolor{abovegreen!70}#1}%
    \else\ifdim \ad pt > 0.20pt {\cellcolor{abovegreen!55}#1}%
    \else\ifdim \ad pt > 0.10pt {\cellcolor{abovegreen!40}#1}%
    \else {\cellcolor{abovegreen!25}#1}%
    \fi\fi\fi
  \fi
}

\ExplSyntaxOn

\cs_new_protected:Npn \aq_system_green:n #1
{
  \fp_compare:nNnTF { abs(\l_tmpa_fp) } > { 100 }
    { \cellcolor{abovegreen!80}#1 }
    {
      \fp_compare:nNnTF { abs(\l_tmpa_fp) } > { 60 }
        { \cellcolor{abovegreen!70}#1 }
        {
          \fp_compare:nNnTF { abs(\l_tmpa_fp) } > { 45 }
            { \cellcolor{abovegreen!60}#1 }
            {
              \fp_compare:nNnTF { abs(\l_tmpa_fp) } > { 30 }
                { \cellcolor{abovegreen!55}#1 }
                {
                  \fp_compare:nNnTF { abs(\l_tmpa_fp) } > { 22 }
                    { \cellcolor{abovegreen!50}#1 }
                    {
                      \fp_compare:nNnTF { abs(\l_tmpa_fp) } > { 15 }
                        { \cellcolor{abovegreen!40}#1 }
                        {
                          \fp_compare:nNnTF { abs(\l_tmpa_fp) } > { 8 }
                            { \cellcolor{abovegreen!30}#1 }
                            { \cellcolor{abovegreen!20}#1 }
                        }
                    }
                }
            }
        }
    }
}

\cs_new_protected:Npn \aq_system_blue:n #1
{
  \fp_compare:nNnTF { abs(\l_tmpa_fp) } > { 100 }
    { \cellcolor{belowblue!80}#1 }
    {
      \fp_compare:nNnTF { abs(\l_tmpa_fp) } > { 60 }
        { \cellcolor{belowblue!70}#1 }
        {
          \fp_compare:nNnTF { abs(\l_tmpa_fp) } > { 45 }
            { \cellcolor{belowblue!60}#1 }
            {
              \fp_compare:nNnTF { abs(\l_tmpa_fp) } > { 30 }
                { \cellcolor{belowblue!55}#1 }
                {
                  \fp_compare:nNnTF { abs(\l_tmpa_fp) } > { 22 }
                    { \cellcolor{belowblue!50}#1 }
                    {
                      \fp_compare:nNnTF { abs(\l_tmpa_fp) } > { 15 }
                        { \cellcolor{belowblue!40}#1 }
                        {
                          \fp_compare:nNnTF { abs(\l_tmpa_fp) } > { 8 }
                            { \cellcolor{belowblue!30}#1 }
                            { \cellcolor{belowblue!20}#1 }
                        }
                    }
                }
            }
        }
    }
}

\NewDocumentCommand{\sgcell}{m}
{
  \fp_set:Nn \l_tmpa_fp
    { 100 * ((#1) - (\base)) / abs(\base) }

  \fp_compare:nNnTF { \l_tmpa_fp } < { 0 }
    { \aq_system_blue:n {#1} }
    { \aq_system_green:n {#1} }
}

\NewDocumentCommand{\spcell}{m}
{
  \fp_set:Nn \l_tmpa_fp
    { 100 * ((\base) - (#1)) / abs(\base) }

  \fp_compare:nNnTF { \l_tmpa_fp } < { 0 }
    { \aq_system_blue:n {#1} }
    { \aq_system_green:n {#1} }
}

\ExplSyntaxOff

\usepackage{tikz}
\newcommand{\ds}[1]{%
  \tikz[baseline=(X.base)]
  \node[draw,circle,inner sep=0.6pt,minimum size=1em]
  (X) {\scriptsize #1};%
}

\newcommand{\modelname}{\textit{AlignQuant}\xspace}

\title{\modelname: Tile-Aligned Mixed-Precision\\Quantization for Efficient LLM Generation}

\author{
Hanzhi Zhang\textsuperscript{1},
Qiao Zhang\textsuperscript{2},
Qinglei Cao\textsuperscript{2},
Heng Fan\textsuperscript{1},
Yan Huang\textsuperscript{1},
Kewei Sha\textsuperscript{3},
Yunhe Feng\textsuperscript{1} \\
\textsuperscript{1}LLaVi Lab, Computer Science \& Engineering, University of North Texas \\
\textsuperscript{2}Computer Science, Saint Louis University;
\textsuperscript{3}Data Science, University of North Texas \\
\texttt{\{hanzhi.zhang,heng.fan,yan.huang,kewei.sha,yunhe.feng\}@unt.edu} \\
\texttt{\{qiao.zhang,qinglei.cao\}@slu.edu}
}

\iclrfinalcopy

\begin{document}

\maketitle
\lhead{}
\renewcommand{\headrulewidth}{0pt}

\vspace{-18pt}
\begin{abstract}
Fine-grained mixed-precision quantization promises efficient large language model inference, but local precision choices can conflict with regular GPU storage and computation units.
This precision-boundary mismatch limits the translation of compression into practical acceleration.
We introduce \modelname, a post-training quantization method that uses GPU-compatible two-dimensional weight tiles as the common unit of precision allocation, compact storage, and execution.
This shared partition lets precision follow sensitivity within output channels.
Joint prefill/decode calibration scores precision reductions using projection-output perturbations weighted by language-model loss gradients under quantized activations.
Phase-normalized scores prioritize higher precision for tiles important to either phase under a model-wide weight-storage budget.
Each tile stores one selected representation, while phase-specialized kernels reuse the packed model and expand lower-bit weights for INT8 computation with 8-bit activations.
Across four LLMs spanning 3B to 14B parameters, \modelname achieves up to $2.50\times$ generation speedup over BF16 while preserving model quality.
Evaluations further cover three GPUs and contexts up to 64K tokens.
These results show that local precision flexibility and regular GPU execution can coexist through a shared tile unit.
The implementation is available at \url{https://github.com/HanzhiZhang-Ulrica/AlignQuant}.
\end{abstract}

\vspace{-10pt}
\section{Introduction}
\label{sec:introduction}
\vspace{-5pt}

Large language model (LLM) inference is increasingly constrained by parameter storage, memory traffic, and computation.
Post-training quantization (PTQ) reduces these costs through lower-precision representations, while the resulting quantization error makes accuracy preservation a central challenge
\citep{dettmers2022gpt3,xiao2023smoothquant,frantar2022gptq,lin2024awq}.
Uniform-bit quantization addresses this challenge by improving the quantized representation at a fixed weight precision
\citep{frantar2022gptq,lin2024awq}.
Mixed-precision quantization additionally exploits differences in sensitivity by assigning higher precision to sensitive weight regions under a weight-storage budget
\citep{chen2024channel,huang2024slim,zheng2026mixllm,li2026scalebits,nie2026sfmp,hooper2025fgmp}.
However, when fine-grained precision boundaries cross regular storage or GPU computation units, the resulting \textbf{precision-boundary mismatch} makes efficient deployment depend on how mixed-bit weights are packed, loaded, and converted for computation
\citep{zhao2024atom,lin2025qserve}.

\begin{figure*}[t]
    \centering
    \includegraphics[width=\textwidth]{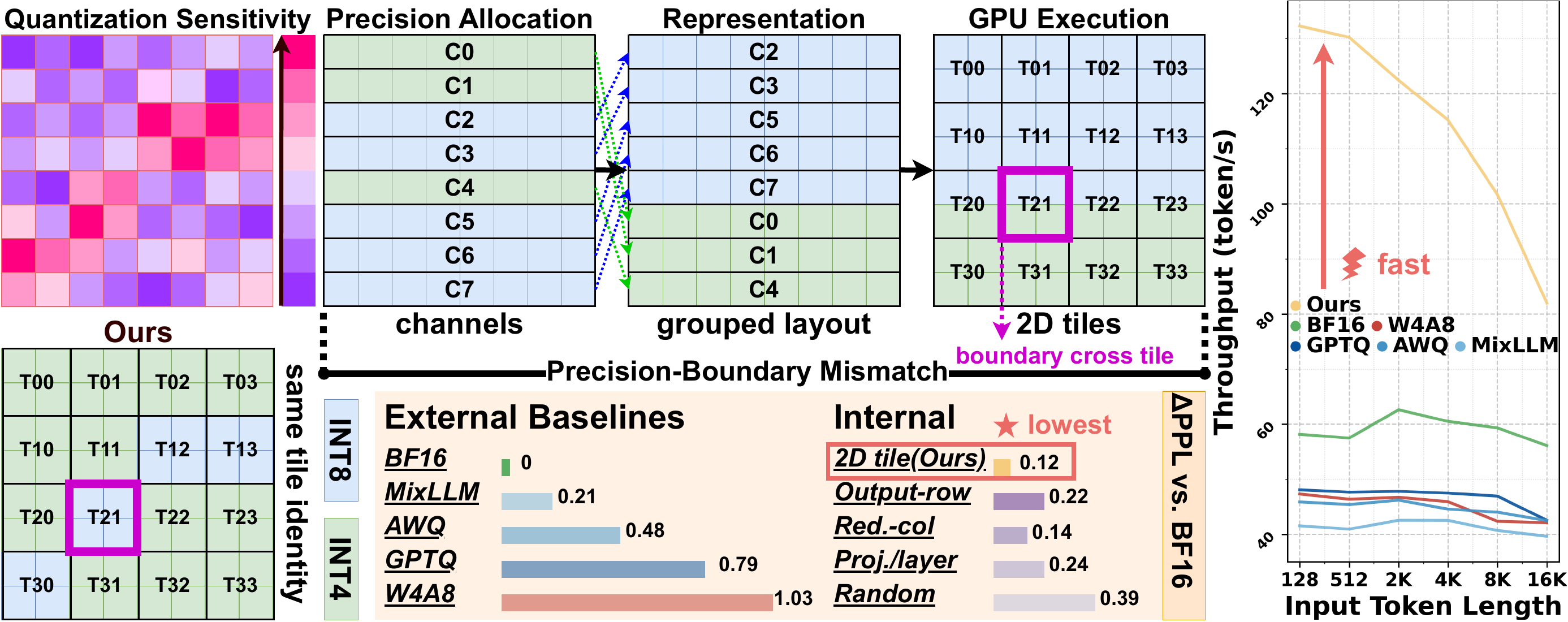}
    \vspace{-20pt}
    \caption{
    Comparison of precision allocation, representation, and GPU execution.
    Colors denote 8-bit/4-bit weight precision, and purple outlines track logical tiles.
    The schematic shows how channel-wise precision grouping can create a precision-boundary mismatch with regular two-dimensional GPU tiling, while \modelname uses a common weight-tile partition across allocation, storage, and execution.
    The quality panel reports WikiText-2 PPL increase over BF16, and the right panel reports generation throughput across prompt lengths on Llama-3.2-3B.
    }
    \label{fig:intro}
    \vspace{-20pt}
\end{figure*}

Figure~\ref{fig:intro} illustrates how precision allocation, weight representation, and GPU execution jointly shape model quality and inference efficiency.
Existing methods manage this tension through different precision-sharing and execution structures.
Channel- and output-feature-level methods maintain regular partitions by sharing one bit-width across each channel or output feature, coupling weights with different sensitivities to the same precision choice
\citep{chen2024channel,zheng2026mixllm}.
Finer-grained approaches expose more local choices through quantization groups and hardware-aligned blocks, together with reordering, mixed formats, or specialized execution support
\citep{huang2024slim,li2026scalebits,nie2026sfmp,hooper2025fgmp}.
System-oriented quantizers co-design low-bit formats with weight layouts, dequantization, and GPU data paths to control these deployment costs
\citep{zhao2024atom,lin2025qserve}.
These approaches expose a structural trade-off: coarser sharing restricts where higher precision can be placed, while finer allocation requires coordinating the compact storage of heterogeneous weights with the formats used for GPU computation.
This leads to our central question:
\emph{How can sensitivity-driven local precision choices be preserved in compact weight storage and direct GPU execution?}

Our key insight is to use GPU-compatible two-dimensional \textbf{weight tiles as a shared unit} for local precision selection, storage, and execution.
Each tile receives its own bit-width, allowing precision to vary within output channels while providing a regular unit for storing and loading the selected weights.
Our analysis supports this choice by showing that quantization sensitivity is localized within output channels.
At a matched 4-bit tile fraction, two-dimensional tile allocation improves WikiText-2 perplexity over the output-row control.
However, turning these local precision choices into efficient inference presents two coupled challenges.
First, prefill processes prompt tokens and decode generates new tokens using the same weights.
A shared precision assignment must therefore combine sensitivity evidence from both phases under one model-wide weight-storage budget.
Second, the selected mixed-bit weights must retain their compact format in GPU memory and during loading while supporting integer computation tailored to each phase's workload
\citep{lin2025qserve}.

We introduce \modelname, a tile-aligned mixed-precision method that constructs one compact model for efficient LLM generation.
Joint loss-aware calibration assigns 4-bit or 8-bit precision under a model-wide budget, giving higher-precision priority to tiles important to either phase.
Packing stores one selected representation per tile, preserving 4-bit weights in compact form through storage and loading.
Phase-specialized kernels expand 4-bit weights on-chip to share INT8 computation with 8-bit weights and activations, directly consuming the same packed model.

Across Llama-3.2-3B, Qwen3-4B, Ministral-3-8B, and Qwen3-14B, \modelname achieves maximum $2.50\times$ inference speedups over BF16 on an NVIDIA A100, NVIDIA H100 NVL, and RTX 4090 GPUs while preserving model quality.
These results show how preserving precision granularity through weight storage and computation translates local quantization choices into practical inference efficiency.
Our contributions are as follows:
\begin{enumerate}[nosep,left=0pt]
    \item \textbf{Tile-Aligned Precision Design:}
    We develop a two-dimensional weight-tile organization that permits local bit-width choices within output channels.
    The same partition defines the regions evaluated during calibration and the units used for compact storage and GPU loading.

    \item \textbf{Joint Prefill/Decode Allocation:}
    We develop a loss-aware rule that scores local projection-output changes from 8-to-4-bit substitutions under 8-bit activations.
    Joint prefill/decode calibration produces one precision map under a model-wide storage budget, prioritizing 8-bit retention for tiles important to either phase.

    \item \textbf{Compact Storage and Direct Execution:}
    We develop prefill- and decode-specialized integer kernels that directly consume a compact model with one selected representation per tile.
    Compact 4-bit weight loads and on-chip expansion let both weight formats share INT8 arithmetic for efficient inference.
\end{enumerate}


\section{Related Work}
\label{sec:related_work}
\vspace{-10pt}

Mixed-precision LLM quantization couples three design dimensions: the granularity at which precision varies, the representation used to realize heterogeneous precision, and the numerical transformations used to reduce quantization error.

\textit{Sensitivity-aware precision allocation} has progressively moved from coarse model structure toward localized decisions.
Layer-wise methods reduce the number of precision decisions and preserve regular operators, while coupling regions of different sensitivity to one bit-width across each layer
\citep{lee2025amq,zhao2025impq}.
Channel-, output-feature-, and group-wise methods capture finer heterogeneity through structured units, coupling the weights within each unit to one precision and, under layer-local budgets, constraining high-precision placement
\citep{chen2024channel,huang2024slim,zheng2026mixllm,harma2026mxsens}.
Block-level schemes expose more local choices through hardware-aligned partitions, using global budgets, Fisher-weighted perturbation, or outlier-aware clustering for bit-width selection
\citep{li2026scalebits,nie2026sfmp,hooper2025fgmp,xie2025fineq}.
Per-weight salience aggregation captures additive contributions, while whole-block precision changes interact through the projection output
\citep{nie2026sfmp,hooper2025fgmp}.
Phase-adaptive methods vary precision across inference phases and generation steps, adding runtime selection among quantized weight representations
\citep{chen2025progressive}.
Allocation therefore depends on spatial granularity, sensitivity evidence, and precision sharing across inference phases.

\textit{Low-bit representation and execution} approach the same trade-off from the systems side.
Low-bit systems co-design numerical formats, packing, data movement, and kernels for efficient GPU computation
\citep{dettmers2022gpt3,zhao2024atom,lin2025qserve,hu2025liquidgemm}.
Within this category, uniform weight formats couple sensitive and less sensitive regions to the same bit-width
\citep{lin2025qserve,hu2025liquidgemm}.
Executable mixed-precision systems organize heterogeneous weights through output-feature partitions, reordered blocks, expert groups, or hardware-supported formats, tying their precision choices to corresponding layouts, indexing, or arithmetic paths
\citep{zheng2026mixllm,nie2026sfmp,duanmu2025mxmoe,liu2026micromix,jang2025blockdialect}.
Representation-centric methods use sparse outliers, residual subspaces, reusable binary codes, or fractional-bit quantizers to improve fidelity or precision flexibility, with corresponding indexed accesses, auxiliary products, or specialized reconstruction
\citep{dettmers2023spqr,saxena2024resq,park2026anybcq,lee2026q}.
Inference efficiency therefore depends on how compact representations map to actual GPU arithmetic.

\textit{Transformation-based quantization} reshapes weight and activation distributions to reduce quantization error.
Scaling methods protect salient weights or redistribute quantization difficulty between weights and activations, with flexibility limited to feature-wise rescaling
\citep{xiao2023smoothquant,lin2024awq}.
Rotation- and permutation-based methods redistribute outliers across feature dimensions, while transforms spanning precision partitions change the weights assigned to each quantized region
\citep{ashkboos2024quarot,liu2025spinquant,lin2024duquant}.
Affine and jointly optimized scaling and orthogonal transformations offer richer distribution shaping, while spatial bit-width placement remains a separate decision
\citep{ma2024affinequant,hu2025ostquant,sun2024flatquant}.
Runtime smoothing adapts channel scales during inference to control dynamic ranges, while loss-sensitive weight placement requires a separate allocation criterion
\citep{ICLR2025_2bda52ac}.
These approaches improve numerical fidelity, while executable mixed precision also requires compatible transformation, allocation, and execution partitions.

\vspace{-15pt}
\section{\modelname}
\label{sec:method}
\vspace{-10pt}

\begin{figure*}[t]
    \centering
    \includegraphics[width=\textwidth]{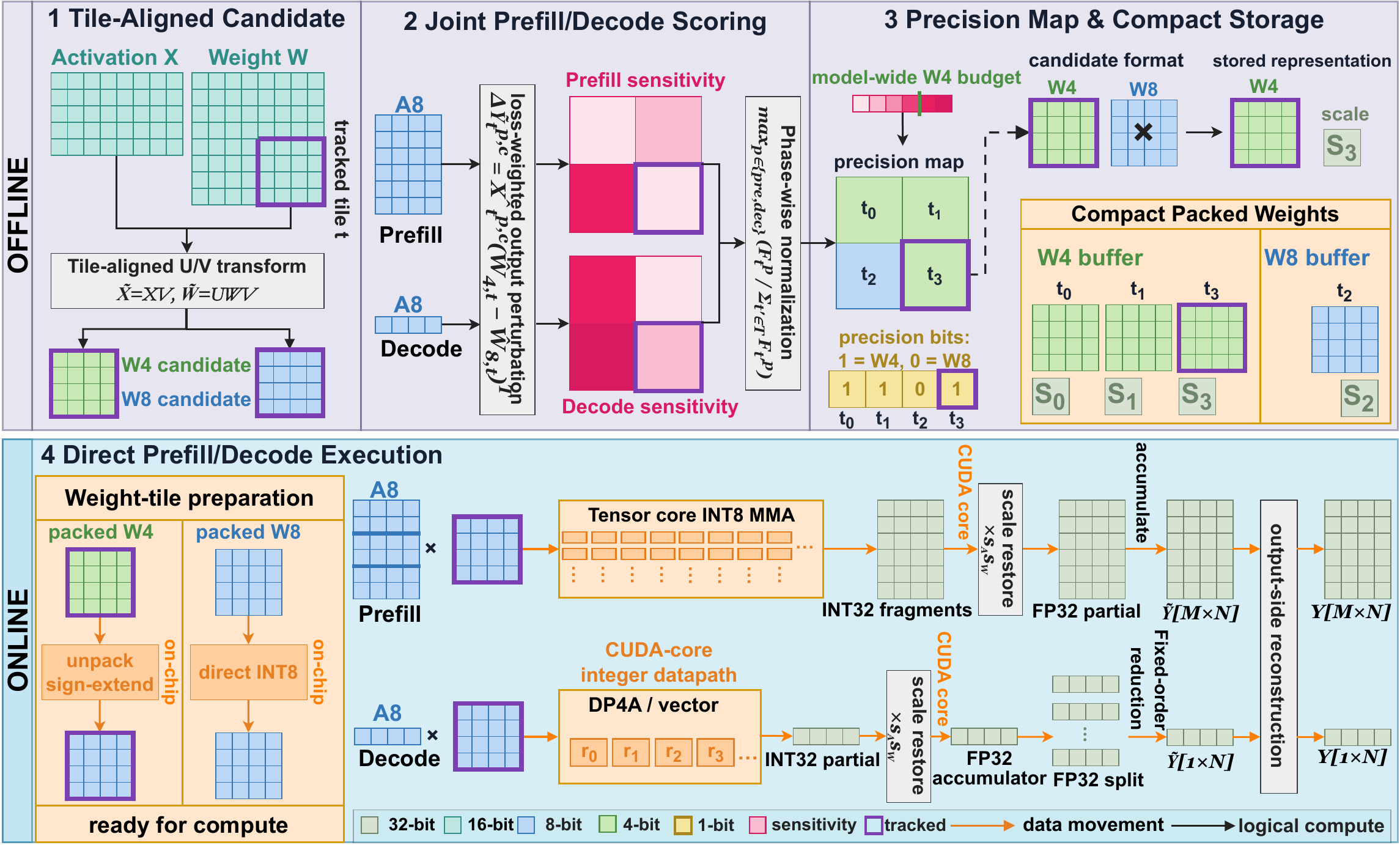}
    \vspace{-10pt}
    \caption{
    \textbf{Overview of \modelname.}
    Purple outlines track one logical weight tile.
    \emph{(1) Candidates.} Tile-aligned $U/V$ transforms prepare 4-bit and 8-bit weight candidates.
    \emph{(2) Joint scoring.} Loss-weighted output changes score each 8-bit to 4-bit replacement in prefill and decode; the larger normalized score determines 8-bit priority.
    \emph{(3) Compact storage.} The budget assigns 4-bit precision to the lowest-scoring tiles, storing one selected representation and scale per tile.
    \emph{(4) Direct execution.} With 8-bit activations, prefill uses Tensor Core multiplication and decode uses DP4A dot products.
    The illustrated path expands 4-bit weights to INT8 on chip; long prefill reuses a temporary INT8 buffer.
    Scaled accumulation and output reconstruction produce $Y$.
    }
    \label{fig:framework}
    \vspace{-10pt}
\end{figure*}

\modelname uses a common two-dimensional weight-tile partition to connect \textbf{precision allocation, compact storage, and GPU execution}.
Each tile uses 4-bit or 8-bit weights, while uniform 8-bit activations provide a common input format for hardware-accelerated INT8 computation.
Offline calibration produces a precision map under a model-wide weight-storage budget.
This map records the selected bit-width at each tile coordinate and determines the weights stored in one packed model shared by prefill and decode.

Figure~\ref{fig:framework} follows a tile from candidate construction and joint calibration to compact storage and integer computation.
We evaluate each replacement of 8-bit weights with 4-bit weights through its local projection-output change, weighted by gradients of the full model loss.
Prefill and decode provide separate calibration evidence, which is combined to select one representation per tile.
The runtime uses the selected weights and scales with execution schedules adapted to the two phases.

\vspace{-10pt}
\subsection{Tile-Level Precision Assignment}
\label{sec:tile_selection}
\vspace{-5pt}

We introduce offline precision allocation to select 8-to-4-bit replacements with the \textbf{smallest estimated impact under a storage budget}. 
The allocation leverages per-tile 4-bit and 8-bit candidates, multi-phase activation and gradient evidence, and a target 4-bit fraction.

For a linear projection $Y=XW^\top$, let $X\in\mathbb{R}^{M\times K}$ be the input activations and $W\in\mathbb{R}^{N\times K}$ the weights, where $M$ is the number of activation rows and $K,N$ are the input and output dimensions. Each row of $W$ corresponds to an output feature and each column to an input feature. We partition all quantized weight matrices into equal-sized two-dimensional tiles, with model-wide set $\mathcal T$. Each $t\in\mathcal T$ is identified by its layer, projection, and matrix position, allowing \textbf{different regions within an output channel to use different bit-widths}.

Let $\rho\in[0,1]$ denote the target fraction of 4-bit tiles. Since all tiles contain the same number of weights, each 8-to-4-bit replacement saves the same storage. Thus, the budget \textbf{determines how many tiles use 4-bit weights, while the scores determine which ones}.

\textbf{Quantized candidates.}
As shown in Figure~\ref{fig:framework}(1), candidate construction applies orthogonal input and output transformations ($V$ along $K$, $U$ along $N$) to redistribute outliers and minimize quantization error \citep{ashkboos2024quarot,liu2025spinquant}, yielding symmetric 4-bit and 8-bit candidates per tile. 
Aligning groups with tile boundaries keeps each transformed weight tile ($\widetilde W=UWV$, $\widetilde X=XV$) within its assigned precision region. 
The paired output reconstruction preserves the floating-point projection (Appendix~\ref{app:tile_transforms}).

Let $\widehat W_{4,t}$ and $\widehat W_{8,t}$ denote the reconstructed 4-bit and 8-bit candidates of the same transformed tile.
Multiplying the values by their own scale puts both candidates in \textbf{the same numerical coordinates for comparison}.
The precision assignment selects the integer values and scale retained for inference.
Calibration and inference use the same 8-bit activation quantizer, so candidate comparisons reflect deployed inputs (Appendices~\ref{app:quantization_details} and~\ref{app:activation_precision}).

\textbf{Output perturbation from precision reduction.}
Joint scoring in panel~(2) of Figure~\ref{fig:framework} begins by measuring the \textbf{local projection-output change} caused by replacing tile $t$'s 8-bit candidate with its 4-bit candidate.
This change depends on both the candidate weight difference and the activations multiplied by it.
Let $p\in\{\mathrm{pre},\mathrm{dec}\}$ denote the inference phase, $\mathcal C_p$ its calibration set, and $c\in\mathcal C_p$ one example.
We write $\widehat X_t^{p,c}$ for the transformed 8-bit activations reconstructed with their scales.
This slice contains the scored activation rows and the input columns associated with tile $t$.
Holding these activations and the other weight-tile candidates fixed, the substitution produces
\vspace{-5pt}
\begin{equation}
\Delta\widetilde Y_t^{p,c}
=
\widehat X_t^{p,c}
\left(
\widehat W_{4,t}
-
\widehat W_{8,t}
\right)^\top.
\label{eq:output-perturbation}
\end{equation}
The matrix product combines candidate errors across the tile's input dimensions, capturing how they \textbf{reinforce or cancel in the projection output}.
We then weight this local change by its importance to the full language-model loss.

\begin{wrapfigure}{r}{0.5\linewidth}
    \vspace{-15pt}
    \centering
    \includegraphics[width=\linewidth]{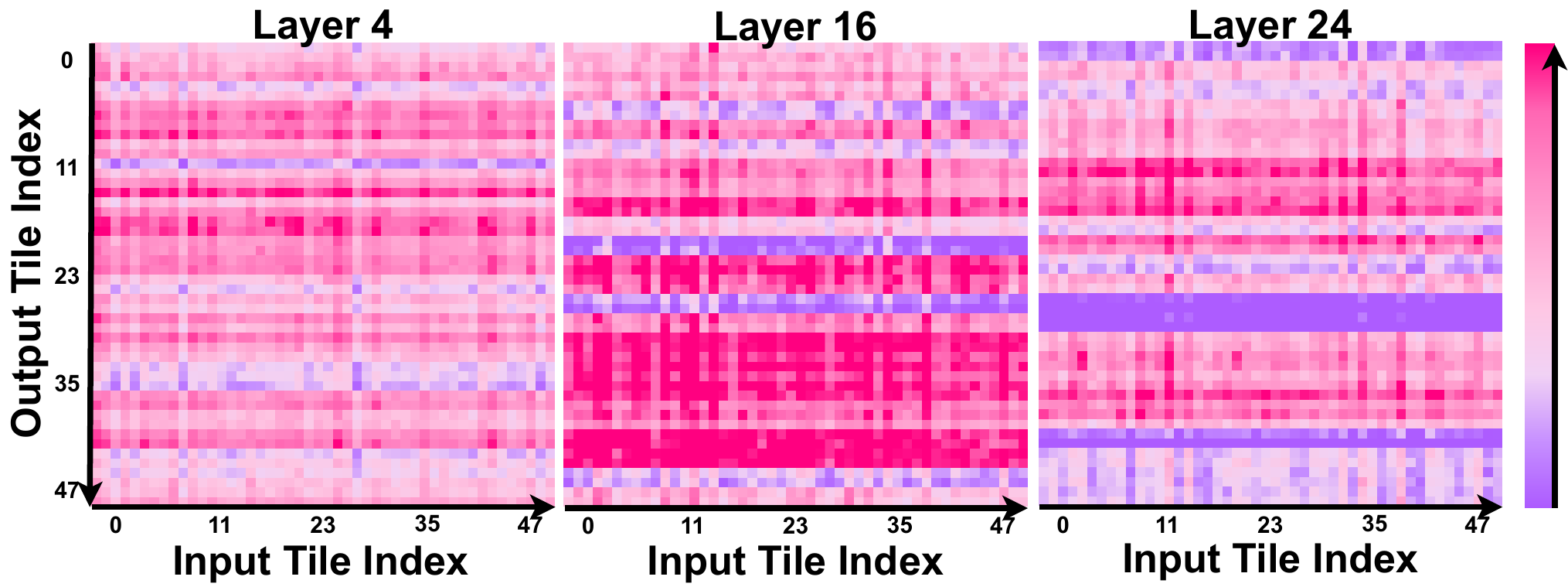}
    \vspace{-15pt}
    \caption{
    Tile-level sensitivity of query-projection weights in Llama-3.2-3B at three decoder depths. Purple and pink indicate lower and higher sensitivity under a shared scale, revealing pronounced within-matrix and across-depth variation.
        }
    \label{fig:qproj}
    \vspace{-20pt}
\end{wrapfigure}

\textbf{Fisher-weighted tile sensitivity.}
Projection-output coordinates differ in their influence on the language-model loss.
We weight squared output perturbations by squared loss gradients using the diagonal empirical Fisher
\citep{hooper2025fgmp}.
The gradients carry information from downstream model computation, giving higher scores to substitutions that affect loss-sensitive outputs.

Let $\mathcal L_c$ denote the full language-model loss averaged over $n_c$ valid prediction tokens.
Let $\widetilde Y_t$ denote the transformed projection-output slice with the same row and feature coordinates as $\Delta\widetilde Y_t^{p,c}$.
Tiles in the same output-feature group share these output coordinates, while their input columns determine their respective perturbations.
The phase-specific score is
\vspace{-5pt}
\begin{equation}
F_t^p
=
\frac{1}{2}
\sum_{c\in\mathcal C_p}
\left\|
 n_c
 \nabla_{\widetilde Y_t}\mathcal L_c
 \odot
 \Delta\widetilde Y_t^{p,c}
\right\|_F^2,
\label{eq:fisher-score}
\end{equation}
where $\odot$ denotes elementwise multiplication and $\|\cdot\|_F$ is the Frobenius norm.
The factor $n_c$ restores the summed-loss gradient from the token-averaged loss.
The squared norm sums the squared, gradient-weighted perturbations over the scored rows and output features, and the outer sum aggregates calibration examples within the phase.

The score combines \textbf{the local projection change} with its \textbf{downstream loss sensitivity}.
Forming the perturbation before weighting captures the combined effect of changing all weights in the tile together.
A larger $F_t^p$ gives tile $t$ higher priority for retaining 8-bit weights in phase $p$.
Figure~\ref{fig:qproj} shows local sensitivity variation within query-projection weights at three decoder depths, illustrating the value of assigning precision to individual tiles.
Calibration forward and gradient-capture details are given in Appendix~\ref{app:calibration_details}, and phase-specific observation construction in Appendix~\ref{app:phase_calibration}.

\textbf{Joint prefill/decode allocation.}
Prefill and decode supply distinct activation and gradient observations for the same weights.
To combine their evidence in one precision map, we first express each tile's score relative to its phase's model-wide total.
We then retain the larger relative score:
\vspace{-5pt}
\begin{equation}
F_t
=
\max_{p\in\{\mathrm{pre},\mathrm{dec}\}}
\frac{F_t^p}
{\sum_{t'\in\mathcal T}F_{t'}^p}.
\label{eq:joint-score}
\end{equation}
Each ratio expresses tile $t$'s share of one phase's total sensitivity score.
Normalization makes the phase scores comparable, while the maximum retains \textbf{each tile's greater relative sensitivity across the two phases}.
A tile with high relative sensitivity in either phase therefore receives higher priority for retaining 8-bit weights.

\begin{figure*}[t]
    \centering
    \includegraphics[width=\textwidth]{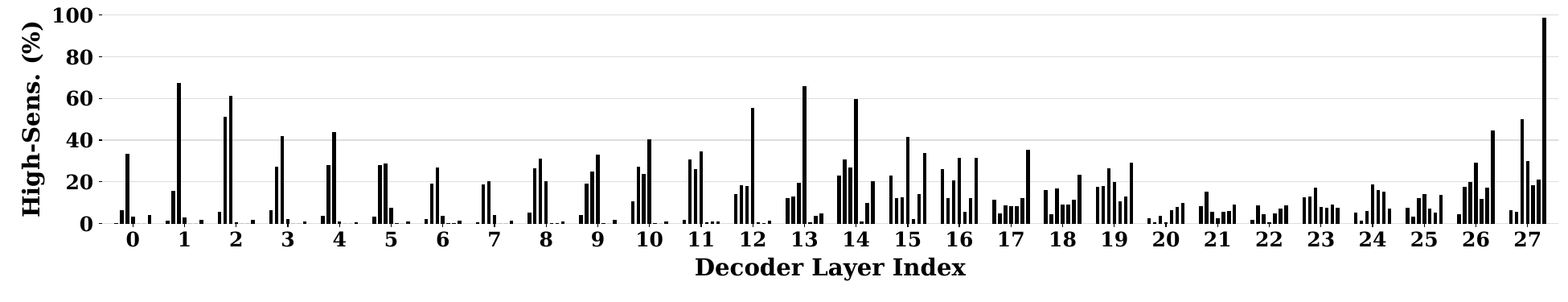}
    \vspace{-20pt}
    \caption{
    Distribution of high-sensitivity tiles across layers.
    Each group contains 7 linear projections, ordered as the query, key, value, attention output projections, feed-forward gate, up, and down projections.
    Bar height reports the percentage of tiles in each projection whose sensitivity scores fall within the global top 10\%, using a single threshold over all tiles in the model.
    The strong concentration across layers and projections motivates model-wide precision allocation.
    }
    \label{fig:distribution}
    \vspace{-15pt}
\end{figure*}

\modelname assigns 4-bit weights to the $\operatorname{round}(\rho|\mathcal T|)$ lowest-scoring tiles and 8-bit weights to the rest.
Since each 4-bit selection saves the same number of weight bytes, the \textbf{model-wide ranking determines where the fixed high-precision budget is spent}.
Figure~\ref{fig:distribution} shows that high-sensitivity tiles are concentrated unevenly across layers and projections.
Model-wide ranking lets the allocation of 8-bit weights follow this distribution, with a common budget coordinating all quantized projections.
Tie handling and the fixed-budget selection argument are given in Appendix~\ref{app:allocation_details}.

\vspace{-8pt}
\subsection{Compact Mixed-Precision Representation}
\label{sec:representation}
\vspace{-5pt}

In panel~(3) of Figure~\ref{fig:framework}, packing stores \textbf{one selected integer representation per tile} together with its FP32 reconstruction scale.
Each byte stores two signed 4-bit values or one signed 8-bit value, so a 4-bit tile uses \textbf{half the integer-weight storage} of an equally sized 8-bit tile.
The selected scale accompanies those values through storage and computation.

Selected 4-bit and 8-bit tiles occupy separate contiguous buffers.
A one-bit precision label selects the buffer for each logical tile.
Within that buffer, the row offset marks the start of its output-tile row, and the count of preceding same-precision tiles determines its position.
This mapping lets the runtime \textbf{locate the selected weights and scale from the original tile coordinate}.
Indexing rules, bit ordering, and kernel layouts are detailed in Appendix~\ref{app:packing_details}.

With a target fraction $\rho$ of 4-bit tiles, nominal integer-weight storage is $4\rho+8(1-\rho)=8-4\rho$ bits per quantized weight; the realized value uses the rounded tile count.
Complete packed storage also includes reconstruction scales and indexing metadata.
Appendices~\ref{app:storage_accounting} and~\ref{app:storage_measurements} give storage accounting and memory-measurement details.

\vspace{-8pt}
\subsection{Direct Prefill/Decode Execution}
\label{sec:runtime}
\vspace{-5pt}

The execution stage prepares the selected weights for GPU integer computation (Figure~\ref{fig:framework}, panel~(4)).
Both weight formats use a common INT8 path, with schedules that exploit weight reuse in prefill and parallel dot products in decode.

\textbf{Integer operand preparation.}
The runtime first applies the input-side transform $V$ and quantizes the activations to 8 bits using the same numerical procedure as calibration.
In the direct mixed-tile kernels, 4-bit values \textbf{stay packed during GPU-memory loading and expand on chip} into signed INT8 operands.
Sign extension preserves each selected integer value, and the stored scale preserves the reconstructed value of the 4-bit candidate.
8-bit weights provide signed INT8 operands directly.

\textbf{Prefill weight reuse.}
Prefill processes multiple activation rows, so its Tensor Core INT8 schedule reuses each weight tile across their output computations.
Short-prefill kernels prepare mixed-tile operands on chip.
For long-prefill projections containing both formats, the runtime expands the selected weights into a temporary INT8 buffer in GPU memory once per projection call.
The Tensor Core schedule \textbf{reuses the prepared weights across activation-row groups}, reducing the conversion cost per group while preserving each tile's values and scale.

\textbf{Decode parallelism.}
Single-sequence decode has one activation row ($M=1$), giving a matrix-vector workload.
The kernel directly loads the packed weight tiles and forms INT8 operands on chip.
DP4A accumulates four products of signed 8-bit operands per instruction into an INT32 result.
Work is distributed \textbf{across output features and input-column groups}, allowing multiple GPU threads to contribute to each output in parallel.
Their partial dot products are combined, drawing parallelism from both weight-matrix dimensions.

\textbf{Output reconstruction.}
For both schedules, each tile product first accumulates in INT32.
The corresponding activation and weight scales restore its contribution before FP32 accumulation across input tiles.
This per-tile rescaling combines contributions from both weight precisions in \textbf{the same numerical coordinates}.
Split-reduction schedules combine their FP32 partials in a fixed order, after which the output-side transform $U$ restores the projection output.
The phase transition reuses the same precision map and selected weights.

\textbf{Shared activation preparation.}
Projections with matching inputs and transforms (e.g., Q/K/V or gate/up) share input preparation. 
Each group \textbf{transforms and quantizes its input once}, reusing 8-bit activations across projections while retaining independent weights, scales, and output transforms (Appendices~\ref{app:execution_details} and~\ref{app:shared_activation}).

Together, these designs preserve local precision flexibility while making each 4-bit choice reduce stored weight bytes.
Joint calibration and phase-specific weight preparation support efficient deployment and generation with one compact model.


\begin{table*}[t]

\centering
\scriptsize
\setlength{\tabcolsep}{3.0pt}
\renewcommand{\arraystretch}{0.75}
\caption{
End-to-end inference efficiency for \textbf{Llama-3.2-3B-Instruct}.
Cell colors are relative to BF16 under the corresponding metric direction: green for better than or equal to BF16, blue for worse than BF16, and gray for BF16.
Bold and underlined values indicate the best and worst results, respectively, among the compared methods excluding the reference columns.
}
\label{tab:system_llama3b}
\vspace{-10pt}
\begin{adjustbox}{width=\textwidth}
\begin{tabular}{l l c c c c c c c}
\toprule
\multirow{2}{*}{\textbf{Metric}} &
\multirow{2}{*}{\textbf{Unit}} &
\multicolumn{1}{c}{\textbf{Upper Ref.}} &
\multicolumn{1}{c}{\textbf{Lower Ref.}} &
\multicolumn{1}{c}{\textbf{Ours}} &
\multicolumn{4}{c}{\textbf{Others}} \\
\cmidrule(lr){3-3}
\cmidrule(lr){4-4}
\cmidrule(lr){5-5}
\cmidrule(lr){6-9}
& &
\textbf{BF16} &
\textbf{W4A8} &
\textbf{\modelname} &
\textbf{GPTQ} &
\textbf{AWQ} &
\textbf{MixLLM} &
\textbf{SFMP} \\
\midrule

TTFT $\downarrow$
& ms
\setbase{17.47}
& \cellcolor{heatneutral}\base
& \spcell{22.51}
& \textbf{\spcell{19.67}}
& \spcell{21.99}
& \spcell{23.07}
& \spcell{27.48}
& \underline{\spcell{86.99}} \\

TPOT $\downarrow$
& ms/token
\setbase{17.19}
& \cellcolor{heatneutral}\base
& \spcell{21.12}
& \textbf{\spcell{7.54}}
& \spcell{20.77}
& \spcell{21.78}
& \spcell{24.05}
& \underline{\spcell{82.33}} \\

Throughput $\uparrow$
& tok/s
\setbase{58.18}
& \cellcolor{heatneutral}\base
& \sgcell{47.33}
& \textbf{\sgcell{132.28}}
& \sgcell{48.13}
& \sgcell{45.91}
& \sgcell{41.58}
& \underline{\sgcell{12.15}} \\

Resident GPU Mem. $\downarrow$
& GiB
\setbase{5.99}
& \cellcolor{heatneutral}\base
& \spcell{2.08}
& \spcell{2.72}
& \textbf{\spcell{2.11}}
& \spcell{2.12}
& \underline{\spcell{2.87}}
& \spcell{2.15} \\

Peak VRAM $\downarrow$
& GiB
\setbase{6.09}
& \cellcolor{heatneutral}\base
& \spcell{2.16}
& \spcell{2.72}
& \textbf{\spcell{2.20}}
& \spcell{2.21}
& \underline{\spcell{2.92}}
& \spcell{2.37} \\

Energy $\downarrow$
& J/1K tokens
\setbase{2280.68}
& \cellcolor{heatneutral}\base
& \spcell{2322.34}
& \textbf{\spcell{1267.92}}
& \spcell{2047.43}
& \spcell{1732.24}
& \spcell{2172.71}
& \underline{\spcell{22535.65}} \\

\bottomrule
\end{tabular}
\end{adjustbox}
\vspace{-12pt}
\end{table*}

\vspace{-10pt}
\section{Experiments}
\label{sec:experiments}
\vspace{-10pt}

\subsection{Experimental Setup}
\label{sec:setup}
\vspace{-5pt}

We evaluate whether \modelname improves inference efficiency while preserving model quality, with allocation controls examining the value of local precision placement.
The main text focuses on Llama-3.2-3B-Instruct; Appendix~\ref{app:additional-results} extends the evaluation to Qwen3-4B-Instruct-2507, Ministral-3-8B-Instruct-2512-BF16, and Qwen3-14B.

\noindent
\textbf{Datasets and Metrics.}
We evaluate language modeling via WikiText-2 perplexity and reasoning/understanding via HellaSwag, MMLU, ARC-Challenge, LAMBADA, PIQA, and WinoGrande. 
Long-context capabilities are assessed using LV-Eval QA (16K--64K length levels) and LongEval line-retrieval exact-match accuracy.

\noindent
\textbf{Configuration and Comparisons.}
The default model uses $64\times64$ tiles, 60\% 4-bit weights, and 8-bit activations.
BF16 is the original-model reference; uniform 4-bit weights provide the low-bit reference.
Structural and random controls test placement (Appendix~\ref{app:comparison_configs}), alongside GPTQ, AWQ, MixLLM, and SFMP \citep{frantar2022gptq,lin2024awq,zheng2026mixllm,nie2026sfmp}.

\noindent
\textbf{Hardware and Measurement.}
Primary measurements use an NVIDIA A100-SXM4 40\,GB GPU, with additional Llama-3.2-3B tests on H100 NVL and RTX 4090. 
We report time to first token (TTFT), time per output token (TPOT), throughput, memory, and energy over the full inference path (Appendix~\ref{app:experimental-details}).

\vspace{-5pt}
\subsection{Inference Efficiency}
\label{sec:efficiency}
\vspace{-5pt}

BF16 retains the original optimized model; Hugging Face FP16, shown as Torch FP16, provides the FP16 reference used in prior quantization evaluations \citep{lin2024awq}.
On Llama-3.2-3B, \modelname reaches 132.28 tokens/s, \textbf{$2.27\times$ BF16 and $2.75\times$ GPTQ throughput}, while reducing resident memory from 5.99 to 2.72\,GiB and achieving the lowest energy in Table~\ref{tab:system_llama3b}.
The compact selected representation therefore converts low-bit projection efficiency into complete-model generation gains.

\begin{figure*}[t]
    \centering

    \includegraphics[width=0.274\linewidth]{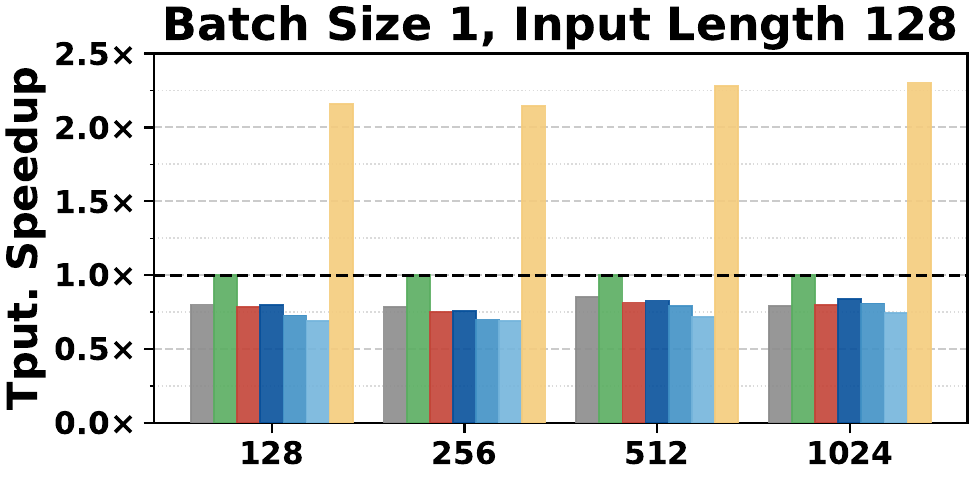}
    \hfill
    \includegraphics[width=0.234\linewidth]{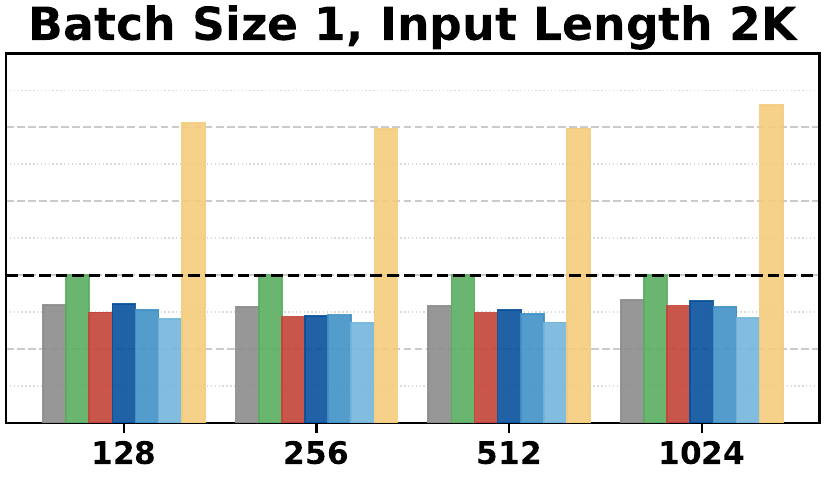}
    \hfill
    \includegraphics[width=0.234\linewidth]{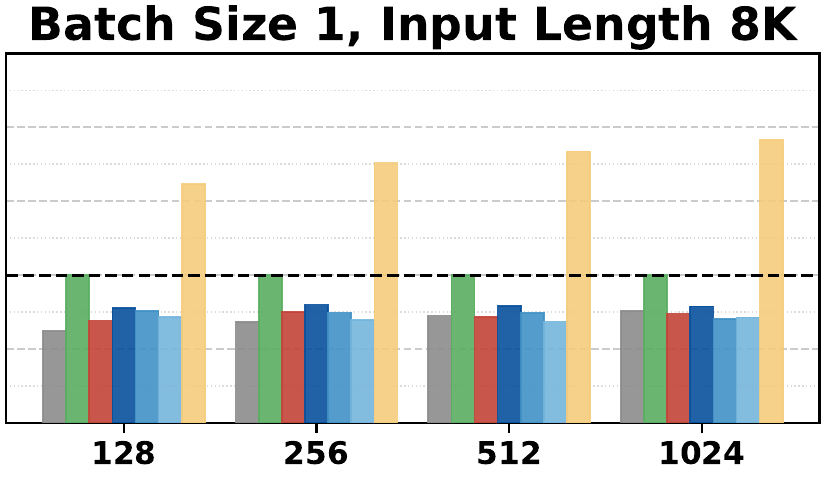}
    \hfill
    \includegraphics[width=0.234\linewidth]{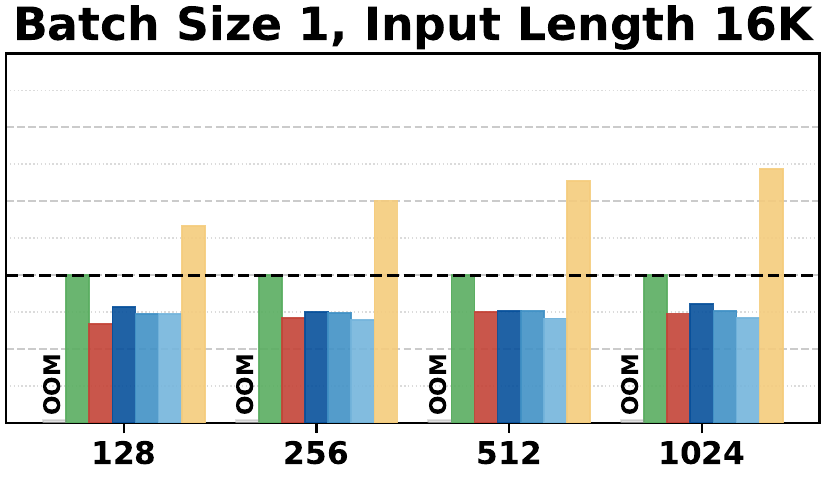}

    \vspace{-3pt}

    \includegraphics[width=0.274\linewidth]{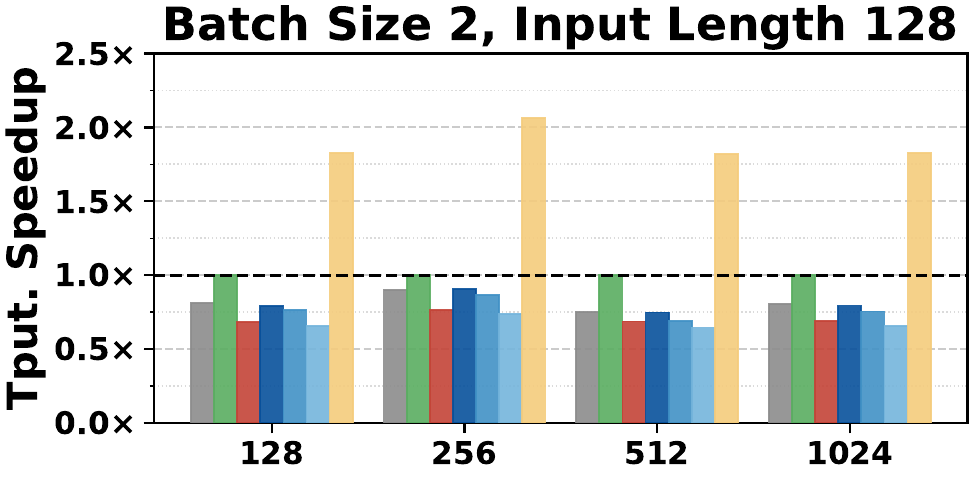}
    \hfill
    \includegraphics[width=0.234\linewidth]{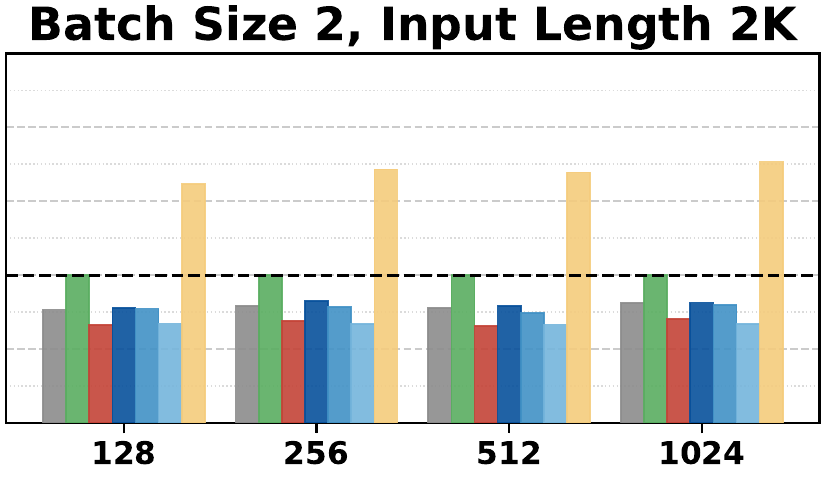}
    \hfill
    \includegraphics[width=0.234\linewidth]{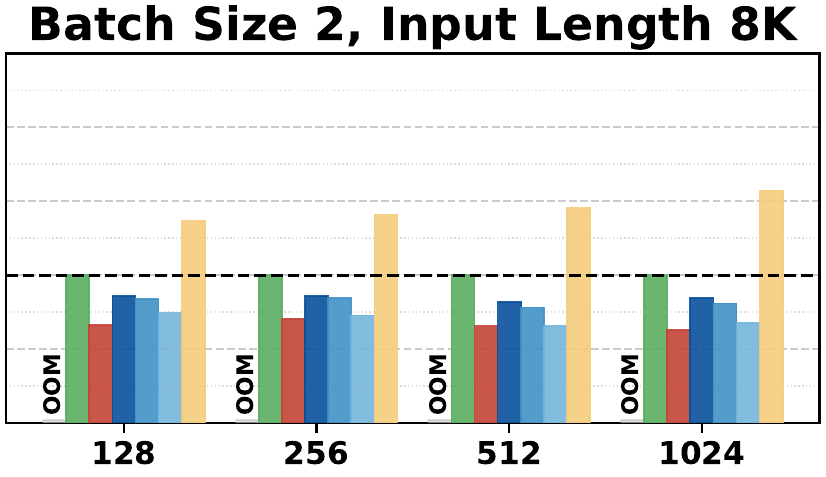}
    \hfill
    \includegraphics[width=0.234\linewidth]{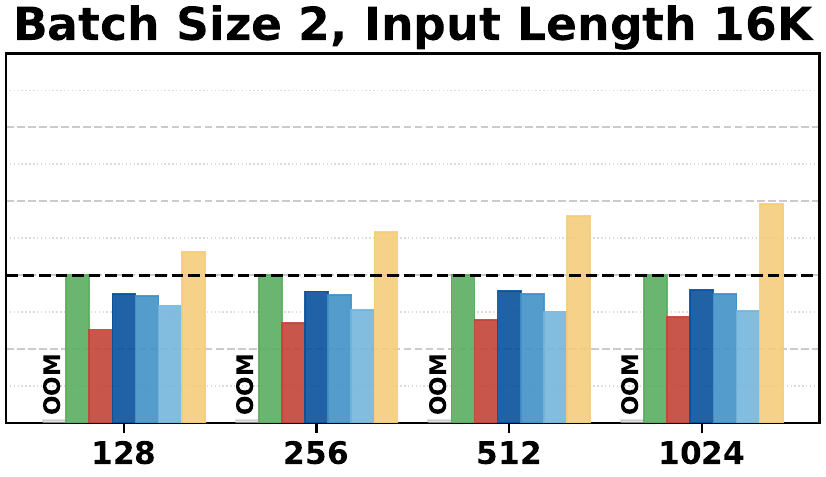}

    \vspace{-3pt}

    \includegraphics[width=0.274\linewidth]{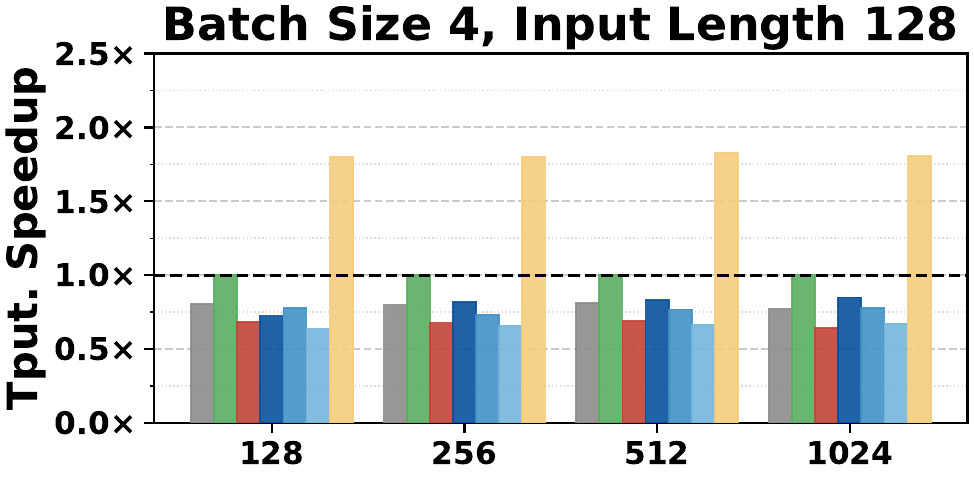}
    \hfill
    \includegraphics[width=0.234\linewidth]{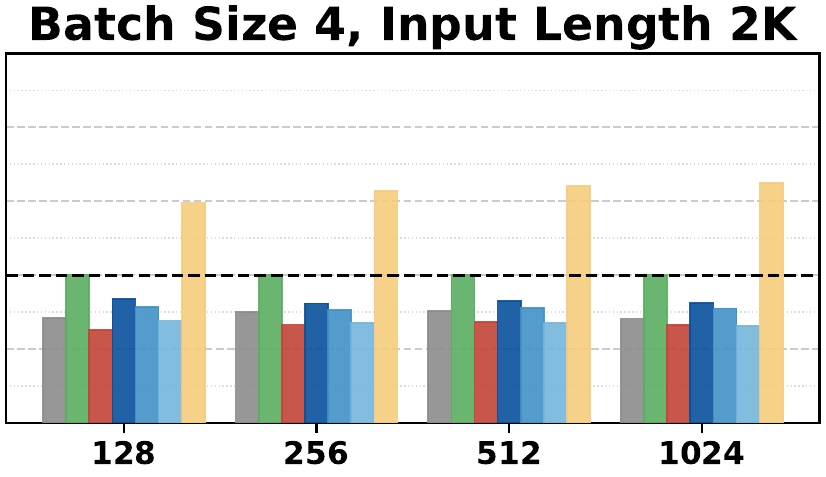}
    \hfill
    \includegraphics[width=0.234\linewidth]{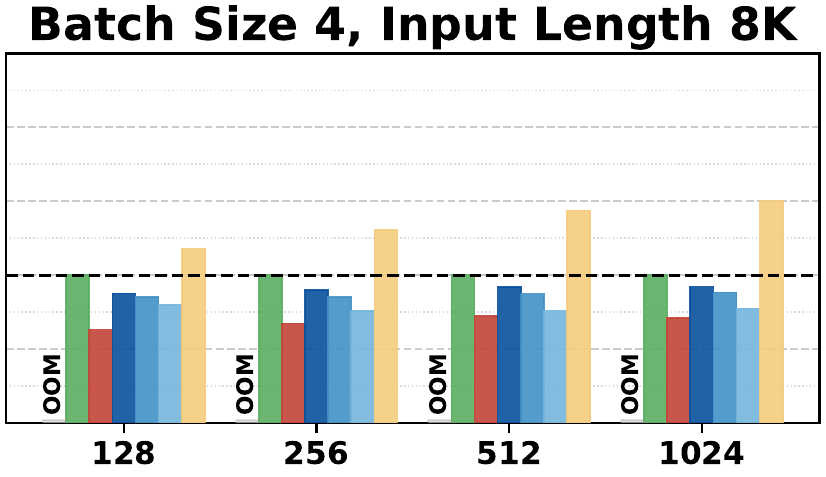}
    \hfill
    \includegraphics[width=0.234\linewidth]{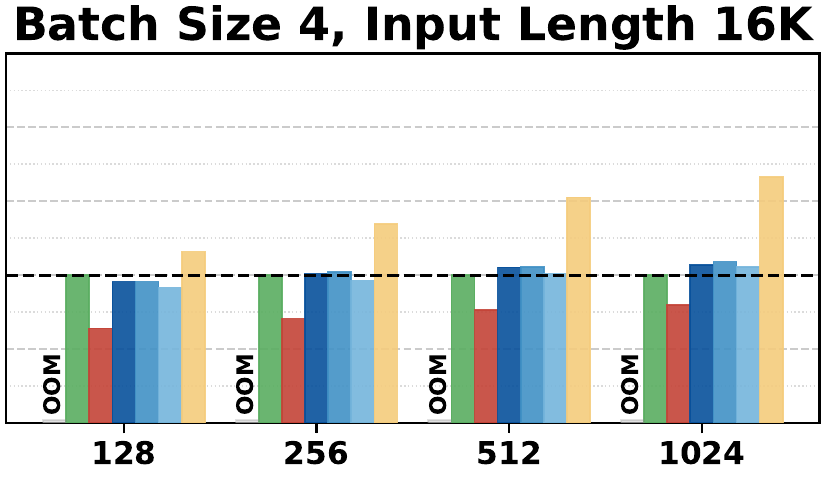}

    \vspace{-3pt}

    \includegraphics[width=0.274\linewidth]{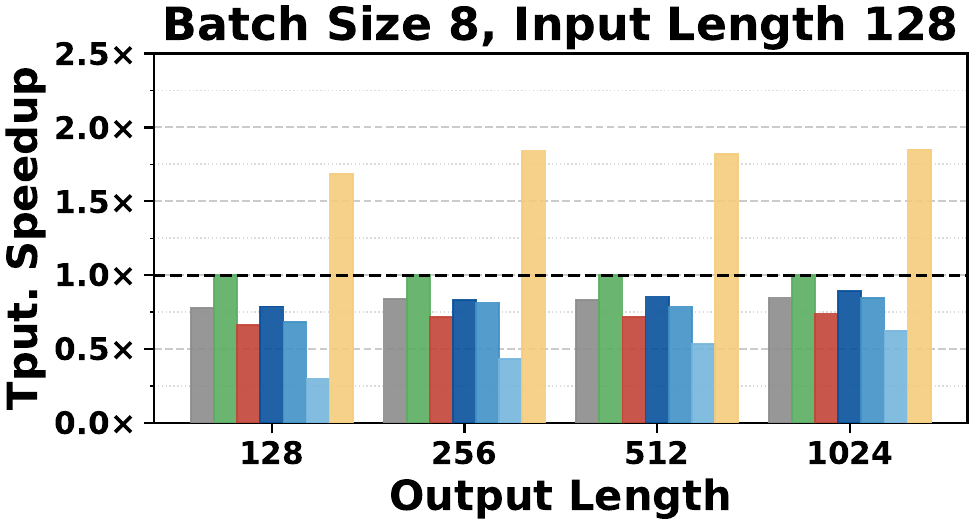}
    \hfill
    \includegraphics[width=0.234\linewidth]{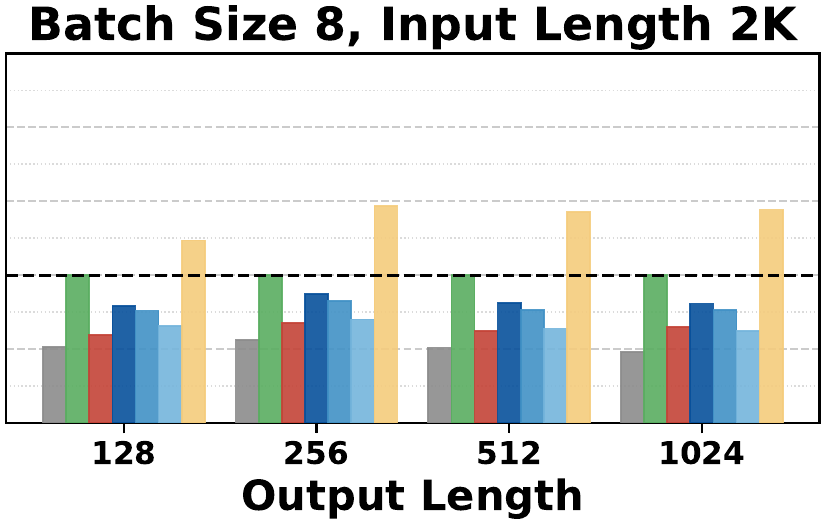}
    \hfill
    \includegraphics[width=0.234\linewidth]{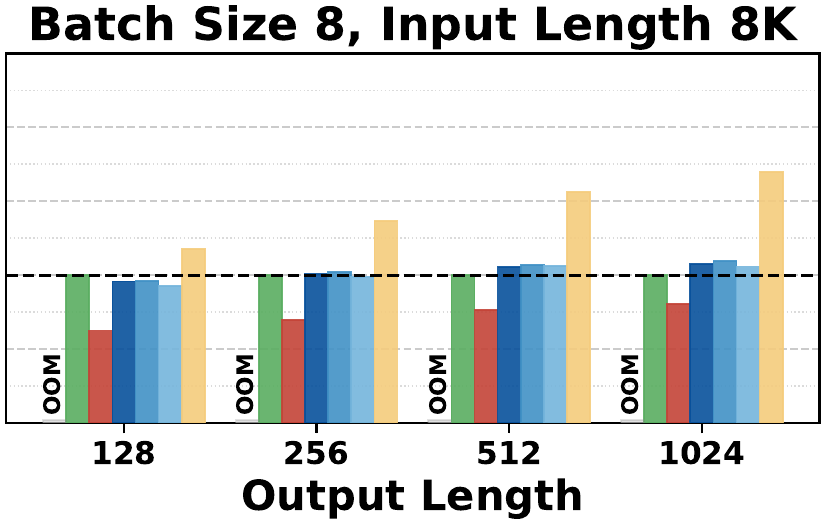}
    \hfill
    \includegraphics[width=0.234\linewidth]{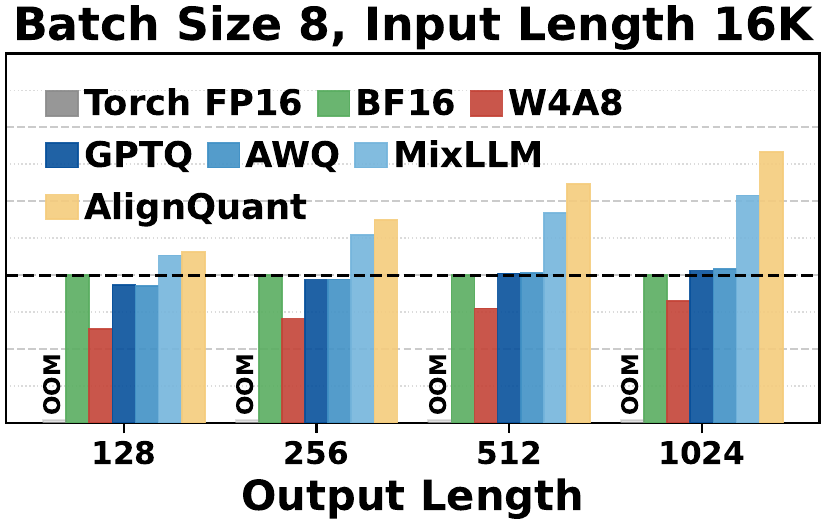}

    \vspace{-8pt}
    \caption{
    Generation-throughput speedup relative to BF16 on Llama-3.2-3B.
    Rows vary batch size over 1, 2, 4, and 8; columns vary prompt length over 128, 2K, 8K, and 16K tokens.
    Each panel varies output length; OOM denotes an out-of-memory run.
    }
    \label{fig:generation_throughput_grid}
    \vspace{-15pt}
\end{figure*}

Varying request shape shows how the benefit changes with the balance between prompt processing and generation.
\modelname leads the displayed throughput comparisons in Figures~\ref{fig:intro} and~\ref{fig:generation_throughput_grid}.
For long prompts, \textbf{longer continuations amplify the gain over BF16}: repeated token generation gives the packed-load decode path a larger share of the request.
The same selected model therefore supports different request shapes while its execution schedules exploit their available reuse and parallelism.
Additional system results and runtime ablations appear in Appendices~\ref{app:system_results} and~\ref{app:runtime_results}.

\begin{table*}[t]

\centering
\scriptsize
\setlength{\tabcolsep}{2.5pt}
\renewcommand{\arraystretch}{0.75}
\caption{
General and ablation results for \textbf{Llama-3.2-3B-Instruct}.
Accuracy values are reported as percentages, while WikiText-2 is reported with word perplexity.
Cell colors are relative to BF16 under the corresponding metric direction: green for better than or equal to BF16, blue for worse than BF16, and gray for BF16.
Bold values indicate the best result among the non-reference methods.
}
\label{tab:general_llama3b}
\vspace{-10pt}
\begin{adjustbox}{width=\textwidth}
\begin{tabular}{l @{\hspace{2pt}} c c c c c c c c c c c c}
\toprule
\multirow{2}{*}{\textbf{Benchmark}} &
\multirow{2}{*}{\textbf{Metric}} &
\multicolumn{1}{c}{\textbf{Upper Ref.}} &
\multicolumn{1}{c}{\textbf{Lower Ref.}} &
\multicolumn{1}{c}{\textbf{Ours}} &
\multicolumn{4}{c}{\textbf{Others}} &
\multicolumn{4}{c}{\textbf{Ablation}} \\
\cmidrule(lr){3-3}
\cmidrule(lr){4-4}
\cmidrule(lr){5-5}
\cmidrule(lr){6-9}
\cmidrule(lr){10-13}
& &
\textbf{BF16} &
\textbf{W4A8} &
\textbf{\modelname} &
\textbf{GPTQ} &
\textbf{AWQ} &
\textbf{MixLLM} &
\textbf{SFMP} &
\textbf{Output-row} &
\textbf{Red.-col} &
\textbf{Proj./layer} &
\textbf{Random} \\
\midrule

WikiText-2 $\downarrow$
& PPL \setbase{12.49}
& \cellcolor{heatneutral}\base
& \pcell{13.51}
& \textbf{\pcell{12.61}}
& \pcell{13.27}
& \pcell{12.97}
& \pcell{12.69}
& \pcell{12.88}
& \pcell{12.71}
& \pcell{12.63}
& \pcell{12.73}
& \pcell{12.88} \\

HellaSwag $\uparrow$
& Acc. \setbase{71.52}
& \cellcolor{heatneutral}\base
& \gcell{68.88}
& \textbf{\gcell{70.74}}
& \gcell{52.62}
& \gcell{50.41}
& \gcell{51.44}
& \gcell{70.38}
& \gcell{70.53}
& \textbf{\gcell{70.74}}
& \gcell{70.45}
& \gcell{70.24} \\

MMLU $\uparrow$
& Acc. \setbase{60.66}
& \cellcolor{heatneutral}\base
& \gcell{58.30}
& \textbf{\gcell{60.53}}
& \gcell{59.91}
& \gcell{59.24}
& \gcell{60.50}
& \gcell{58.53}
& \gcell{60.05}
& \gcell{59.81}
& \gcell{59.79}
& \gcell{59.89} \\

ARC-C $\uparrow$
& Acc. \setbase{52.56}
& \cellcolor{heatneutral}\base
& \gcell{51.11}
& \textbf{\gcell{52.73}}
& \gcell{45.73}
& \gcell{47.44}
& \gcell{47.35}
& \gcell{50.77}
& \gcell{50.68}
& \gcell{51.88}
& \gcell{50.68}
& \gcell{50.34} \\

LAMBADA $\uparrow$
& Acc. \setbase{64.23}
& \cellcolor{heatneutral}\base
& \gcell{61.65}
& \textbf{\gcell{65.66}}
& \gcell{63.79}
& \gcell{64.42}
& \gcell{65.28}
& \gcell{64.55}
& \gcell{63.67}
& \gcell{63.30}
& \gcell{63.57}
& \gcell{63.81} \\

PIQA $\uparrow$
& Acc. \setbase{77.31}
& \cellcolor{heatneutral}\base
& \gcell{76.61}
& \textbf{\gcell{76.68}}
& \gcell{75.08}
& \gcell{75.73}
& \gcell{76.01}
& \gcell{76.22}
& \gcell{76.61}
& \gcell{75.79}
& \gcell{76.44}
& \gcell{76.66} \\

WinoGrande $\uparrow$
& Acc. \setbase{71.43}
& \cellcolor{heatneutral}\base
& \gcell{69.22}
& \textbf{\gcell{71.22}}
& \gcell{70.24}
& \gcell{68.75}
& \gcell{70.43}
& \gcell{68.82}
& \gcell{69.53}
& \gcell{69.61}
& \gcell{69.61}
& \gcell{68.75} \\

\bottomrule
\end{tabular}
\end{adjustbox}
\vspace{-10pt}
\end{table*}

\vspace{-5pt}
\subsection{General Quality}
\label{sec:model-performance}
\vspace{-5pt}

Beyond generation speed, we evaluate general language tasks (Table~\ref{tab:general_llama3b}). 
\modelname achieves \textbf{the lowest perplexity and highest or tied-highest accuracy} among quantized configurations, showing that one loss-aware allocation supports diverse tasks while retaining efficient inference.

\textbf{Precision-allocation ablation.}
\modelname reaches 65.66\% on LAMBADA and 71.22\% on WinoGrande, versus 63.30\%--63.81\% and 68.75\%--69.61\% for structured and random placements.
These gains support \textbf{assigning precision to local regions within output channels}; additional models appear in Appendices~\ref{app:general_results} and~\ref{app:ablation_results}.

\begin{table*}[!t]

\centering
\scriptsize
\setlength{\tabcolsep}{2.0pt}
\renewcommand{\arraystretch}{0.75}
\caption{
Long-context QA results for
\textbf{Llama-3.2-3B-Instruct} on LV-Eval.
Scores are reported as percentages.
Cell colors are relative to BF16: green for higher scores,
blue for lower scores, and gray for BF16.
Bold values indicate the best result among the non-reference methods.
}
\label{tab:lveval_llama3b}
\vspace{-10pt}
\begin{adjustbox}{width=\textwidth}
\begin{tabular}{c c c c c c c c c c c c c}
\toprule
\multirow{2}{*}{\textbf{Dataset}} &
\multirow{2}{*}{\textbf{Len}} &
\multicolumn{1}{c}{\textbf{Upper Ref.}} &
\multicolumn{1}{c}{\textbf{Lower Ref.}} &
\multicolumn{1}{c}{\textbf{Ours}} &
\multicolumn{4}{c}{\textbf{Others}} &
\multicolumn{4}{c}{\textbf{Ablation}} \\
\cmidrule(lr){3-3}
\cmidrule(lr){4-4}
\cmidrule(lr){5-5}
\cmidrule(lr){6-9}
\cmidrule(lr){10-13}
& &
\textbf{BF16} &
\textbf{W4A8} &
\textbf{\modelname} &
\textbf{GPTQ} &
\textbf{AWQ} &
\textbf{MixLLM} &
\textbf{SFMP} &
\textbf{Output-row} &
\textbf{Red.-col} &
\textbf{Proj./layer} &
\textbf{Random} \\
\midrule

\multirow{3}{*}{\ds{1}}
& 16k \setbase{21.41}
& \cellcolor{heatneutral}\base
& \gcell{17.78}
& \textbf{\gcell{22.28}}
& \gcell{16.63}
& \gcell{18.29}
& \gcell{20.05}
& \gcell{18.21}
& \gcell{21.61}
& \gcell{21.52}
& \gcell{19.05}
& \gcell{18.43} \\
& 32k \setbase{18.61}
& \cellcolor{heatneutral}\base
& \gcell{13.55}
& \textbf{\gcell{17.97}}
& \gcell{13.85}
& \gcell{16.39}
& \gcell{16.13}
& \gcell{15.57}
& \gcell{15.43}
& \gcell{15.67}
& \gcell{14.98}
& \gcell{13.58} \\
& 64k \setbase{14.19}
& \cellcolor{heatneutral}\base
& \gcell{11.75}
& \textbf{\gcell{13.14}}
& \gcell{12.02}
& \gcell{11.99}
& \gcell{12.40}
& \gcell{12.88}
& \gcell{12.66}
& \gcell{12.87}
& \gcell{11.93}
& \gcell{11.94} \\
\midrule

\multirow{3}{*}{\ds{2}}
& 16k \setbase{16.46}
& \cellcolor{heatneutral}\base
& \gcell{15.05}
& \textbf{\gcell{17.42}}
& \gcell{16.23}
& \gcell{15.41}
& \gcell{17.20}
& \gcell{17.15}
& \gcell{17.38}
& \gcell{17.34}
& \gcell{16.63}
& \gcell{16.57} \\
& 32k \setbase{17.16}
& \cellcolor{heatneutral}\base
& \gcell{15.18}
& \textbf{\gcell{15.99}}
& \gcell{15.23}
& \gcell{15.97}
& \gcell{15.29}
& \gcell{15.73}
& \gcell{15.20}
& \gcell{15.08}
& \gcell{15.14}
& \gcell{14.69} \\
& 64k \setbase{17.17}
& \cellcolor{heatneutral}\base
& \gcell{14.69}
& \textbf{\gcell{17.40}}
& \gcell{14.94}
& \gcell{16.84}
& \gcell{16.52}
& \gcell{15.86}
& \gcell{15.42}
& \gcell{14.98}
& \gcell{14.75}
& \gcell{14.12} \\
\midrule

\multirow{3}{*}{\ds{3}}
& 16k \setbase{23.76}
& \cellcolor{heatneutral}\base
& \gcell{20.90}
& \textbf{\gcell{23.39}}
& \gcell{20.46}
& \gcell{22.44}
& \gcell{23.01}
& \gcell{22.73}
& \gcell{18.99}
& \gcell{22.02}
& \gcell{21.58}
& \gcell{21.18} \\
& 32k \setbase{29.75}
& \cellcolor{heatneutral}\base
& \gcell{19.55}
& \textbf{\gcell{29.16}}
& \gcell{27.97}
& \gcell{27.59}
& \gcell{28.35}
& \gcell{28.19}
& \gcell{22.88}
& \gcell{26.87}
& \gcell{26.44}
& \gcell{22.76} \\
& 64k \setbase{24.79}
& \cellcolor{heatneutral}\base
& \gcell{20.01}
& \textbf{\gcell{24.38}}
& \gcell{23.20}
& \gcell{23.54}
& \gcell{23.80}
& \gcell{23.70}
& \gcell{22.70}
& \gcell{23.81}
& \gcell{23.87}
& \gcell{22.24} \\
\midrule

\multirow{3}{*}{\ds{4}}
& 16k \setbase{11.23}
& \cellcolor{heatneutral}\base
& \gcell{9.57}
& \gcell{12.86}
& \gcell{9.85}
& \gcell{12.65}
& \gcell{10.38}
& \gcell{12.38}
& \textbf{\gcell{13.21}}
& \gcell{12.60}
& \gcell{11.92}
& \gcell{11.48} \\
& 32k \setbase{11.74}
& \cellcolor{heatneutral}\base
& \gcell{8.93}
& \textbf{\gcell{11.83}}
& \gcell{9.24}
& \gcell{11.09}
& \gcell{10.70}
& \gcell{11.27}
& \gcell{11.01}
& \gcell{11.30}
& \gcell{10.78}
& \gcell{9.97} \\
& 64k \setbase{9.77}
& \cellcolor{heatneutral}\base
& \gcell{7.13}
& \textbf{\gcell{10.00}}
& \gcell{7.29}
& \gcell{9.90}
& \gcell{9.27}
& \gcell{9.23}
& \gcell{9.89}
& \gcell{9.97}
& \gcell{8.78}
& \gcell{8.24} \\
\midrule

\multirow{3}{*}{\ds{5}}
& 16k \setbase{26.11}
& \cellcolor{heatneutral}\base
& \gcell{23.05}
& \textbf{\gcell{25.96}}
& \gcell{25.17}
& \gcell{25.36}
& \gcell{24.57}
& \gcell{25.50}
& \gcell{23.51}
& \gcell{25.30}
& \gcell{23.36}
& \gcell{23.73} \\
& 32k \setbase{25.69}
& \cellcolor{heatneutral}\base
& \gcell{22.13}
& \textbf{\gcell{25.94}}
& \gcell{24.19}
& \gcell{25.09}
& \gcell{25.19}
& \textbf{\gcell{25.94}}
& \gcell{23.02}
& \gcell{24.68}
& \gcell{22.96}
& \gcell{23.45} \\
& 64k \setbase{24.34}
& \cellcolor{heatneutral}\base
& \gcell{20.42}
& \textbf{\gcell{24.81}}
& \gcell{23.24}
& \gcell{24.77}
& \gcell{23.00}
& \gcell{24.68}
& \gcell{20.57}
& \gcell{22.42}
& \gcell{20.31}
& \gcell{20.97} \\
\midrule

\multirow{3}{*}{\ds{6}}
& 16k \setbase{18.55}
& \cellcolor{heatneutral}\base
& \gcell{17.34}
& \gcell{19.87}
& \gcell{19.28}
& \gcell{19.48}
& \gcell{19.39}
& \textbf{\gcell{20.29}}
& \gcell{17.49}
& \gcell{19.30}
& \gcell{19.07}
& \gcell{17.39} \\
& 32k \setbase{19.06}
& \cellcolor{heatneutral}\base
& \gcell{16.53}
& \textbf{\gcell{18.79}}
& \gcell{18.26}
& \gcell{18.60}
& \gcell{17.93}
& \gcell{18.59}
& \gcell{16.93}
& \gcell{17.44}
& \gcell{17.60}
& \gcell{16.78} \\
& 64k \setbase{19.19}
& \cellcolor{heatneutral}\base
& \gcell{16.85}
& \textbf{\gcell{19.40}}
& \gcell{18.31}
& \gcell{17.19}
& \gcell{19.14}
& \gcell{17.37}
& \gcell{17.53}
& \gcell{18.04}
& \gcell{18.12}
& \gcell{17.17} \\
\midrule

\multirow{3}{*}{\ds{7}}
& 16k \setbase{53.80}
& \cellcolor{heatneutral}\base
& \gcell{44.96}
& \textbf{\gcell{52.94}}
& \gcell{45.93}
& \gcell{50.59}
& \gcell{52.42}
& \gcell{49.71}
& \gcell{51.22}
& \gcell{52.91}
& \gcell{51.33}
& \gcell{49.81} \\
& 32k \setbase{49.48}
& \cellcolor{heatneutral}\base
& \gcell{32.97}
& \textbf{\gcell{49.59}}
& \gcell{36.73}
& \gcell{46.50}
& \gcell{49.45}
& \gcell{41.38}
& \gcell{46.42}
& \gcell{48.23}
& \gcell{46.76}
& \gcell{45.31} \\
& 64k \setbase{48.81}
& \cellcolor{heatneutral}\base
& \gcell{30.48}
& \textbf{\gcell{48.06}}
& \gcell{40.28}
& \gcell{41.69}
& \gcell{47.24}
& \gcell{43.21}
& \gcell{45.87}
& \gcell{47.62}
& \gcell{46.05}
& \gcell{40.33} \\
\midrule

\multirow{3}{*}{\ds{8}}
& 16k \setbase{34.83}
& \cellcolor{heatneutral}\base
& \gcell{34.15}
& \textbf{\gcell{35.89}}
& \gcell{32.47}
& \gcell{32.55}
& \gcell{34.16}
& \gcell{34.46}
& \gcell{34.35}
& \gcell{35.21}
& \gcell{32.56}
& \gcell{33.72} \\
& 32k \setbase{32.85}
& \cellcolor{heatneutral}\base
& \gcell{26.47}
& \textbf{\gcell{32.96}}
& \gcell{30.52}
& \gcell{30.18}
& \gcell{31.21}
& \gcell{32.74}
& \gcell{32.64}
& \gcell{31.11}
& \gcell{30.78}
& \gcell{29.30} \\
& 64k \setbase{36.52}
& \cellcolor{heatneutral}\base
& \gcell{29.12}
& \gcell{34.55}
& \gcell{30.06}
& \textbf{\gcell{34.90}}
& \gcell{33.69}
& \gcell{34.28}
& \gcell{34.33}
& \gcell{34.10}
& \gcell{34.49}
& \gcell{30.63} \\
\midrule

\multirow{3}{*}{\ds{9}}
& 16k \setbase{19.05}
& \cellcolor{heatneutral}\base
& \gcell{13.67}
& \textbf{\gcell{20.32}}
& \gcell{20.16}
& \gcell{13.08}
& \gcell{19.91}
& \gcell{14.35}
& \gcell{17.77}
& \gcell{17.47}
& \gcell{19.65}
& \gcell{15.66} \\
& 32k \setbase{13.80}
& \cellcolor{heatneutral}\base
& \gcell{7.63}
& \textbf{\gcell{12.48}}
& \gcell{12.41}
& \gcell{11.14}
& \gcell{12.03}
& \gcell{11.41}
& \gcell{11.08}
& \gcell{11.86}
& \gcell{11.99}
& \gcell{10.90} \\
& 64k \setbase{11.64}
& \cellcolor{heatneutral}\base
& \gcell{8.53}
& \textbf{\gcell{13.20}}
& \gcell{12.08}
& \gcell{9.98}
& \gcell{10.41}
& \gcell{10.44}
& \gcell{10.08}
& \gcell{11.67}
& \gcell{10.24}
& \gcell{9.86} \\

\bottomrule
\end{tabular}
\end{adjustbox}

\begin{minipage}{\textwidth}

    \centering

    \begin{minipage}[b]{0.497\linewidth}
        \centering
        \includegraphics[width=\linewidth]{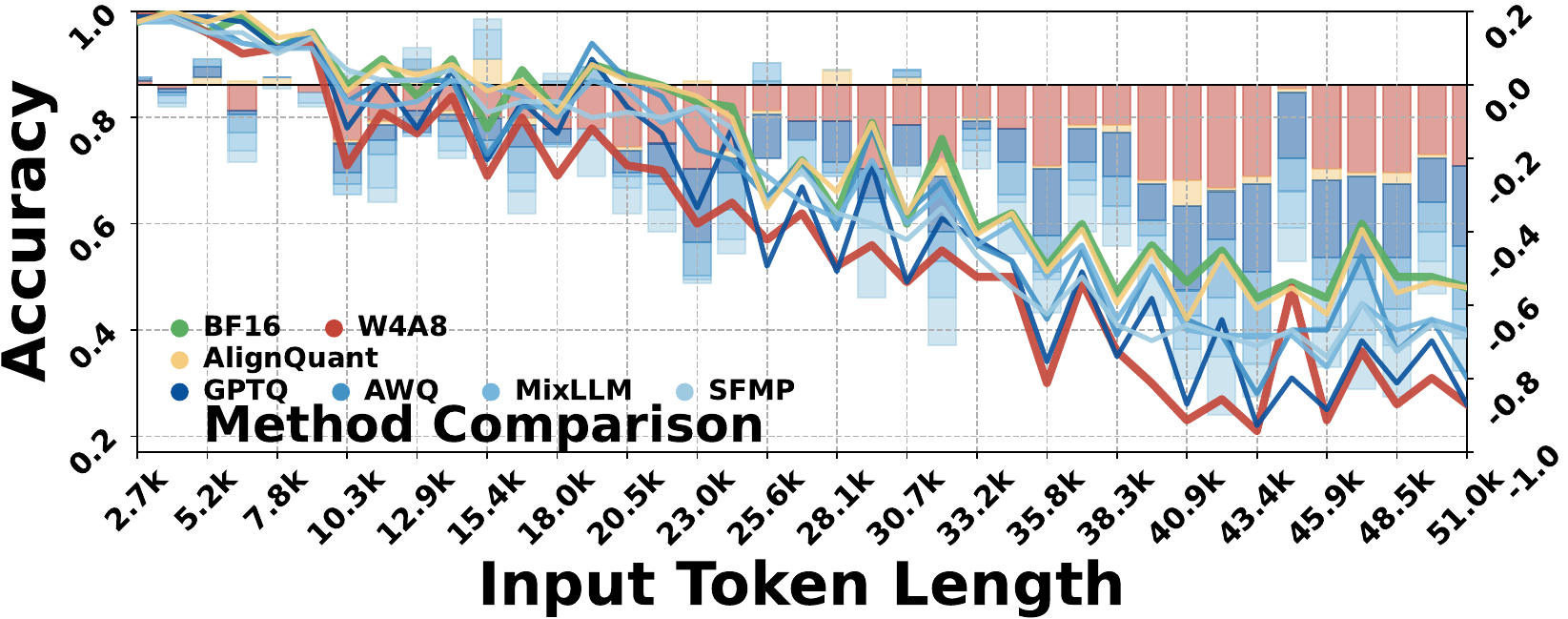}
    \end{minipage}
    \hfill
    \begin{minipage}[b]{0.497\linewidth}
        \centering
        \includegraphics[width=\linewidth]{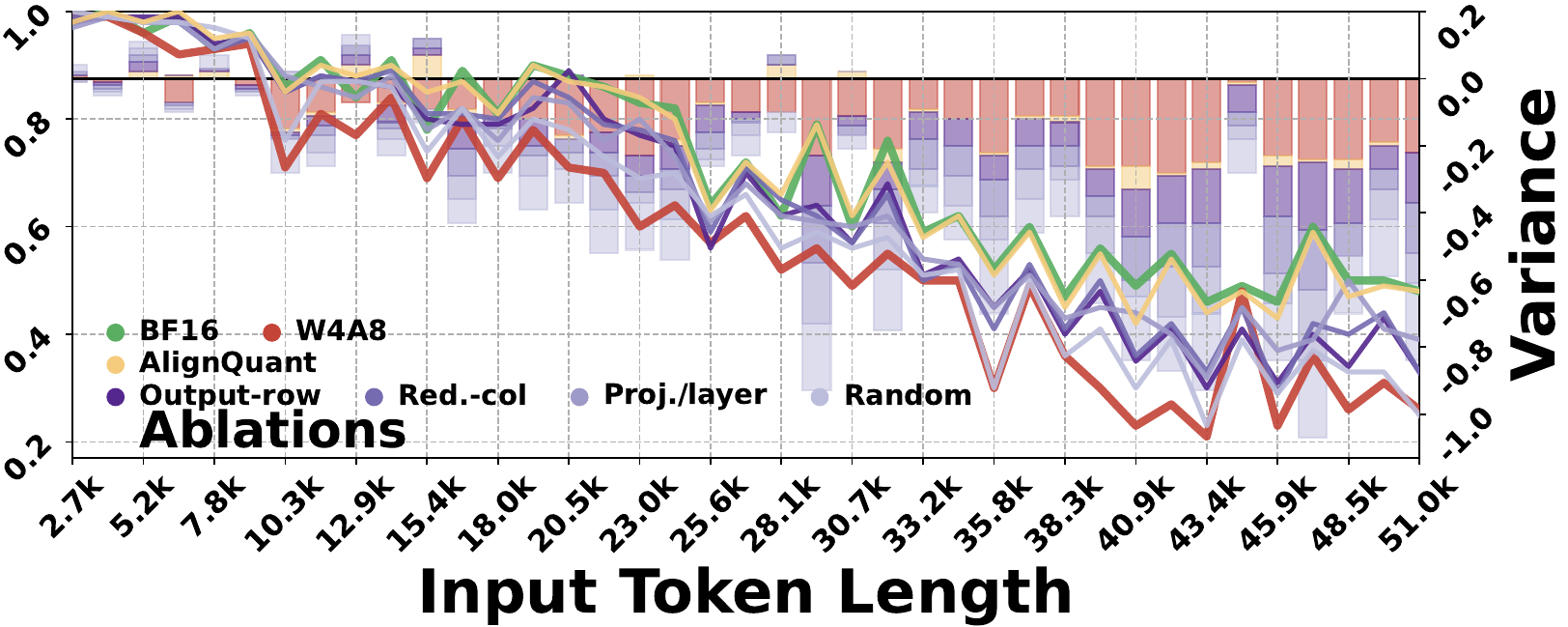}
    \end{minipage}

    \vspace{-5pt}
    \captionof{figure}{
    LongEval line-retrieval accuracy for \textbf{Llama-3.2-3B-Instruct} across input lengths from 2.7K to 51.0K tokens.
    Lines report exact-match accuracy and bars show differences relative to BF16.
    \textbf{Left:} \modelname compared with baselines.
    \textbf{Right:} \modelname compared with ablations.
    }
    \vspace{-15pt}
    \label{fig:longeval}
\end{minipage}
\end{table*}

\vspace{-5pt}
\subsection{Long-Context Evaluation}
\label{sec:long-context}
\vspace{-5pt}

Beyond general benchmarks, we evaluate long-context information retrieval and reasoning at LV-Eval's 16K, 32K, and 64K length levels.
Using the dataset indices in Table~\ref{tab:lveval_llama3b}: \ds{1} \textit{cmrc\_mixup} (Chinese reading comprehension); \ds{2} \textit{dureader\_mixup} (Chinese QA); \ds{3} \textit{hotpotwikiqa\_mixup} (multi-hop Wikipedia evidence retrieval); \ds{4} \textit{lic\_mixup} (long-input comprehension and localization); \ds{5}--\ds{7} \textit{loogle\_CR/MIR/SD\_mixup} (comprehension/reasoning, multiple-information retrieval, and short-dependency QA); and \ds{8}--\ds{9} \textit{multifieldqa\_en/zh\_mixup} (English and Chinese multi-field retrieval).

LooGLE-SD (\ds{7}) shows how selective precision behaves as context grows (Table~\ref{tab:lveval_llama3b}).
At the 64K level, \modelname scores 48.06 alongside BF16's 48.81, versus 47.24 for MixLLM and 41.69 for AWQ.
Its margin over AWQ grows from 2.35 to 6.37 points, showing that \textbf{precision retained in sensitive regions remains useful at longer contexts}.

\textbf{Allocation at long contexts.}
Comprehension and reasoning further distinguish precision placement: at the 64K level, \modelname scores 24.81 on LooGLE-CR (\ds{5}), versus 22.42 for reduction-column and 20.57 for output-row allocation.
This supports \textbf{local precision choices within output channels} (Appendix~\ref{app:lveval_results}).

\textbf{Retrieval across input lengths.}
LongEval resolves length effects across 39 settings from 2.7K to 51.0K tokens (Figure~\ref{fig:longeval}).
Short inputs are similar, while longer inputs separate the methods: at the longest input, \modelname matches BF16 at 48\%, versus 40\% for MixLLM and 31\% for AWQ, preserving successful retrieval at the demanding end of the sweep.

\textbf{Retrieval allocation controls.}
At the longest inputs, \modelname achieves 48\% accuracy, outperforming structured (33\%--39\%) and random (25\%) placements. This demonstrates the value of \textbf{preserving precision within targeted weight regions} to retain information as context grows.

Across these evaluations, \modelname combines \textbf{quality preservation and faster generation in one compact model}.
Loss-aware tile scoring identifies the weight regions worth protecting, while tile-aligned storage and integer execution make those selected weights efficient to use.
The common tile structure connects these numerical and computational choices throughout deployment.

\vspace{-10pt}
\section{Conclusion}
\label{sec:conclusion}
\vspace{-10pt}

\modelname demonstrates that fine-grained precision choices can coexist with regular GPU computation when the precision unit supports both sensitivity evaluation and efficient execution. 
Our results show this preserves quality while boosting inference efficiency, suggesting a broader principle for future quantizers: design precision units for compact storage and regular execution, then allocate bits based on local output sensitivity.

\clearpage

\subsection*{AI Use Statement}

Generative AI tools were used to assist with manuscript polishing and code development.
The authors reviewed all AI-assisted content and take responsibility for the final work.

\subsection*{Reproducibility Statement}

Appendices~\ref{app:numerical_construction} through~\ref{app:packed_execution} detail the numerical construction, calibration procedure, allocation derivations, and GPU implementation.
Appendix~\ref{app:experimental-details} specifies the experimental configurations, evaluation protocols, and storage and runtime measurements.

\bibliography{reference}
\bibliographystyle{iclr2027_conference}


\clearpage
\appendix
\setlength{\textfloatsep}{8pt}
\setlength{\dbltextfloatsep}{8pt}

\section{Notation and Reading Guide}
\label{app:notation}

Table~\ref{tab:notation} collects the notation used throughout the method.
Appendix~\ref{app:numerical_construction} defines the quantized candidates, Appendix~\ref{app:calibration_algorithm} describes calibration, and Appendix~\ref{app:score_derivations} derives the allocation rule.
Storage and execution details appear in Appendix~\ref{app:packed_execution}, followed by the design discussion in Appendix~\ref{app:design_rationale}.

\begin{table*}[t]
\centering
\caption{Notation for the numerical construction, allocation, and execution of quantized weight tiles.}
\small
\setlength{\tabcolsep}{4pt}
\renewcommand{\arraystretch}{1.08}
\begin{tabular}{c l l}
\toprule
\textbf{Category} & \textbf{Symbol} & \textbf{Description} \\
\midrule
\multirow{4}{*}{\textbf{Projection}}
& $X\in\mathbb R^{M\times K}$ & Input activations \\
& $W\in\mathbb R^{N\times K}$ & Projection weights \\
& $Y=XW^\top$ & Projection output \\
& $M,K,N$ & Activation rows, input features, and output features \\
\midrule
\multirow{2}{*}{\textbf{Coordinates}}
& $U,V$ & Orthogonal output- and input-feature transforms \\
& $\widetilde{\cdot},\widehat{\cdot}$ & Transformed coordinates and quantized reconstructions \\
\midrule
\multirow{4}{*}{\textbf{Tiling}}
& $b_N,b_K$ & Weight-tile dimensions \\
& $R=N/b_N,\ C=K/b_K$ & Tile-row and tile-column counts per projection \\
& $\mathcal T,\ t$ & Model-wide tile set and one tile identifier \\
& $\rho,\ z_t$ & Target W4 fraction and selected precision $z_t\in\{4,8\}$ \\
\midrule
\multirow{3}{*}{\textbf{Calibration}}
& $p\in\{\mathrm{pre},\mathrm{dec}\}$ & Inference phase \\
& $\mathcal C_p,\ c$ & Phase-specific observations and one observation \\
& $\mathcal L_c,\ n_c$ & Mean prediction loss and its valid token count \\
\midrule
\multirow{4}{*}{\textbf{Scoring}}
& $\widehat W_{4,t},\widehat W_{8,t}$ & Reconstructed weight candidates \\
& $\widehat X_t^{p,c}$ & Reconstructed A8 input slice associated with tile $t$ \\
& $\widetilde Y_t,\Delta\widetilde Y_t^{p,c}$ & Output slice and its candidate-induced change \\
& $F_t^p,\ F_t$ & Phase-specific score and joint allocation score \\
\midrule
\multirow{3}{*}{\textbf{Storage}}
& $Q_{b,t},s_{b,t};\ Q_t,s_t$ & Candidate and selected integer values and scales \\
& $\beta_{r,j}$ & Precision label: 1 for W4 and 0 for W8 \\
& $T_4,T_8$ & Selected W4 and W8 tile counts per projection \\
\bottomrule
\end{tabular}

\label{tab:notation}
\end{table*}

Tiles in one output-tile row share output features but span different input columns.
Their scores pair the shared output-gradient slice with each tile's input-dependent perturbation.

\section{Numerical Construction of Quantized Tiles}
\label{app:numerical_construction}

\subsection{Tile-Aligned Transformations}
\label{app:tile_transforms}

Block-diagonal transforms $U=\operatorname{diag}(U_0,\ldots,U_{R-1})$ and $V=\operatorname{diag}(V_0,\ldots,V_{C-1})$ redistribute values \textbf{within each weight tile} \citep{ashkboos2024quarot,liu2025spinquant}.
The transformed tile is $\widetilde W_{r,j}=U_rW_{r,j}V_j$.
With $\widetilde X=XV$ and $\widetilde W=UWV$, output reconstruction preserves the floating-point projection:
\begin{equation}
\widetilde YU=(XV)(UWV)^\top U=XVV^\top W^\top U^\top U=XW^\top.
\label{eq:app_transform_identity}
\end{equation}

For the default $64\times64$ configuration, stored seeds define $V_j=\Phi_{v_j}$ and $U_r=\Phi_{u_r}^{\top}$, with $\Phi_{-1}=I$, $\Phi_0=H_{64}$, and $\Phi_s=D_{L,s}P_sH_{64}D_{R,s}$ for $s=1,\ldots,6$.
Here, $P_s$ is a permutation, $D_{L,s}$ and $D_{R,s}$ are diagonal sign matrices, and $H_{64}$ is the normalized Hadamard transform with final butterfly scaling $1/8$.

\subsection{Weight Candidates}
\label{app:quantization_details}

For each precision $b\in\{4,8\}$, let $q_{\min,b}=-2^{b-1}$ and $q_{\max,b}=2^{b-1}-1$.
Each tile uses zero point zero and nearest-integer rounding with ties to even:
\begin{equation}
Q_b(A;s)=\operatorname{clip}_{[q_{\min,b},q_{\max,b}]}\!\left(\operatorname{round}_{\mathrm{even}}(A/s)\right).
\label{eq:app_weight_quantizer}
\end{equation}
The initial scale covers both tile endpoints:
\begin{equation}
s_{0,b}(A)=\max\left\{\frac{\max(0,\max A)}{q_{\max,b}},\frac{\max(0,-\min A)}{|q_{\min,b}|},10^{-12}\right\}.
\label{eq:app_weight_base_scale}
\end{equation}
We fit \textbf{a separate scale at each precision} by testing $s_i=\max\{(1/4+i/128)s_{0,b}(A),10^{-12}\}$ for $i=0,\ldots,128$ and selecting
\begin{equation}
s_b^\star(A)=\underset{s_i}{\arg\min}\ \|A-s_iQ_b(A;s_i)\|_F^2.
\label{eq:app_weight_scale_search}
\end{equation}
Search and integer construction use FP64, with grid order resolving equal errors.
The candidate contains $Q_{b,t}=Q_b(\widetilde W_t;s_b^\star)$ and $s_{b,t}=\operatorname{FP32}(s_b^\star)$.
Scoring compares their reconstructions $\widehat W_{b,t}=s_{b,t}Q_{b,t}$, and packing retains the selected pair $(Q_t,s_t)=(Q_{z_t,t},s_{z_t,t})$.

\subsection{Activation Quantization and Padding}
\label{app:a8_quantization}

After applying $V$, activation quantization uses one scale per group of 64 features and $b_M$ rows, where $b_M=1$ for $M=1$ and $b_M=16$ otherwise.
Each group $A$ uses $s_A^\star=s_{0,8}(A)$ from \eqref{eq:app_weight_base_scale} and $Q_A=Q_8(A;s_A^\star)$.
CUDA preparation computes in FP64 and stores signed INT8 values with FP32 scales; calibration follows the same grouping and endpoint rule with FP64 reconstruction.
Zero padding completes $M'=\lceil M/b_M\rceil b_M$ rows, and the output is sliced back to $M$ rows.
The prediction mask selects the tokens contributing to calibration loss.

\subsection{Transform Selection}
\label{app:transform_selection}

For each compatible input-sharing group $\mathcal G$, transform selection uses collected inputs $X_{\mathcal G}$, a shared $V$, and projection-specific $U_\ell$.
Let $\mathcal R_4$ denote 4-bit quantization and reconstruction with FP64 fitted scales, using $16\times64$ activation groups and $64\times64$ weight tiles.
The search minimizes
\begin{equation}
E_{\mathcal G}=\sum_{\ell\in\mathcal G}\left\|\mathcal R_4(X_{\mathcal G}V)\mathcal R_4(U_\ell W_\ell V)^\top U_\ell-X_{\mathcal G}W_\ell^\top\right\|_F^2.
\label{eq:app_transform_objective}
\end{equation}
Search over the eight seeds initializes a uniform input transform and then a uniform output transform for each projection.
Two coordinate sweeps refine input groups followed by output groups, with seed order resolving ties.
The resulting transforms feed W4 and W8 candidate scoring under A8 inputs.

\section{Calibration Procedure}
\label{app:calibration_algorithm}

\subsection{Forward Values and Gradient Observations}
\label{app:calibration_details}

Calibration uses reconstructed 8-bit weights, the selected transforms, and 8-bit activations while keeping model parameters fixed.
Outputs return to original coordinates and the model's activation type.
For a captured output gradient $G$, the frozen-weight straight-through rule computes $G_X=GW_{\mathrm{original}}$ using the pretrained weights.
Since $Y=\widetilde YU$, scoring pairs the reconstructed input slices with gradients in the same transformed output coordinates:
\begin{equation}
\widetilde G=GU^\top.
\label{eq:app_gradient_transform}
\end{equation}

\subsection{Prefill and Decode Observations}
\label{app:phase_calibration}

Each record contains a prompt and continuation, with loss $\mathcal L_c=n_c^{-1}\sum_{\tau\in\mathcal I_c}\ell_{c,\tau}$ over $n_c=|\mathcal I_c|$ prediction targets.
Prefill scores \textbf{all continuation targets} in one causal pass, beginning with the target predicted from the final prompt position.

Decode builds the prompt's key/value cache and feeds continuation tokens sequentially to predict the next token.
Up to four deterministically spaced steps are scored, each with a \textbf{one-token loss} ($n_c=1$) and one-row projection observations.
The accumulated cache supplies fixed context for each scored step.

\subsection{Computing the Precision Map}
\label{app:allocation_algorithm}

For each input-tile column, the scorer evaluates equations~\ref{eq:output-perturbation} and~\ref{eq:fisher-score} using FP64 inputs, reconstructed candidates, and accumulation.
This produces one tile-score matrix per projection and phase.
\Eqref{eq:joint-score} combines the scores for model-wide ranking.
The allocator selects the target number of 4-bit tiles and packs each tile's selected integer values and scale, following the ordering rules in Appendix~\ref{app:allocation_details}.

\section{Derivations and Properties of the Allocation Score}
\label{app:score_derivations}

\subsection{Local Output Perturbation}
\label{app:perturbation_derivation}

A tile replaces several weights at once, and their errors combine through the input.
For output group $r$, the reconstructed projection is $\widehat{\widetilde Y}_{:,r}=\sum_j\widehat X_{:,j}\widehat W_{r,j}^{\top}$.
Holding the inputs and other tiles fixed, the W8-to-W4 substitution at $t=(r,j)$ gives
\begin{equation}
\Delta\widetilde Y_t^{p,c}=\widehat X_t^{p,c}\widehat W_{4,t}^{\top}-\widehat X_t^{p,c}\widehat W_{8,t}^{\top}=\widehat X_t^{p,c}(\widehat W_{4,t}-\widehat W_{8,t})^\top.
\label{eq:app_local_difference}
\end{equation}

For one output coordinate, write the input as $x$ and the candidate weight difference as $\delta w$.
Then
\begin{equation}
(x^\top\delta w)^2=\sum_i x_i^2\delta w_i^2+2\sum_{i<i'}x_ix_{i'}\delta w_i\delta w_{i'}.
\label{eq:app_within_tile_interactions}
\end{equation}
The cross terms capture reinforcement and cancellation between errors contributing to that output.

\subsection{Loss-Weighted Quadratic Form}
\label{app:fisher_derivation}

The factor $n_c$ converts the gradient of the mean token loss to the gradient of its sum:
\begin{equation}
\nabla_{\widetilde Y_t}\sum_{\tau\in\mathcal I_c}\ell_{c,\tau}=n_c\nabla_{\widetilde Y_t}\mathcal L_c.
\label{eq:app_loss_scale}
\end{equation}
For each observation, let $d_t^{p,c}=\operatorname{vec}(\Delta\widetilde Y_t^{p,c})$ and $g_t^{p,c}=\operatorname{vec}(n_c\nabla_{\widetilde Y_t}\mathcal L_c)$.
Equivalently, \eqref{eq:fisher-score} becomes
\begin{equation}
F_t^p=\frac12\sum_{c\in\mathcal C_p}(d_t^{p,c})^\top\operatorname{diag}(g_t^{p,c}\odot g_t^{p,c})d_t^{p,c}.
\label{eq:app_fisher_quadratic}
\end{equation}
The diagonal weights act on projection-output coordinates, after the weight errors within each tile have combined through the input.
This gives a loss-weighted score for the whole tile substitution.
Scores use the captured calibration state and the gradient rule in Appendix~\ref{app:calibration_details}; downstream quality is measured on the assembled model with all selected substitutions applied.

\subsection{Phase Normalization}
\label{app:phase_aggregation}

Assume $F_t^p\ge0$ and a positive total score in each phase.
The normalized scores $\overline F_t^p=F_t^p/\sum_{t'}F_{t'}^p$ sum to one.
For any phase-wide multiplier $\alpha_p>0$,
\begin{equation}
\frac{\alpha_pF_t^p}{\sum_{t'}\alpha_pF_{t'}^p}=\overline F_t^p.
\label{eq:app_phase_scale_invariance}
\end{equation}
Normalization therefore retains relative tile importance when the overall score scale changes.

The joint score $F_t=\max\{\overline F_t^{\mathrm{pre}},\overline F_t^{\mathrm{dec}}\}$ retains each tile's stronger relative importance.
For any set $\mathcal S$ of tiles assigned lower precision, $\max_p\sum_{t\in\mathcal S}\overline F_t^p\leq\sum_{t\in\mathcal S}F_t$.
Fixed-budget ranking therefore minimizes \textbf{a common upper bound on both phases' total normalized scores}.
The budget then determines which positions in this joint ranking receive W8.

\subsection{Fixed-Budget Selection}
\label{app:allocation_details}

Let $k=\operatorname{round}(\rho|\mathcal T|)$ and keep the candidate scores fixed.
Equal-area tiles give every W8-to-W4 substitution the same integer-storage saving, so allocation selects
\begin{equation}
\underset{\mathcal S\subseteq\mathcal T,\;|\mathcal S|=k}{\arg\min}\ \sum_{t\in\mathcal S}F_t.
\label{eq:app_fixed_budget_objective}
\end{equation}

Order the scores as $F_{t_1}\le\cdots\le F_{t_{|\mathcal T|}}$.
For any feasible set with sorted ranks $j_1<\cdots<j_k$, each $j_i\ge i$, giving
\[
\sum_{i=1}^kF_{t_{j_i}}\ge\sum_{i=1}^kF_{t_i}.
\]
The first $k$ tiles attain this minimum.
The endpoints $k=0$ and $k=|\mathcal T|$ select the empty set and the full tile set, respectively.

The implementation resolves equal scores by lexicographic order of the module-and-coordinate identifier.
The tile count uses nearest-integer rounding with ties to even.

\section{Packed Representation and GPU Execution}
\label{app:packed_execution}

\subsection{Locating and Unpacking a Tile}
\label{app:packing_details}

The kernel starts from a logical tile coordinate $(r,j)$.
Its precision label $\beta_{r,j}=\mathbf 1[z_{r,j}=4]$ selects the W4 or W8 buffer.
Labels follow row-major tile order, with eight labels per byte and the first label in the least significant bit.

Each buffer stores its selected tiles in row order.
For precision $b\in\{4,8\}$, the physical tile index is
\begin{equation}
\operatorname{index}_b(r,j)=\underbrace{\sum_{r'<r}\sum_{j'=0}^{C-1}\mathbf 1[z_{r',j'}=b]}_{\text{stored row offset}}+\underbrace{\sum_{j'<j}\mathbf 1[z_{r,j'}=b]}_{\text{position within the row}}.
\label{eq:app_buffer_index}
\end{equation}
The weight scale uses the same index.
For example, a row with precisions W4, W8, W4 stores its first and third tiles consecutively in the W4 buffer.

Within each tile, values are viewed in input-by-output order and permuted into the CUTLASS \texttt{ColumnMajorTensorOpMultiplicandCrosswise<8,32>} layout used by the integer kernels.
Both precisions use this element order.
W8 stores one signed value per byte, while W4 packs consecutive values $q_0,q_1$ as
\begin{equation}
\operatorname{byte}(q_0,q_1)=(q_0\bmod16)+16(q_1\bmod16).
\label{eq:app_nibble_pack}
\end{equation}
A nibble $v$ expands to $v$ for $v<8$ and to $v-16$ otherwise.
This sign extension preserves the selected integer value and its reconstruction with the stored scale.

\subsection{Stored Bytes}
\label{app:storage_accounting}

Let $T_4+T_8=RC$ and $B=b_Nb_K$.
The integer weights occupy $BT_4/2+BT_8$ bytes.
The packed projection additionally contains one FP32 scale per tile, the precision labels, and two INT32 row-offset arrays of length $R+1$:
\begin{equation}
S_{\mathrm{packed}}=\frac{B}{2}T_4+BT_8+4(T_4+T_8)+\left\lceil\frac{RC}{8}\right\rceil+8(R+1).
\label{eq:app_packed_bytes}
\end{equation}
For $64\times64$ tiles, the W4 and W8 buffers have shapes $[T_4,64,32]$ and $[T_8,64,64]$, respectively.

With realized W4 fraction $\rho_{\mathrm{real}}=T_4/(T_4+T_8)$, the integer-weight cost is $8-4\rho_{\mathrm{real}}$ bits per weight.
Complete packed bits per weight are $8S_{\mathrm{packed}}/[B(T_4+T_8)]$.
Transform seeds and matrix geometry accompany the serialized tensors.

Runtime allocations also include prepared activations, outputs, and execution workspace.
For a mixed projection using long-prefill preparation, the temporary INT8 weights, scales, and row offsets occupy $BRC+4RC+4(R+1)$ bytes for that call.

\subsection{Operand Preparation and Scheduling}
\label{app:execution_details}

\begin{figure*}[!htb]
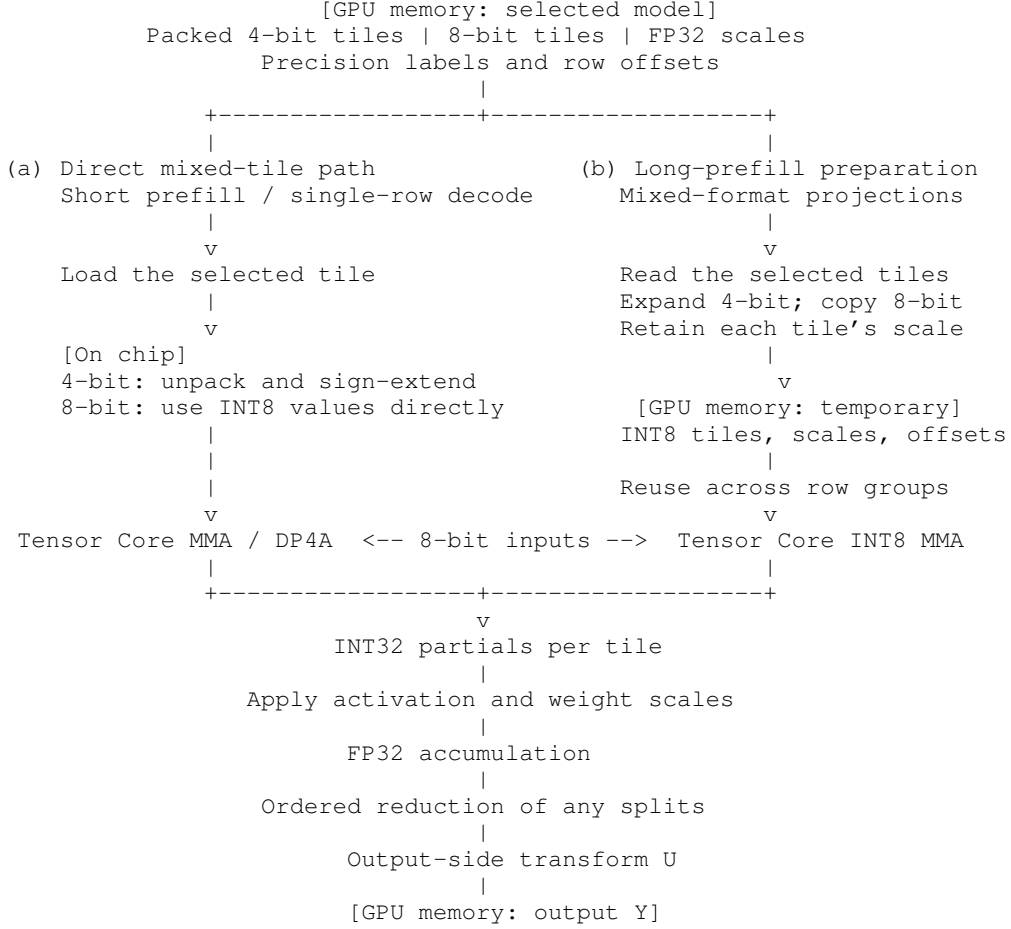

\centering
\begin{adjustbox}{max width=\textwidth,center}
\begin{BVerbatim}[fontsize=\small]
                      [GPU memory: selected model]
          Packed 4-bit tiles | 8-bit tiles | FP32 scales
                  Precision labels and row offsets
                                 |
              +------------------+-------------------+
              |                                      |
(a) Direct mixed-tile path              (b) Long-prefill preparation
    Short prefill / single-row decode      Mixed-format projections
              |                                      |
              v                                      v
    Load the selected tile                 Read the selected tiles
              |                            Expand 4-bit; copy 8-bit
              v                            Retain each tile's scale
    [On chip]                                        |
    4-bit: unpack and sign-extend                     v
    8-bit: use INT8 values directly         [GPU memory: temporary]
              |                            INT8 tiles, scales, offsets
              |                                      |
              |                            Reuse across row groups
              v                                      v
 Tensor Core MMA / DP4A  <-- 8-bit inputs -->  Tensor Core INT8 MMA
              |                                      |
              +------------------+-------------------+
                                 v
                       INT32 partials per tile
                                 |
                 Apply activation and weight scales
                                 |
                        FP32 accumulation
                                 |
                  Ordered reduction of any splits
                                 |
                        Output-side transform U
                                 |
                        [GPU memory: output Y]
\end{BVerbatim}
\end{adjustbox}
\caption{
\textbf{Weight preparation and data movement.}
Both paths start from the same selected mixed-precision model.
Direct kernels expand packed 4-bit weights on chip, while long prefill prepares a temporary INT8 buffer once per projection call and reuses it across activation-row groups.
Both paths retain the selected integer values and scales and receive separately prepared 8-bit activations.
Integer partials are rescaled before FP32 accumulation and output reconstruction.
}
\label{fig:weight_dataflow}
\end{figure*}

Figure~\ref{fig:weight_dataflow} shows the two weight-preparation paths, both supplied with activations transformed by $V$ and quantized to 8 bits.
Table~\ref{tab:phase_kernel_geometry} lists the prefill and decode thread-block geometries.

\begin{table*}[t]
\centering
\caption{Computation assigned to one thread block before optional splitting of the input dimension.}
\small
\setlength{\tabcolsep}{7pt}
\begin{tabular}{l c c l}
\toprule
\textbf{Schedule} & \textbf{Output region} & \textbf{Threads} & \textbf{Arithmetic} \\
\midrule
Prefill M16 & $16\times64$ & 128 & Tensor Core INT8 MMA \\
Prefill M64 & $64\times64$ & 256 & Tensor Core INT8 MMA \\
Single-sequence decode & $1\times64$ & 256 & DP4A dot products \\
\bottomrule
\end{tabular}
\label{tab:phase_kernel_geometry}
\end{table*}

\textbf{Prefill.}
Prefill selects M16 below 128 padded rows and M64 otherwise, using $16\times8\times32$ INT8 MMA instructions.
The M64 path receives activation tiles of shape $[M'/16,C,16,64]$, with one scale per 16-row group.
Two shared-memory stages overlap tile loading and multiplication.

Direct kernels prepare selected weights on chip.
For mixed-format projections with $M'\ge1024$, a temporary INT8 buffer is prepared once per call, retaining the selected values and scales in logical tile order with its own row offsets.
Reuse across activation-row groups amortizes this preparation over the prompt.

\textbf{Decode.}
For $M=1$, the kernel directly loads packed tiles, expands 4-bit values on chip, and distributes work across output features and input-column groups.
DP4A accumulates four signed 8-bit products per instruction into INT32.
In split execution, four adjacent lanes cooperate on one output feature, each handling 16 input coordinates per tile.

\textbf{Accumulation and reconstruction.}
For activation-row group $m$ and output-tile row $r$, integer tile products are rescaled before FP32 accumulation:
\begin{equation}
\widehat{\widetilde Y}_{m,r}=\sum_{j=0}^{C-1}s^X_{m,j}s^W_{r,j}\left(Q^X_{m,j}(Q^W_{r,j})^\top\right).
\label{eq:app_quantized_runtime}
\end{equation}
Each parenthesized product accumulates in INT32.
For $S$ splits, split $s$ covers input-tile columns $\lfloor Cs/S\rfloor$ through $\lfloor C(s+1)/S\rfloor-1$.
FP32 partials are summed in increasing split order before $U_r$ restores output coordinates.
The interface returns the original row count and model activation type, adding bias when present.

\subsection{Shared Projection Inputs}
\label{app:shared_activation}

\begin{figure*}[!htbp]
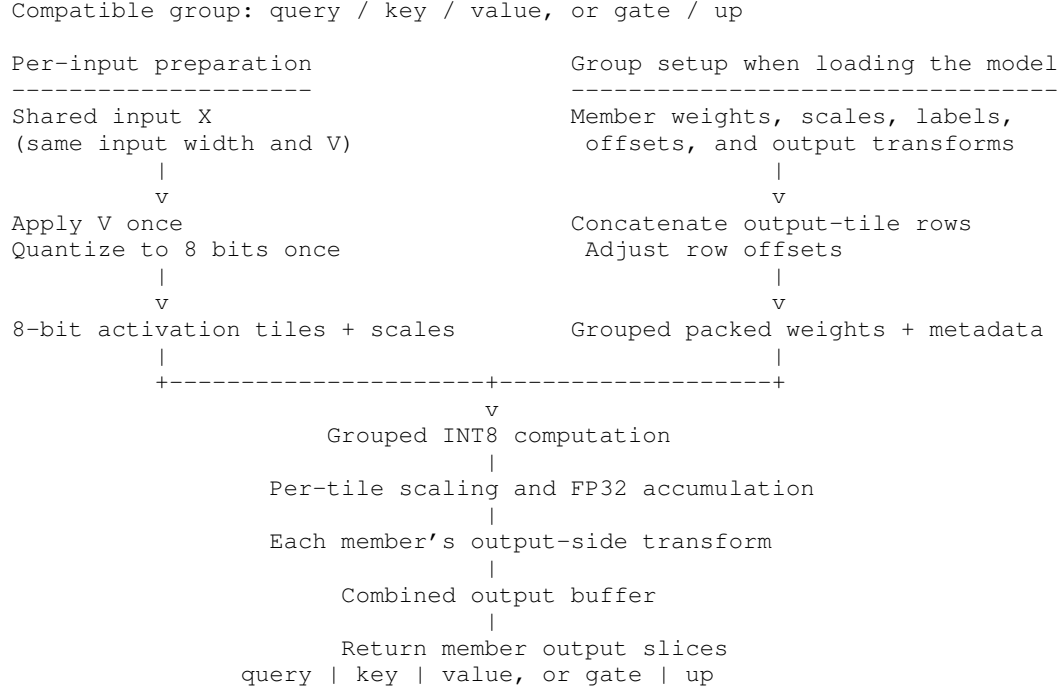

\centering
\begin{adjustbox}{max width=\textwidth,center}
\begin{BVerbatim}[fontsize=\small]
Compatible group: query / key / value, or gate / up

Per-input preparation                  Group setup when loading the model
---------------------                  ----------------------------------
Shared input X                         Member weights, scales, labels,
(same input width and V)                offsets, and output transforms
          |                                          |
          v                                          v
Apply V once                           Concatenate output-tile rows
Quantize to 8 bits once                 Adjust row offsets
          |                                          |
          v                                          v
8-bit activation tiles + scales        Grouped packed weights + metadata
          |                                          |
          +----------------------+-------------------+
                                 v
                      Grouped INT8 computation
                                 |
                  Per-tile scaling and FP32 accumulation
                                 |
                  Each member's output-side transform
                                 |
                       Combined output buffer
                                 |
                       Return member output slices
                query | key | value, or gate | up
\end{BVerbatim}
\end{adjustbox}
\caption{
\textbf{Shared input preparation across compatible projections.}
A group contains query, key, and value projections, or gate and up projections, sharing the same input, input width, and transform $V$.
Their selected weights are assembled along the output dimension during model setup.
Each input is transformed and quantized once for grouped execution, which preserves each member's weight scales and output transform.
The resulting combined output supplies the individual projection slices.
}
\label{fig:shared_input_reuse}
\end{figure*}

Compatible query, key, and value projections, or gate and up projections, share the same input, input width, and input-side transform $V$.
During model setup, their packed weight rows and output-transform metadata are concatenated, with row offsets adjusted to locate each member's tiles.
As shown in Figure~\ref{fig:shared_input_reuse}, each input is transformed and quantized once, and grouped execution computes a combined output.
Each projection retains its selected weights, scales, and output transform and receives its corresponding output slice.
The combined result is retained until all group members have consumed their slices.

\section{Design Rationale}
\label{app:design_rationale}

\subsection{Activation Precision}
\label{app:activation_precision}

One A8 input group contributes to several output tiles, each of which can independently select W4 or W8.
Using the same reconstructed input makes the candidate comparison specific to the weight substitution and lets compatible projections share activation preparation.
A8 also supplies the operand width used by both Tensor Core INT8 multiplication and DP4A \citep{xiao2023smoothquant,lin2025qserve}.

For a fixed represented range, A8 provides 256 integer levels and A4 provides 16.
An A4 variant would recalibrate the candidate scores using its reconstructed inputs and prepare those values for the chosen arithmetic.
Mixed activation precision would add format choices for input groups and coordinate them across their consuming projections.
Uniform A8 keeps the input representation shared while the weight allocation follows local sensitivity.

\subsection{Precision Granularity and Execution}
\label{app:tile_geometry_discussion}

The weight tile determines which values share a precision choice, while activation groups and thread blocks determine how those values are used.
A $64\times64$ weight tile can therefore support a single decode row or multiple prefill rows.
Smaller weight tiles provide more allocation choices and increase scale and label storage, whose contribution is approximately $33/(b_Nb_K)$ bits per weight before row offsets and byte alignment.

Joint calibration assigns the shared weights using evidence from both phases.
At runtime, direct loading forms INT8 operands on chip, while long prefill reuses prepared operands across many rows.
These schedules preserve the selected values and scales while adapting their preparation to the available reuse.

\section{Experimental Protocols}
\label{app:experimental-details}

\subsection{Models and Comparison Configurations}
\label{app:comparison_configs}

Evaluated models are listed in Section~\ref{sec:setup}.
General quality and LV-Eval compare GPTQ and AWQ on all four models, adding MixLLM and SFMP on Llama-3.2-3B and the two Qwen models.
The reported LongEval results use Llama-3.2-3B with the comparison set in Figure~\ref{fig:longeval}.
Cross-model A100 measurements include all four external methods on every model.

GPTQ and AWQ use 4-bit weights with group size 128.
MixLLM uses \texttt{kernels\_mixllm::gemm}; the evaluated SFMP implementation unpacks and dequantizes INT4 weights for floating-point multiplication.
Reported SFMP system measurements characterize this evaluated dequantization-based backend.
The four allocation controls share the 60\% 4-bit tile target.
Output-row shares precision across reduction-direction tiles within an output-feature group, while reduction-column shares precision across output groups within a reduction-tile column.
At the default tile size, these groups span 64 output features for Output-row and 64 input features for Red.-col.
Projection/layer-constrained allocation restricts the budget within projection/layer partitions, while random allocation selects the same number of 4-bit tiles using a sensitivity-independent random ranking.
The endpoints use uniform 4-bit or 8-bit weights with 8-bit activations.

Exact checkpoint revisions are pinned in the released model presets.
\modelname quantizes the decoder query, key, value, and attention-output projections together with the feed-forward gate, up, and down projections; other modules retain their original representation.

\subsection{Calibration Settings}
\label{app:calibration_evaluation}

Calibration uses 64 ordered document-summary pairs from Multi-News training rows 544--607 \citep{fabbri2019multi}.
The first 24 records provide prefill observations.
Prompt text is truncated to 4096 tokens, and summary text to 128 tokens before the end-of-sequence token is appended.
Decode uses 24 deterministic matched-prefix windows constructed by the released calibration script, each with 32 continuation tokens, and scores up to four deterministic decode steps per window.
The default record order and observation positions are selected deterministically.
Candidate construction and scoring follow Appendices~\ref{app:numerical_construction} and~\ref{app:calibration_algorithm}, and the default model-wide allocation assigns 4-bit weights to the lowest-scoring 60\% of tiles.

\subsection{General Quality Evaluation}
\label{app:general_protocol}

WikiText-2 reports word-level perplexity \citep{merity2016pointer}.
General Avg is the equal-weight mean accuracy on HellaSwag, MMLU, ARC-Challenge, LAMBADA, PIQA, and WinoGrande \citep{zellers2019hellaswag,hendrycks2020measuring,clark2018think,paperno2016lambada,bisk2020piqa,sakaguchi2021winogrande}.
For each model, all methods use the same evaluator, tokenizer, task configuration, prompt format, evaluation examples, and reported metric field.

\subsection{Long-Context Evaluation}
\label{app:long_context_protocol}

LongEval \citep{li2023long} measures exact-match retrieval at line counts $200,300,\ldots,4000$, with token-length axes determined by each model's tokenizer.
LV-Eval \citep{yuan2024lv} uses the nine datasets identified in Section~\ref{sec:long-context} at its 16K, 32K, and 64K length levels, retaining the official task scores.
Qwen3-14B uses the adopted YaRN configuration at 32K and 64K.

\subsection{System Evaluation}
\label{app:system_protocols}

Hardware settings follow Section~\ref{sec:setup}, with A100 runs conducted at Polaris.
The evaluated \modelname runtime uses adaptive decode scheduling, shared activation preparation, and CUDA Graph execution.
All compared methods use matched CUDA Graph settings, attention backend, KV-cache configuration, and measurement protocol; method-specific quantization kernels follow their evaluated implementations.
The cross-model workload uses batch size 1, a 128-token prompt, and 512 generated tokens.
The reference Llama-3.2-3B A100 measurement uses two warm-up runs and five timed repetitions, reporting median latency, throughput, and energy.
Measurements include input and weight preparation, projection computation, scale restoration, reduction, and output conversion.
Headline generation speedups are BF16-to-\modelname TPOT ratios for matching models, GPUs, and workloads.
Candidate construction, calibration, and packing precede request timing; online operand preparation, including the long-prefill INT8 buffer, is included.

The 80-setting sweep combines prompt lengths $\{128,2048,8192,16384\}$ tokens, output lengths $\{128,256,512,1024\}$ tokens, and batch sizes $\{1,2,4,8,16\}$.
It compares \modelname, BF16, Hugging Face FP16, uniform 4-bit weights with 8-bit activations, GPTQ, AWQ, and MixLLM; Figure~\ref{fig:generation_throughput_grid} shows batches 1, 2, 4, and 8.
The separate peak-memory sweep uses prompt lengths of 128, 512, 2K, 4K, 8K, and 16K tokens (Table~\ref{tab:peak_vram_scaling_llama3b}).

\subsection{Ablation Protocols}
\label{app:methodology_suite}

Sensitivity-localization statistics cover all four models.
Tile-size and calibration-data studies use Llama-3.2-3B, Qwen3-4B, and Ministral-3-8B.
Tile sizes $32\times32$, $64\times64$, and $128\times128$ are evaluated with held-out prefill and decode NLL.
Calibration uses 8, 16, 24, and 48 documents with seeds 0, 1, and 2; Table~\ref{tab:calibration_robustness} reports the mean, sample standard deviation, minimum, and maximum of General Avg across seeds.
Main-table results use the deterministic record construction in Appendix~\ref{app:calibration_evaluation}.
Means and standard deviations in this sweep summarize the sampled runs within each calibration-set size.

On Llama-3.2-3B, the budget sweep uses $\rho\in\{0,0.4,0.5,0.6,0.7,0.8,1\}$ with the final runtime (Table~\ref{tab:w4_budget_sweep}).
Runtime controls use fixed decode splitting, separate activation preparation, or individual kernel launches under the same A100 generation workload (Table~\ref{tab:runtime_optimization_ablation}).

\subsection{Storage and Memory Accounting}
\label{app:storage_measurements}

Packed BPW sums the bytes in \eqref{eq:app_packed_bytes} over quantized projections, multiplies by eight, and divides by their total weight count.
Full-model storage adds the remaining parameters and serialization metadata.
Resident and peak allocated GPU memory are reported in GiB, with activations, key/value states, and temporary weights contributing according to their runtime lifetimes.

\section{Additional Results and Analysis}
\label{app:additional-results}

\subsection{General Quality Results}
\label{app:general_results}

To examine whether the quality preservation extends across model families and sizes, we repeat language modeling and downstream evaluation on Qwen3-4B, Ministral-3-8B, and Qwen3-14B.
Tables~\ref{tab:general_qwen4b}, \ref{tab:general_ministral8b}, and~\ref{tab:general_qwen14b} show a recurring pattern: selective W8 retention improves the reported task scores over W4A8 while preserving the predictive behavior of the original model.
For Qwen3-4B, LAMBADA accuracy reaches 63.38\%, alongside 63.40\% for BF16 and 58.14\% for W4A8.
For Qwen3-14B, HellaSwag accuracy is 78.77\% alongside 78.83\% for BF16, and WikiText-2 perplexity is 10.94 alongside 10.78.
The preservation therefore appears in both correct-answer prediction and the probabilities assigned to text.

This behavior is consistent with the numerical object evaluated during allocation.
As detailed in Appendices~\ref{app:perturbation_derivation} and~\ref{app:fisher_derivation}, each score measures the output change caused by replacing a reconstructed W8 candidate with its W4 candidate under the observed A8 inputs.
The loss-gradient weighting connects that local change to the model's prediction objective.
Each model supplies its own candidate differences, activations, and gradients, so the same procedure can assign W8 to different regions under a common storage budget.
The downstream results support these selected regions as useful locations for retaining additional numerical resolution.

\textbf{Allocation controls.}
The matched-budget controls examine how much of this quality depends on the spatial freedom of the allocation.
On Qwen3-4B, \modelname achieves 13.15 perplexity, compared with 13.32 to 13.88 for the structured and random placements (Table~\ref{tab:general_qwen4b_ablation}).
On Qwen3-14B, LAMBADA accuracy reaches 69.65\%, compared with 67.55\% to 67.79\% for the structured controls and 66.45\% for random placement (Table~\ref{tab:general_qwen14b_ablation}).
These comparisons hold the W4 fraction fixed, so the improved predictions reflect a more effective placement of the available W8 tiles.
The model-wide selection in Appendix~\ref{app:allocation_details} permits this capacity to follow the tile scores across projections.
Together, these results support retaining local choices in model-wide allocation.

\begin{table*}[t]
\centering
\scriptsize
\setlength{\tabcolsep}{3.0pt}
\renewcommand{\arraystretch}{0.75}
\caption{
General benchmark results for \textbf{Qwen3-4B-Instruct-2507}.
Accuracy values are reported as percentages, while WikiText-2 is reported with word perplexity.
Cell colors are relative to BF16 under the corresponding metric direction: green for better than or equal to BF16, blue for worse than BF16, and gray for BF16.
Bold values indicate the best result among the compared quantized methods, excluding the reference columns.
}
\label{tab:general_qwen4b}

\begin{adjustbox}{width=\textwidth}
\begin{tabular}{l c c c c c c c c}
\toprule
\multirow{2}{*}{\textbf{Benchmark}} &
\multirow{2}{*}{\textbf{Metric}} &
\multicolumn{1}{c}{\textbf{Upper Ref.}} &
\multicolumn{1}{c}{\textbf{Lower Ref.}} &
\multicolumn{1}{c}{\textbf{Ours}} &
\multicolumn{4}{c}{\textbf{Others}} \\
\cmidrule(lr){3-3}
\cmidrule(lr){4-4}
\cmidrule(lr){5-5}
\cmidrule(lr){6-9}
& &
\textbf{BF16} &
\textbf{W4A8} &
\textbf{\modelname} &
\textbf{GPTQ} &
\textbf{AWQ} &
\textbf{MixLLM} &
\textbf{SFMP} \\
\midrule

WikiText-2 $\downarrow$
& PPL \setbase{13.07}
& \cellcolor{heatneutral}\base
& \pcell{14.44}
& \textbf{\pcell{13.15}}
& \pcell{13.72}
& \pcell{13.71}
& \pcell{13.19}
& \pcell{13.96} \\

HellaSwag $\uparrow$
& Acc. \setbase{69.14}
& \cellcolor{heatneutral}\base
& \gcell{66.87}
& \textbf{\gcell{68.85}}
& \gcell{51.53}
& \gcell{51.71}
& \gcell{68.81}
& \gcell{67.91} \\

MMLU $\uparrow$
& Acc. \setbase{74.72}
& \cellcolor{heatneutral}\base
& \gcell{72.74}
& \textbf{\gcell{74.22}}
& \gcell{73.09}
& \gcell{73.03}
& \gcell{72.10}
& \gcell{71.00} \\

ARC-Challenge $\uparrow$
& Acc. \setbase{66.38}
& \cellcolor{heatneutral}\base
& \gcell{64.08}
& \textbf{\gcell{66.64}}
& \gcell{61.35}
& \gcell{61.09}
& \gcell{66.54}
& \gcell{64.76} \\

LAMBADA $\uparrow$
& Acc. \setbase{63.40}
& \cellcolor{heatneutral}\base
& \gcell{58.14}
& \textbf{\gcell{63.38}}
& \gcell{59.95}
& \gcell{62.51}
& \gcell{63.21}
& \gcell{61.25} \\

PIQA $\uparrow$
& Acc. \setbase{76.06}
& \cellcolor{heatneutral}\base
& \gcell{76.66}
& \textbf{\gcell{76.67}}
& \gcell{75.19}
& \gcell{75.08}
& \gcell{76.33}
& \gcell{76.06} \\

WinoGrande $\uparrow$
& Acc. \setbase{66.38}
& \cellcolor{heatneutral}\base
& \gcell{66.54}
& \textbf{\gcell{66.77}}
& \gcell{66.22}
& \gcell{65.27}
& \gcell{65.98}
& \gcell{65.90} \\

\bottomrule
\end{tabular}
\end{adjustbox}
\end{table*}

\begin{table*}[t]
\centering
\scriptsize
\setlength{\tabcolsep}{3.0pt}
\renewcommand{\arraystretch}{0.75}
\caption{
Ablation results for \textbf{Qwen3-4B-Instruct-2507}.
Accuracy values are reported as percentages, while WikiText-2 is reported with word perplexity.
Cell colors are relative to BF16 under the corresponding metric direction: green for better than or equal to BF16, blue for worse than BF16, and gray for BF16.
Bold values indicate the best result among the non-reference methods.
}
\label{tab:general_qwen4b_ablation}

\begin{adjustbox}{width=\textwidth}
\begin{tabular}{l c c c c c c c c}
\toprule
\multirow{2}{*}{\textbf{Benchmark}} &
\multirow{2}{*}{\textbf{Metric}} &
\multicolumn{1}{c}{\textbf{Upper Ref.}} &
\multicolumn{1}{c}{\textbf{Lower Ref.}} &
\multicolumn{1}{c}{\textbf{Ours}} &
\multicolumn{4}{c}{\textbf{Ablation}} \\
\cmidrule(lr){3-3}
\cmidrule(lr){4-4}
\cmidrule(lr){5-5}
\cmidrule(lr){6-9}
& &
\textbf{BF16} &
\textbf{W4A8} &
\textbf{\modelname} &
\textbf{Output-row} &
\textbf{Red.-col} &
\textbf{Proj./layer} &
\textbf{Random} \\
\midrule

WikiText-2 $\downarrow$
& PPL \setbase{13.07}
& \cellcolor{heatneutral}\base
& \pcell{14.44}
& \textbf{\pcell{13.15}}
& \pcell{13.40}
& \pcell{13.32}
& \pcell{13.35}
& \pcell{13.88} \\

HellaSwag $\uparrow$
& Acc. \setbase{69.14}
& \cellcolor{heatneutral}\base
& \gcell{66.87}
& \textbf{\gcell{68.85}}
& \gcell{68.15}
& \gcell{68.54}
& \gcell{67.91}
& \gcell{67.34} \\

MMLU $\uparrow$
& Acc. \setbase{74.72}
& \cellcolor{heatneutral}\base
& \gcell{72.74}
& \gcell{74.22}
& \gcell{74.14}
& \textbf{\gcell{74.30}}
& \gcell{74.19}
& \gcell{73.41} \\

ARC-Challenge $\uparrow$
& Acc. \setbase{66.38}
& \cellcolor{heatneutral}\base
& \gcell{64.08}
& \textbf{\gcell{66.64}}
& \gcell{65.70}
& \gcell{66.30}
& \gcell{65.96}
& \gcell{65.44} \\

LAMBADA $\uparrow$
& Acc. \setbase{63.40}
& \cellcolor{heatneutral}\base
& \gcell{58.14}
& \gcell{63.38}
& \gcell{62.88}
& \textbf{\gcell{63.42}}
& \gcell{63.13}
& \gcell{59.93} \\

PIQA $\uparrow$
& Acc. \setbase{76.06}
& \cellcolor{heatneutral}\base
& \gcell{76.66}
& \textbf{\gcell{76.67}}
& \gcell{76.17}
& \gcell{76.33}
& \gcell{76.17}
& \gcell{76.01} \\

WinoGrande $\uparrow$
& Acc. \setbase{66.38}
& \cellcolor{heatneutral}\base
& \gcell{66.54}
& \textbf{\gcell{66.77}}
& \gcell{66.14}
& \gcell{66.69}
& \gcell{66.22}
& \gcell{65.27} \\

\bottomrule
\end{tabular}
\end{adjustbox}
\end{table*}

\begin{table*}[t]
\centering
\scriptsize
\setlength{\tabcolsep}{3.0pt}
\renewcommand{\arraystretch}{0.75}
\caption{
General benchmark results for \textbf{Qwen3-14B}.
Accuracy values are reported as percentages, while WikiText-2 is reported with word perplexity.
Cell colors are relative to BF16 under the corresponding metric direction: green for better than or equal to BF16, blue for worse than BF16, and gray for BF16.
Bold values indicate the best result among the compared quantized methods, excluding the reference columns.
}
\label{tab:general_qwen14b}

\begin{adjustbox}{width=\textwidth}
\begin{tabular}{l c c c c c c c c}
\toprule
\multirow{2}{*}{\textbf{Benchmark}} &
\multirow{2}{*}{\textbf{Metric}} &
\multicolumn{1}{c}{\textbf{Upper Ref.}} &
\multicolumn{1}{c}{\textbf{Lower Ref.}} &
\multicolumn{1}{c}{\textbf{Ours}} &
\multicolumn{4}{c}{\textbf{Others}} \\
\cmidrule(lr){3-3}
\cmidrule(lr){4-4}
\cmidrule(lr){5-5}
\cmidrule(lr){6-9}
& &
\textbf{BF16} &
\textbf{W4A8} &
\textbf{\modelname} &
\textbf{GPTQ} &
\textbf{AWQ} &
\textbf{MixLLM} &
\textbf{SFMP} \\
\midrule

WikiText-2 $\downarrow$
& PPL \setbase{10.78}
& \cellcolor{heatneutral}\base
& \pcell{11.49}
& \textbf{\pcell{10.94}}
& \pcell{11.14}
& \pcell{11.24}
& \pcell{10.96}
& \pcell{11.39} \\

HellaSwag $\uparrow$
& Acc. \setbase{78.83}
& \cellcolor{heatneutral}\base
& \gcell{77.92}
& \textbf{\gcell{78.77}}
& \gcell{59.97}
& \gcell{60.18}
& \gcell{78.70}
& \gcell{78.25} \\

MMLU $\uparrow$
& Acc. \setbase{80.72}
& \cellcolor{heatneutral}\base
& \gcell{79.62}
& \textbf{\gcell{80.15}}
& \gcell{79.74}
& \gcell{79.92}
& \gcell{78.58}
& \gcell{77.98} \\

ARC-Challenge $\uparrow$
& Acc. \setbase{69.97}
& \cellcolor{heatneutral}\base
& \gcell{68.94}
& \textbf{\gcell{69.65}}
& \gcell{67.32}
& \gcell{65.70}
& \gcell{69.62}
& \gcell{69.11} \\

LAMBADA $\uparrow$
& Acc. \setbase{67.86}
& \cellcolor{heatneutral}\base
& \gcell{65.86}
& \textbf{\gcell{69.65}}
& \gcell{67.01}
& \gcell{67.22}
& \gcell{69.12}
& \gcell{68.19} \\

PIQA $\uparrow$
& Acc. \setbase{79.65}
& \cellcolor{heatneutral}\base
& \gcell{78.56}
& \textbf{\gcell{79.87}}
& \gcell{79.33}
& \gcell{79.82}
& \gcell{79.33}
& \gcell{79.82} \\

WinoGrande $\uparrow$
& Acc. \setbase{75.43}
& \cellcolor{heatneutral}\base
& \gcell{74.59}
& \textbf{\gcell{75.66}}
& \gcell{75.14}
& \gcell{73.95}
& \gcell{74.27}
& \gcell{72.69} \\

\bottomrule
\end{tabular}
\end{adjustbox}
\end{table*}

\begin{table*}[t]
\centering
\scriptsize
\setlength{\tabcolsep}{3.0pt}
\renewcommand{\arraystretch}{0.75}
\caption{
Ablation results for \textbf{Qwen3-14B}.
Accuracy values are reported as percentages, while WikiText-2 is reported with word perplexity.
Cell colors are relative to BF16 under the corresponding metric direction: green for better than or equal to BF16, blue for worse than BF16, and gray for BF16.
Bold values indicate the best result among the non-reference methods.
}
\label{tab:general_qwen14b_ablation}

\begin{adjustbox}{width=\textwidth}
\begin{tabular}{l c c c c c c c c}
\toprule
\multirow{2}{*}{\textbf{Benchmark}} &
\multirow{2}{*}{\textbf{Metric}} &
\multicolumn{1}{c}{\textbf{Upper Ref.}} &
\multicolumn{1}{c}{\textbf{Lower Ref.}} &
\multicolumn{1}{c}{\textbf{Ours}} &
\multicolumn{4}{c}{\textbf{Ablation}} \\
\cmidrule(lr){3-3}
\cmidrule(lr){4-4}
\cmidrule(lr){5-5}
\cmidrule(lr){6-9}
& &
\textbf{BF16} &
\textbf{W4A8} &
\textbf{\modelname} &
\textbf{Output-row} &
\textbf{Red.-col} &
\textbf{Proj./layer} &
\textbf{Random} \\
\midrule

WikiText-2 $\downarrow$
& PPL \setbase{10.78}
& \cellcolor{heatneutral}\base
& \pcell{11.49}
& \pcell{10.94}
& \pcell{10.97}
& \textbf{\pcell{10.88}}
& \pcell{11.02}
& \pcell{11.22} \\

HellaSwag $\uparrow$
& Acc. \setbase{78.83}
& \cellcolor{heatneutral}\base
& \gcell{77.92}
& \textbf{\gcell{78.77}}
& \gcell{78.65}
& \gcell{78.56}
& \gcell{78.61}
& \gcell{78.45} \\

MMLU $\uparrow$
& Acc. \setbase{80.72}
& \cellcolor{heatneutral}\base
& \gcell{79.62}
& \gcell{80.15}
& \textbf{\gcell{80.30}}
& \gcell{80.24}
& \gcell{80.03}
& \gcell{80.00} \\

ARC-Challenge $\uparrow$
& Acc. \setbase{69.97}
& \cellcolor{heatneutral}\base
& \gcell{68.94}
& \textbf{\gcell{69.65}}
& \gcell{69.03}
& \gcell{69.20}
& \gcell{69.54}
& \gcell{68.34} \\

LAMBADA $\uparrow$
& Acc. \setbase{67.86}
& \cellcolor{heatneutral}\base
& \gcell{65.86}
& \textbf{\gcell{69.65}}
& \gcell{67.79}
& \gcell{67.55}
& \gcell{67.79}
& \gcell{66.45} \\

PIQA $\uparrow$
& Acc. \setbase{79.65}
& \cellcolor{heatneutral}\base
& \gcell{78.56}
& \gcell{79.87}
& \gcell{79.60}
& \gcell{79.87}
& \textbf{\gcell{80.36}}
& \gcell{79.16} \\

WinoGrande $\uparrow$
& Acc. \setbase{75.43}
& \cellcolor{heatneutral}\base
& \gcell{74.59}
& \textbf{\gcell{75.66}}
& \gcell{75.06}
& \gcell{74.51}
& \gcell{74.59}
& \gcell{74.35} \\

\bottomrule
\end{tabular}
\end{adjustbox}
\end{table*}

\begin{table*}[t]
\centering
\scriptsize
\setlength{\tabcolsep}{2.5pt}
\renewcommand{\arraystretch}{0.75}
\caption{
General benchmark and ablation results for \textbf{Ministral-3-8B-Instruct-2512-BF16}.
Accuracy values are reported as percentages, while WikiText-2 is reported with word perplexity.
Cell colors are relative to BF16 under the corresponding metric direction: green for better than or equal to BF16, blue for worse than BF16, and gray for BF16.
Bold values indicate the best result among the non-reference methods.
}
\label{tab:general_ministral8b}

\begin{adjustbox}{width=\textwidth}
\begin{tabular}{l c c c c c c c c c c}
\toprule
\multirow{2}{*}{\textbf{Benchmark}} &
\multirow{2}{*}{\textbf{Metric}} &
\multicolumn{1}{c}{\textbf{Upper Ref.}} &
\multicolumn{1}{c}{\textbf{Lower Ref.}} &
\multicolumn{1}{c}{\textbf{Ours}} &
\multicolumn{2}{c}{\textbf{Others}} &
\multicolumn{4}{c}{\textbf{Ablation}} \\
\cmidrule(lr){3-3}
\cmidrule(lr){4-4}
\cmidrule(lr){5-5}
\cmidrule(lr){6-7}
\cmidrule(lr){8-11}
& &
\textbf{BF16} &
\textbf{W4A8} &
\textbf{\modelname} &
\textbf{GPTQ} &
\textbf{AWQ} &
\textbf{Output-row} &
\textbf{Red.-col} &
\textbf{Proj./layer} &
\textbf{Random} \\
\midrule

WikiText-2 $\downarrow$
& PPL \setbase{9.73}
& \cellcolor{heatneutral}\base
& \pcell{10.24}
& \pcell{9.87}
& \pcell{9.90}
& \pcell{9.93}
& \pcell{9.90}
& \textbf{\pcell{9.86}}
& \pcell{9.90}
& \pcell{10.01} \\

HellaSwag $\uparrow$
& Acc. \setbase{78.79}
& \cellcolor{heatneutral}\base
& \gcell{78.09}
& \textbf{\gcell{78.74}}
& \gcell{58.17}
& \gcell{58.13}
& \gcell{78.40}
& \gcell{78.62}
& \gcell{78.59}
& \gcell{78.42} \\

MMLU $\uparrow$
& Acc. \setbase{77.85}
& \cellcolor{heatneutral}\base
& \gcell{76.46}
& \gcell{77.22}
& \gcell{77.05}
& \gcell{76.88}
& \gcell{77.18}
& \textbf{\gcell{77.26}}
& \gcell{76.98}
& \gcell{76.77} \\

ARC-Challenge $\uparrow$
& Acc. \setbase{69.03}
& \cellcolor{heatneutral}\base
& \gcell{68.00}
& \textbf{\gcell{68.77}}
& \gcell{64.25}
& \gcell{63.57}
& \gcell{68.17}
& \gcell{68.34}
& \gcell{68.34}
& \gcell{68.52} \\

LAMBADA $\uparrow$
& Acc. \setbase{70.99}
& \cellcolor{heatneutral}\base
& \gcell{69.67}
& \gcell{71.76}
& \gcell{70.83}
& \gcell{71.03}
& \gcell{70.75}
& \gcell{71.73}
& \textbf{\gcell{71.86}}
& \gcell{71.45} \\

PIQA $\uparrow$
& Acc. \setbase{80.52}
& \cellcolor{heatneutral}\base
& \gcell{79.27}
& \textbf{\gcell{79.92}}
& \gcell{78.51}
& \gcell{79.05}
& \gcell{79.22}
& \gcell{79.87}
& \textbf{\gcell{79.92}}
& \gcell{79.71} \\

WinoGrande $\uparrow$
& Acc. \setbase{77.98}
& \cellcolor{heatneutral}\base
& \gcell{75.06}
& \gcell{76.87}
& \gcell{76.72}
& \gcell{76.09}
& \textbf{\gcell{77.03}}
& \gcell{76.87}
& \gcell{76.48}
& \gcell{76.40} \\

\bottomrule
\end{tabular}
\end{adjustbox}
\end{table*}

\subsection{LV-Eval Task Results}
\label{app:lveval_results}

The additional LV-Eval experiments examine whether the shared precision maps preserve information use as context length increases across different models.
Qwen3-4B maintains a nearly level CMRC trajectory, scoring 24.95, 25.30, and 24.74 at 16K, 32K, and 64K, with the final value matching BF16 (Table~\ref{tab:lveval_qwen4b}).
Ministral-3-8B preserves a different length response on LooGLE-SD: its score changes from 59.44 at 16K to 56.18 at 64K, alongside the BF16 change from 58.47 to 55.09 (Table~\ref{tab:lveval_ministral8b}).
Preservation here follows each model's task-specific trajectory across lengths.
The selected representation retains the first example's nearly level response and the second's gradual change.

Qwen3-14B shows a stronger dependence on selective precision at the largest context length (Table~\ref{tab:lveval_qwen14b}).
On LooGLE-SD, \modelname scores 21.51, 18.96, and 17.86 from 16K to 64K, while W4A8 scores 18.54, 15.58, and 8.55.
A related pattern appears in English multi-field question answering (\ds{8}), where the 64K score is 14.28 with \modelname, 14.66 with BF16, and 5.89 with W4A8.
The separation across these tasks shows that the value of higher precision extends to several forms of long-input information use.
The retained W8 regions help preserve the model's ability to extract and use relevant evidence at lengths where the precision choice has a larger effect on the answer.

\textbf{Allocation controls.}
The fixed-budget comparisons further distinguish the effect of precision placement.
For Qwen3-14B on LooGLE-SD, the strongest structured control changes from 21.03 at 16K to 10.11 at 64K, compared with 21.51 and 17.86 for \modelname.
For Ministral-3-8B on LooGLE-CR at 64K, \modelname scores 26.77 versus 21.45 to 23.37 for the structured and random allocations.
The available W8 capacity is the same, yet the locally selected map preserves more contextual information in both cases.

The shared-map design connects these results to the calibration procedure in Appendices~\ref{app:phase_calibration} and~\ref{app:phase_aggregation}.
Prefill and decode contribute their own activation and gradient observations, and their normalized scores determine one selected representation per tile.
That representation is reused across the evaluated context lengths.
The task trajectories show that the resulting choices continue to protect useful predictions as the amount of input context changes.

\begin{table*}[t]
\centering
\scriptsize
\setlength{\tabcolsep}{2.0pt}
\renewcommand{\arraystretch}{0.75}
\caption{
Long-context question answering results for
\textbf{Qwen3-4B-Instruct-2507} on LV-Eval across nine datasets
and three context lengths.
Scores are reported as percentages.
Cell colors are relative to BF16: green for higher scores,
blue for lower scores, and gray for BF16.
Bold values indicate the best result among the non-reference methods.
}
\label{tab:lveval_qwen4b}

\begin{adjustbox}{width=\textwidth}
\begin{tabular}{c c c c c c c c c c c c c}
\toprule
\multirow{2}{*}{\textbf{Dataset}} &
\multirow{2}{*}{\textbf{Len}} &
\multicolumn{1}{c}{\textbf{Upper Ref.}} &
\multicolumn{1}{c}{\textbf{Lower Ref.}} &
\multicolumn{1}{c}{\textbf{Ours}} &
\multicolumn{4}{c}{\textbf{Others}} &
\multicolumn{4}{c}{\textbf{Ablation}} \\
\cmidrule(lr){3-3}
\cmidrule(lr){4-4}
\cmidrule(lr){5-5}
\cmidrule(lr){6-9}
\cmidrule(lr){10-13}
& &
\textbf{BF16} &
\textbf{W4A8} &
\textbf{\modelname} &
\textbf{GPTQ} &
\textbf{AWQ} &
\textbf{MixLLM} &
\textbf{SFMP} &
\textbf{Output-row} &
\textbf{Red.-col} &
\textbf{Proj./layer} &
\textbf{Random} \\
\midrule

\multirow{3}{*}{\ds{1}}
& 16k \setbase{24.57}
& \cellcolor{heatneutral}\base
& \gcell{23.04}
& \gcell{24.95}
& \gcell{23.86}
& \gcell{24.94}
& \gcell{24.20}
& \gcell{23.89}
& \textbf{\gcell{25.57}}
& \gcell{24.89}
& \gcell{23.49}
& \gcell{23.37} \\
& 32k \setbase{25.68}
& \cellcolor{heatneutral}\base
& \gcell{21.27}
& \gcell{25.30}
& \gcell{23.90}
& \textbf{\gcell{26.56}}
& \gcell{25.01}
& \gcell{26.04}
& \gcell{23.79}
& \gcell{22.89}
& \gcell{21.48}
& \gcell{21.38} \\
& 64k \setbase{24.74}
& \cellcolor{heatneutral}\base
& \gcell{20.39}
& \textbf{\gcell{24.74}}
& \gcell{21.12}
& \gcell{23.99}
& \gcell{23.75}
& \gcell{23.18}
& \gcell{22.57}
& \gcell{21.66}
& \gcell{20.03}
& \gcell{20.96} \\
\midrule

\multirow{3}{*}{\ds{2}}
& 16k \setbase{19.20}
& \cellcolor{heatneutral}\base
& \gcell{18.75}
& \textbf{\gcell{19.90}}
& \gcell{16.96}
& \gcell{18.89}
& \gcell{18.92}
& \gcell{19.35}
& \gcell{19.67}
& \gcell{19.56}
& \gcell{19.40}
& \gcell{19.70} \\
& 32k \setbase{19.77}
& \cellcolor{heatneutral}\base
& \gcell{18.01}
& \textbf{\gcell{19.26}}
& \gcell{18.41}
& \gcell{17.63}
& \gcell{19.01}
& \gcell{18.68}
& \gcell{18.94}
& \gcell{19.06}
& \gcell{18.76}
& \gcell{18.40} \\
& 64k \setbase{19.83}
& \cellcolor{heatneutral}\base
& \gcell{19.51}
& \textbf{\gcell{20.65}}
& \gcell{18.92}
& \gcell{19.15}
& \gcell{19.86}
& \gcell{19.15}
& \gcell{18.62}
& \gcell{19.40}
& \gcell{19.50}
& \gcell{19.19} \\
\midrule

\multirow{3}{*}{\ds{3}}
& 16k \setbase{4.78}
& \cellcolor{heatneutral}\base
& \gcell{2.72}
& \textbf{\gcell{5.01}}
& \gcell{4.76}
& \gcell{4.80}
& \gcell{4.85}
& \gcell{4.43}
& \gcell{4.95}
& \gcell{4.94}
& \gcell{4.80}
& \gcell{3.89} \\
& 32k \setbase{4.46}
& \cellcolor{heatneutral}\base
& \gcell{2.65}
& \gcell{4.77}
& \gcell{4.66}
& \textbf{\gcell{4.87}}
& \gcell{4.80}
& \gcell{4.53}
& \gcell{4.75}
& \gcell{4.63}
& \gcell{4.45}
& \gcell{3.48} \\
& 64k \setbase{4.67}
& \cellcolor{heatneutral}\base
& \gcell{2.62}
& \textbf{\gcell{4.76}}
& \gcell{3.76}
& \gcell{4.64}
& \gcell{4.51}
& \gcell{4.12}
& \gcell{4.37}
& \gcell{4.49}
& \gcell{4.37}
& \gcell{4.06} \\
\midrule

\multirow{3}{*}{\ds{4}}
& 16k \setbase{22.31}
& \cellcolor{heatneutral}\base
& \gcell{21.31}
& \textbf{\gcell{22.57}}
& \gcell{20.93}
& \gcell{20.98}
& \gcell{22.54}
& \gcell{21.03}
& \gcell{21.32}
& \gcell{22.02}
& \gcell{21.15}
& \gcell{20.64} \\
& 32k \setbase{22.91}
& \cellcolor{heatneutral}\base
& \gcell{18.51}
& \textbf{\gcell{22.20}}
& \gcell{20.99}
& \gcell{21.61}
& \gcell{21.16}
& \gcell{21.67}
& \gcell{18.46}
& \gcell{20.04}
& \gcell{19.01}
& \gcell{18.62} \\
& 64k \setbase{21.65}
& \cellcolor{heatneutral}\base
& \gcell{17.97}
& \textbf{\gcell{22.63}}
& \gcell{22.05}
& \gcell{22.49}
& \gcell{22.26}
& \gcell{22.10}
& \gcell{17.41}
& \gcell{19.63}
& \gcell{18.47}
& \gcell{17.69} \\
\midrule

\multirow{3}{*}{\ds{5}}
& 16k \setbase{14.77}
& \cellcolor{heatneutral}\base
& \gcell{12.19}
& \textbf{\gcell{14.65}}
& \gcell{13.92}
& \gcell{13.90}
& \gcell{14.36}
& \gcell{14.06}
& \gcell{14.54}
& \gcell{14.55}
& \gcell{14.30}
& \gcell{13.38} \\
& 32k \setbase{14.42}
& \cellcolor{heatneutral}\base
& \gcell{11.16}
& \textbf{\gcell{14.30}}
& \gcell{13.12}
& \textbf{\gcell{14.30}}
& \gcell{14.04}
& \gcell{14.11}
& \gcell{13.88}
& \gcell{13.95}
& \gcell{14.01}
& \gcell{13.34} \\
& 64k \setbase{14.24}
& \cellcolor{heatneutral}\base
& \gcell{10.01}
& \textbf{\gcell{13.75}}
& \gcell{12.60}
& \gcell{13.71}
& \gcell{13.04}
& \gcell{12.84}
& \gcell{12.95}
& \gcell{13.07}
& \gcell{13.23}
& \gcell{12.36} \\
\midrule

\multirow{3}{*}{\ds{6}}
& 16k \setbase{10.26}
& \cellcolor{heatneutral}\base
& \gcell{8.10}
& \gcell{10.91}
& \gcell{10.16}
& \gcell{10.66}
& \gcell{10.79}
& \gcell{10.27}
& \textbf{\gcell{11.00}}
& \gcell{10.82}
& \gcell{10.49}
& \gcell{9.02} \\
& 32k \setbase{9.88}
& \cellcolor{heatneutral}\base
& \gcell{7.33}
& \textbf{\gcell{9.89}}
& \gcell{9.19}
& \gcell{9.58}
& \gcell{9.15}
& \gcell{9.47}
& \gcell{9.63}
& \gcell{9.21}
& \gcell{9.66}
& \gcell{7.60} \\
& 64k \setbase{9.07}
& \cellcolor{heatneutral}\base
& \gcell{6.30}
& \textbf{\gcell{9.60}}
& \gcell{8.97}
& \gcell{8.64}
& \gcell{9.38}
& \gcell{8.70}
& \gcell{8.57}
& \gcell{8.21}
& \gcell{8.67}
& \gcell{6.72} \\
\midrule

\multirow{3}{*}{\ds{7}}
& 16k \setbase{24.43}
& \cellcolor{heatneutral}\base
& \gcell{21.08}
& \gcell{23.96}
& \gcell{22.75}
& \gcell{23.54}
& \gcell{23.74}
& \gcell{22.65}
& \gcell{23.57}
& \gcell{23.60}
& \textbf{\gcell{24.06}}
& \gcell{22.46} \\
& 32k \setbase{22.95}
& \cellcolor{heatneutral}\base
& \gcell{18.90}
& \gcell{22.15}
& \gcell{21.54}
& \gcell{21.90}
& \textbf{\gcell{22.20}}
& \gcell{22.14}
& \gcell{21.31}
& \gcell{21.37}
& \gcell{22.04}
& \gcell{20.47} \\
& 64k \setbase{21.96}
& \cellcolor{heatneutral}\base
& \gcell{17.88}
& \textbf{\gcell{20.46}}
& \gcell{19.99}
& \gcell{20.05}
& \gcell{20.30}
& \gcell{19.46}
& \gcell{19.04}
& \gcell{19.11}
& \gcell{19.75}
& \gcell{18.26} \\
\midrule

\multirow{3}{*}{\ds{8}}
& 16k \setbase{18.89}
& \cellcolor{heatneutral}\base
& \gcell{16.46}
& \textbf{\gcell{19.78}}
& \gcell{18.84}
& \gcell{19.15}
& \gcell{19.38}
& \gcell{18.39}
& \gcell{19.42}
& \gcell{19.38}
& \gcell{19.60}
& \gcell{17.46} \\
& 32k \setbase{18.32}
& \cellcolor{heatneutral}\base
& \gcell{16.35}
& \textbf{\gcell{19.51}}
& \gcell{19.43}
& \gcell{18.66}
& \gcell{18.60}
& \gcell{18.72}
& \gcell{19.40}
& \gcell{19.18}
& \gcell{18.86}
& \gcell{17.12} \\
& 64k \setbase{18.61}
& \cellcolor{heatneutral}\base
& \gcell{15.14}
& \gcell{17.93}
& \gcell{18.00}
& \gcell{17.30}
& \gcell{17.89}
& \gcell{16.83}
& \gcell{17.92}
& \textbf{\gcell{18.02}}
& \gcell{17.82}
& \gcell{15.80} \\
\midrule

\multirow{3}{*}{\ds{9}}
& 16k \setbase{17.70}
& \cellcolor{heatneutral}\base
& \gcell{13.11}
& \textbf{\gcell{17.28}}
& \gcell{15.66}
& \gcell{15.69}
& \textbf{\gcell{17.28}}
& \gcell{16.54}
& \gcell{16.32}
& \gcell{17.02}
& \gcell{16.25}
& \gcell{13.94} \\
& 32k \setbase{16.25}
& \cellcolor{heatneutral}\base
& \gcell{12.33}
& \textbf{\gcell{16.77}}
& \gcell{15.01}
& \gcell{15.59}
& \gcell{16.76}
& \gcell{14.83}
& \gcell{15.01}
& \gcell{16.25}
& \gcell{15.19}
& \gcell{13.21} \\
& 64k \setbase{15.20}
& \cellcolor{heatneutral}\base
& \gcell{12.58}
& \textbf{\gcell{15.12}}
& \gcell{13.75}
& \gcell{14.84}
& \gcell{14.29}
& \gcell{14.74}
& \gcell{14.85}
& \gcell{14.88}
& \gcell{14.33}
& \gcell{12.70} \\

\bottomrule
\end{tabular}
\end{adjustbox}
\end{table*}

\begin{table*}[t]
\centering
\scriptsize
\setlength{\tabcolsep}{2.0pt}
\renewcommand{\arraystretch}{0.75}
\caption{
Long-context question answering results for
\textbf{Ministral-3-8B-Instruct-2512-BF16} on LV-Eval across nine datasets
and three context lengths.
Scores are reported as percentages.
Cell colors are relative to BF16: green for higher scores,
blue for lower scores, and gray for BF16.
Bold values indicate the best result among the non-reference methods.
}
\label{tab:lveval_ministral8b}

\begin{adjustbox}{width=\textwidth}
\begin{tabular}{c c c c c c c c c c c}
\toprule
\multirow{2}{*}{\textbf{Dataset}} &
\multirow{2}{*}{\textbf{Len}} &
\multicolumn{1}{c}{\textbf{Upper Ref.}} &
\multicolumn{1}{c}{\textbf{Lower Ref.}} &
\multicolumn{1}{c}{\textbf{Ours}} &
\multicolumn{2}{c}{\textbf{Others}} &
\multicolumn{4}{c}{\textbf{Ablation}} \\
\cmidrule(lr){3-3}
\cmidrule(lr){4-4}
\cmidrule(lr){5-5}
\cmidrule(lr){6-7}
\cmidrule(lr){8-11}
& &
\textbf{BF16} &
\textbf{W4A8} &
\textbf{\modelname} &
\textbf{GPTQ} &
\textbf{AWQ} &
\textbf{Output-row} &
\textbf{Red.-col} &
\textbf{Proj./layer} &
\textbf{Random} \\
\midrule

\multirow{3}{*}{\ds{1}}
& 16k \setbase{39.93}
& \cellcolor{heatneutral}\base
& \gcell{39.98}
& \textbf{\gcell{43.79}}
& \gcell{42.75}
& \gcell{41.71}
& \gcell{40.24}
& \gcell{42.30}
& \gcell{40.41}
& \gcell{41.51} \\
& 32k \setbase{40.88}
& \cellcolor{heatneutral}\base
& \gcell{37.41}
& \textbf{\gcell{41.10}}
& \gcell{40.45}
& \gcell{39.48}
& \gcell{39.46}
& \gcell{40.44}
& \gcell{39.78}
& \gcell{38.81} \\
& 64k \setbase{39.94}
& \cellcolor{heatneutral}\base
& \gcell{35.49}
& \textbf{\gcell{41.00}}
& \gcell{40.10}
& \gcell{40.31}
& \gcell{39.54}
& \gcell{40.21}
& \gcell{39.70}
& \gcell{38.76} \\
\midrule

\multirow{3}{*}{\ds{2}}
& 16k \setbase{14.16}
& \cellcolor{heatneutral}\base
& \gcell{14.94}
& \textbf{\gcell{15.82}}
& \gcell{14.38}
& \gcell{14.22}
& \gcell{14.62}
& \gcell{14.84}
& \gcell{14.55}
& \gcell{14.91} \\
& 32k \setbase{15.80}
& \cellcolor{heatneutral}\base
& \gcell{14.77}
& \textbf{\gcell{16.51}}
& \gcell{15.44}
& \gcell{15.73}
& \gcell{15.58}
& \gcell{14.71}
& \gcell{15.37}
& \gcell{15.46} \\
& 64k \setbase{17.23}
& \cellcolor{heatneutral}\base
& \gcell{14.92}
& \textbf{\gcell{17.35}}
& \gcell{16.91}
& \gcell{15.74}
& \gcell{16.49}
& \gcell{16.51}
& \gcell{17.16}
& \gcell{15.68} \\
\midrule

\multirow{3}{*}{\ds{3}}
& 16k \setbase{31.86}
& \cellcolor{heatneutral}\base
& \gcell{30.46}
& \textbf{\gcell{33.00}}
& \gcell{32.54}
& \gcell{31.64}
& \gcell{31.42}
& \gcell{32.77}
& \gcell{31.42}
& \gcell{30.86} \\
& 32k \setbase{32.58}
& \cellcolor{heatneutral}\base
& \gcell{28.81}
& \gcell{32.24}
& \gcell{31.39}
& \gcell{31.64}
& \textbf{\gcell{32.71}}
& \gcell{31.57}
& \gcell{31.19}
& \gcell{31.02} \\
& 64k \setbase{25.24}
& \cellcolor{heatneutral}\base
& \gcell{24.10}
& \textbf{\gcell{25.48}}
& \gcell{24.77}
& \gcell{24.49}
& \gcell{25.14}
& \gcell{24.32}
& \gcell{24.13}
& \gcell{23.69} \\
\midrule

\multirow{3}{*}{\ds{4}}
& 16k \setbase{20.97}
& \cellcolor{heatneutral}\base
& \gcell{21.42}
& \textbf{\gcell{22.18}}
& \gcell{21.88}
& \gcell{20.24}
& \gcell{20.38}
& \gcell{21.33}
& \gcell{21.95}
& \gcell{21.61} \\
& 32k \setbase{21.37}
& \cellcolor{heatneutral}\base
& \gcell{19.44}
& \textbf{\gcell{21.08}}
& \gcell{19.79}
& \gcell{20.15}
& \gcell{19.36}
& \gcell{19.94}
& \gcell{20.34}
& \gcell{18.96} \\
& 64k \setbase{20.72}
& \cellcolor{heatneutral}\base
& \gcell{17.99}
& \textbf{\gcell{20.69}}
& \gcell{19.76}
& \gcell{19.55}
& \gcell{18.13}
& \gcell{18.74}
& \gcell{19.21}
& \gcell{17.82} \\
\midrule

\multirow{3}{*}{\ds{5}}
& 16k \setbase{27.38}
& \cellcolor{heatneutral}\base
& \gcell{24.21}
& \textbf{\gcell{27.64}}
& \gcell{25.41}
& \gcell{26.78}
& \gcell{25.51}
& \gcell{26.91}
& \gcell{26.91}
& \gcell{24.34} \\
& 32k \setbase{26.19}
& \cellcolor{heatneutral}\base
& \gcell{21.41}
& \textbf{\gcell{27.47}}
& \gcell{27.24}
& \gcell{26.73}
& \gcell{23.54}
& \gcell{22.41}
& \gcell{22.56}
& \gcell{22.47} \\
& 64k \setbase{26.72}
& \cellcolor{heatneutral}\base
& \gcell{23.32}
& \gcell{26.77}
& \gcell{25.21}
& \textbf{\gcell{27.42}}
& \gcell{22.14}
& \gcell{21.45}
& \gcell{21.66}
& \gcell{23.37} \\
\midrule

\multirow{3}{*}{\ds{6}}
& 16k \setbase{21.04}
& \cellcolor{heatneutral}\base
& \gcell{19.63}
& \textbf{\gcell{20.99}}
& \gcell{20.43}
& \gcell{20.35}
& \gcell{20.15}
& \gcell{20.05}
& \gcell{19.53}
& \gcell{20.42} \\
& 32k \setbase{19.64}
& \cellcolor{heatneutral}\base
& \gcell{18.55}
& \gcell{19.81}
& \gcell{19.36}
& \gcell{19.61}
& \textbf{\gcell{20.06}}
& \gcell{19.59}
& \gcell{19.33}
& \gcell{18.41} \\
& 64k \setbase{19.24}
& \cellcolor{heatneutral}\base
& \gcell{17.07}
& \textbf{\gcell{18.95}}
& \gcell{18.27}
& \gcell{18.66}
& \gcell{18.43}
& \gcell{18.17}
& \gcell{17.74}
& \gcell{16.73} \\
\midrule

\multirow{3}{*}{\ds{7}}
& 16k \setbase{58.47}
& \cellcolor{heatneutral}\base
& \gcell{56.57}
& \textbf{\gcell{59.44}}
& \gcell{58.60}
& \gcell{56.78}
& \gcell{59.18}
& \gcell{56.01}
& \gcell{59.37}
& \gcell{57.50} \\
& 32k \setbase{57.72}
& \cellcolor{heatneutral}\base
& \gcell{54.13}
& \textbf{\gcell{58.76}}
& \gcell{57.91}
& \gcell{57.13}
& \gcell{56.88}
& \gcell{53.29}
& \gcell{56.41}
& \gcell{55.37} \\
& 64k \setbase{55.09}
& \cellcolor{heatneutral}\base
& \gcell{51.39}
& \gcell{56.18}
& \gcell{55.79}
& \textbf{\gcell{57.74}}
& \gcell{54.47}
& \gcell{52.13}
& \gcell{55.46}
& \gcell{54.47} \\
\midrule

\multirow{3}{*}{\ds{8}}
& 16k \setbase{33.81}
& \cellcolor{heatneutral}\base
& \gcell{32.61}
& \textbf{\gcell{34.40}}
& \gcell{32.80}
& \gcell{34.10}
& \gcell{33.68}
& \gcell{33.84}
& \gcell{32.35}
& \gcell{33.65} \\
& 32k \setbase{31.98}
& \cellcolor{heatneutral}\base
& \gcell{31.46}
& \textbf{\gcell{32.65}}
& \gcell{31.46}
& \gcell{30.70}
& \gcell{32.02}
& \gcell{32.17}
& \gcell{31.02}
& \gcell{31.41} \\
& 64k \setbase{31.26}
& \cellcolor{heatneutral}\base
& \gcell{29.62}
& \textbf{\gcell{33.00}}
& \gcell{32.99}
& \gcell{32.01}
& \gcell{29.97}
& \gcell{30.75}
& \gcell{28.96}
& \gcell{28.92} \\
\midrule

\multirow{3}{*}{\ds{9}}
& 16k \setbase{41.92}
& \cellcolor{heatneutral}\base
& \gcell{42.80}
& \gcell{45.63}
& \gcell{45.18}
& \gcell{42.75}
& \gcell{44.78}
& \gcell{44.05}
& \textbf{\gcell{46.91}}
& \gcell{44.50} \\
& 32k \setbase{42.31}
& \cellcolor{heatneutral}\base
& \gcell{42.48}
& \textbf{\gcell{43.61}}
& \gcell{41.92}
& \gcell{41.47}
& \gcell{43.08}
& \gcell{42.60}
& \gcell{42.63}
& \gcell{41.29} \\
& 64k \setbase{43.98}
& \cellcolor{heatneutral}\base
& \gcell{41.30}
& \gcell{42.95}
& \gcell{42.79}
& \gcell{42.68}
& \gcell{42.87}
& \gcell{42.73}
& \textbf{\gcell{44.13}}
& \gcell{40.29} \\

\bottomrule
\end{tabular}
\end{adjustbox}
\end{table*}

\begin{table*}[t]
\centering
\scriptsize
\setlength{\tabcolsep}{2.0pt}
\renewcommand{\arraystretch}{0.75}
\caption{
Long-context question answering results for
\textbf{Qwen3-14B} on LV-Eval across nine datasets
and three context lengths.
Scores are reported as percentages.
Cell colors are relative to BF16: green for higher scores,
blue for lower scores, and gray for BF16.
Bold values indicate the best result among the non-reference methods.
}
\label{tab:lveval_qwen14b}

\begin{adjustbox}{width=\textwidth}
\begin{tabular}{c c c c c c c c c c c c c}
\toprule
\multirow{2}{*}{\textbf{Dataset}} &
\multirow{2}{*}{\textbf{Len}} &
\multicolumn{1}{c}{\textbf{Upper Ref.}} &
\multicolumn{1}{c}{\textbf{Lower Ref.}} &
\multicolumn{1}{c}{\textbf{Ours}} &
\multicolumn{4}{c}{\textbf{Others}} &
\multicolumn{4}{c}{\textbf{Ablation}} \\
\cmidrule(lr){3-3}
\cmidrule(lr){4-4}
\cmidrule(lr){5-5}
\cmidrule(lr){6-9}
\cmidrule(lr){10-13}
& &
\textbf{BF16} &
\textbf{W4A8} &
\textbf{\modelname} &
\textbf{GPTQ} &
\textbf{AWQ} &
\textbf{MixLLM} &
\textbf{SFMP} &
\textbf{Output-row} &
\textbf{Red.-col} &
\textbf{Proj./layer} &
\textbf{Random} \\
\midrule

\multirow{3}{*}{\ds{1}}
& 16k \setbase{21.12}
& \cellcolor{heatneutral}\base
& \gcell{19.45}
& \textbf{\gcell{21.11}}
& \gcell{20.11}
& \gcell{20.86}
& \gcell{20.34}
& \gcell{21.06}
& \gcell{20.99}
& \gcell{19.67}
& \gcell{19.95}
& \gcell{19.46} \\
& 32k \setbase{24.02}
& \cellcolor{heatneutral}\base
& \gcell{20.42}
& \gcell{23.57}
& \gcell{21.21}
& \textbf{\gcell{24.00}}
& \gcell{22.66}
& \gcell{20.71}
& \gcell{20.33}
& \gcell{19.43}
& \gcell{20.25}
& \gcell{19.30} \\
& 64k \setbase{21.60}
& \cellcolor{heatneutral}\base
& \gcell{13.45}
& \textbf{\gcell{20.96}}
& \gcell{12.86}
& \gcell{13.78}
& \gcell{18.33}
& \gcell{19.18}
& \gcell{14.53}
& \gcell{12.86}
& \gcell{13.12}
& \gcell{13.11} \\
\midrule

\multirow{3}{*}{\ds{2}}
& 16k \setbase{19.04}
& \cellcolor{heatneutral}\base
& \gcell{16.34}
& \textbf{\gcell{19.73}}
& \gcell{16.90}
& \gcell{17.02}
& \gcell{18.13}
& \gcell{18.58}
& \gcell{17.27}
& \gcell{17.46}
& \gcell{18.50}
& \gcell{17.97} \\
& 32k \setbase{21.10}
& \cellcolor{heatneutral}\base
& \gcell{16.28}
& \gcell{19.72}
& \textbf{\gcell{19.84}}
& \gcell{18.69}
& \gcell{17.34}
& \gcell{19.62}
& \gcell{18.61}
& \gcell{18.91}
& \gcell{18.33}
& \gcell{17.46} \\
& 64k \setbase{17.68}
& \cellcolor{heatneutral}\base
& \gcell{14.87}
& \textbf{\gcell{17.32}}
& \gcell{16.54}
& \gcell{16.41}
& \gcell{15.44}
& \gcell{15.84}
& \gcell{16.02}
& \gcell{16.49}
& \gcell{16.77}
& \gcell{16.53} \\
\midrule

\multirow{3}{*}{\ds{3}}
& 16k \setbase{5.64}
& \cellcolor{heatneutral}\base
& \gcell{3.22}
& \textbf{\gcell{5.35}}
& \gcell{4.26}
& \gcell{3.69}
& \gcell{4.67}
& \gcell{4.19}
& \gcell{3.88}
& \gcell{4.14}
& \gcell{4.93}
& \gcell{4.60} \\
& 32k \setbase{5.33}
& \cellcolor{heatneutral}\base
& \gcell{2.97}
& \textbf{\gcell{4.76}}
& \gcell{3.67}
& \gcell{3.02}
& \gcell{4.74}
& \gcell{3.59}
& \gcell{3.46}
& \gcell{3.75}
& \gcell{4.33}
& \gcell{4.01} \\
& 64k \setbase{6.18}
& \cellcolor{heatneutral}\base
& \gcell{2.13}
& \textbf{\gcell{5.04}}
& \gcell{1.06}
& \gcell{1.90}
& \gcell{4.66}
& \gcell{4.12}
& \gcell{2.34}
& \gcell{2.79}
& \gcell{3.50}
& \gcell{3.23} \\
\midrule

\multirow{3}{*}{\ds{4}}
& 16k \setbase{22.47}
& \cellcolor{heatneutral}\base
& \gcell{20.91}
& \textbf{\gcell{21.12}}
& \gcell{18.84}
& \gcell{20.92}
& \gcell{17.75}
& \gcell{20.32}
& \gcell{20.03}
& \gcell{18.96}
& \gcell{20.34}
& \gcell{20.90} \\
& 32k \setbase{19.66}
& \cellcolor{heatneutral}\base
& \gcell{17.84}
& \textbf{\gcell{19.74}}
& \gcell{19.43}
& \gcell{18.12}
& \gcell{18.51}
& \gcell{18.47}
& \gcell{16.88}
& \gcell{14.94}
& \gcell{17.31}
& \gcell{17.89} \\
& 64k \setbase{18.33}
& \cellcolor{heatneutral}\base
& \gcell{12.22}
& \gcell{17.20}
& \gcell{12.81}
& \gcell{14.89}
& \gcell{15.19}
& \textbf{\gcell{18.79}}
& \gcell{12.09}
& \gcell{11.14}
& \gcell{12.36}
& \gcell{12.98} \\
\midrule

\multirow{3}{*}{\ds{5}}
& 16k \setbase{13.88}
& \cellcolor{heatneutral}\base
& \gcell{11.97}
& \textbf{\gcell{13.38}}
& \gcell{13.10}
& \gcell{12.24}
& \gcell{12.75}
& \gcell{13.10}
& \gcell{12.67}
& \gcell{12.75}
& \gcell{12.90}
& \gcell{12.44} \\
& 32k \setbase{13.56}
& \cellcolor{heatneutral}\base
& \gcell{11.99}
& \textbf{\gcell{12.34}}
& \gcell{12.26}
& \gcell{11.11}
& \gcell{11.64}
& \gcell{12.00}
& \gcell{12.03}
& \gcell{12.10}
& \gcell{12.21}
& \gcell{12.19} \\
& 64k \setbase{13.95}
& \cellcolor{heatneutral}\base
& \gcell{9.84}
& \textbf{\gcell{12.41}}
& \gcell{8.16}
& \gcell{8.27}
& \gcell{10.28}
& \gcell{10.18}
& \gcell{10.12}
& \gcell{10.66}
& \gcell{10.94}
& \gcell{10.54} \\
\midrule

\multirow{3}{*}{\ds{6}}
& 16k \setbase{11.36}
& \cellcolor{heatneutral}\base
& \gcell{8.63}
& \textbf{\gcell{10.13}}
& \gcell{9.49}
& \gcell{9.13}
& \gcell{9.49}
& \gcell{9.18}
& \gcell{9.23}
& \gcell{9.01}
& \gcell{8.76}
& \gcell{8.64} \\
& 32k \setbase{10.05}
& \cellcolor{heatneutral}\base
& \gcell{7.01}
& \textbf{\gcell{8.96}}
& \gcell{7.96}
& \gcell{7.42}
& \gcell{8.82}
& \gcell{7.71}
& \gcell{8.23}
& \gcell{8.14}
& \gcell{7.88}
& \gcell{7.78} \\
& 64k \setbase{8.93}
& \cellcolor{heatneutral}\base
& \gcell{4.12}
& \textbf{\gcell{8.35}}
& \gcell{4.44}
& \gcell{4.88}
& \gcell{7.82}
& \gcell{7.99}
& \gcell{5.26}
& \gcell{5.01}
& \gcell{4.90}
& \gcell{4.63} \\
\midrule

\multirow{3}{*}{\ds{7}}
& 16k \setbase{21.41}
& \cellcolor{heatneutral}\base
& \gcell{18.54}
& \textbf{\gcell{21.51}}
& \gcell{21.50}
& \textbf{\gcell{21.51}}
& \gcell{20.76}
& \gcell{16.53}
& \gcell{20.21}
& \gcell{20.48}
& \gcell{21.03}
& \gcell{20.70} \\
& 32k \setbase{21.43}
& \cellcolor{heatneutral}\base
& \gcell{15.58}
& \textbf{\gcell{18.96}}
& \gcell{18.35}
& \gcell{18.24}
& \gcell{17.15}
& \gcell{14.98}
& \gcell{16.51}
& \gcell{16.22}
& \gcell{16.97}
& \gcell{16.74} \\
& 64k \setbase{20.32}
& \cellcolor{heatneutral}\base
& \gcell{8.55}
& \textbf{\gcell{17.86}}
& \gcell{8.44}
& \gcell{8.19}
& \gcell{17.17}
& \gcell{14.24}
& \gcell{9.23}
& \gcell{9.45}
& \gcell{10.11}
& \gcell{10.06} \\
\midrule

\multirow{3}{*}{\ds{8}}
& 16k \setbase{17.49}
& \cellcolor{heatneutral}\base
& \gcell{14.84}
& \gcell{15.91}
& \gcell{15.83}
& \gcell{14.94}
& \gcell{14.66}
& \gcell{14.97}
& \gcell{13.97}
& \gcell{14.79}
& \textbf{\gcell{16.93}}
& \gcell{14.84} \\
& 32k \setbase{15.96}
& \cellcolor{heatneutral}\base
& \gcell{11.21}
& \textbf{\gcell{13.04}}
& \gcell{11.76}
& \gcell{11.89}
& \gcell{11.74}
& \gcell{11.87}
& \gcell{10.33}
& \gcell{11.43}
& \gcell{12.61}
& \gcell{11.15} \\
& 64k \setbase{14.66}
& \cellcolor{heatneutral}\base
& \gcell{5.89}
& \textbf{\gcell{14.28}}
& \gcell{6.33}
& \gcell{7.15}
& \gcell{12.56}
& \gcell{12.28}
& \gcell{5.11}
& \gcell{6.23}
& \gcell{7.19}
& \gcell{6.47} \\
\midrule

\multirow{3}{*}{\ds{9}}
& 16k \setbase{16.47}
& \cellcolor{heatneutral}\base
& \gcell{13.90}
& \textbf{\gcell{16.52}}
& \gcell{14.62}
& \gcell{13.76}
& \gcell{12.14}
& \gcell{14.97}
& \gcell{15.59}
& \gcell{14.58}
& \gcell{16.31}
& \gcell{14.55} \\
& 32k \setbase{16.86}
& \cellcolor{heatneutral}\base
& \gcell{13.84}
& \textbf{\gcell{15.84}}
& \gcell{13.24}
& \gcell{14.65}
& \gcell{12.76}
& \gcell{13.44}
& \gcell{14.32}
& \gcell{15.49}
& \gcell{15.10}
& \gcell{14.94} \\
& 64k \setbase{13.85}
& \cellcolor{heatneutral}\base
& \gcell{10.41}
& \textbf{\gcell{14.04}}
& \gcell{10.11}
& \gcell{12.03}
& \gcell{11.65}
& \gcell{12.20}
& \gcell{11.44}
& \gcell{12.01}
& \gcell{13.46}
& \gcell{12.19} \\

\bottomrule
\end{tabular}
\end{adjustbox}
\end{table*}

\subsection{Allocation and Numerical Ablations}
\label{app:ablation_results}

\subsubsection{Allocation Structure and Sensitivity Localization}

We examine sensitivity variation and precision maps to explain the quality gains from the allocation controls defined in Appendix~\ref{app:comparison_configs}.
Table~\ref{tab:sensitivity_localization} reports mixed W4/W8 assignments in 59.81\% of the output-tile rows of Llama-3.2-3B and 87.37\% to 90.17\% of those in the other models.
Each such row contains input regions contributing to the same output-feature group but receiving different precisions.
The selected maps therefore use the within-channel allocation freedom throughout substantial portions of the models.
The sensitivity CV values also show that the extent of local variation differs across models, providing different distributions for the common allocation procedure.

The perturbation construction explains how these regions receive distinct scores.
Tiles in one output-tile row share output-gradient coordinates, while their input slices and W4/W8 candidate differences produce different local output changes in \eqref{eq:app_local_difference}.
The allocator can therefore distinguish their effects even though they contribute to the same output features.
Figure~\ref{fig:qproj} shows this local variation, and Figure~\ref{fig:distribution} shows how high-sensitivity regions are distributed across projections.
The precision maps connect those observations to the actual model representation: each selected region retains its tile coordinate and one stored format through the addressing scheme in Appendix~\ref{app:packing_details}.

\begin{table*}[t]
\centering
\scriptsize
\setlength{\tabcolsep}{4.0pt}
\renewcommand{\arraystretch}{0.80}
\caption{
Quantitative characterization of sensitivity localization and the resulting precision maps.
Sensitivity CV measures within-unit variation in sensitivity scores, state entropy summarizes W4/W8 assignment diversity, and mixed-row fraction reports the percentage of output-tile rows containing both precisions.
Nonzero variance is the share of units with varying scores.
}
\label{tab:sensitivity_localization}

\begin{adjustbox}{width=\textwidth}
\begin{tabular}{l c c c c r}
\toprule
\textbf{Model} &
\textbf{Sensitivity CV} &
\textbf{State Entropy} &
\textbf{Mixed-row} &
\textbf{Nonzero Variance} &
\textbf{W4 Tiles} \\
&
&
\textbf{(bits)} &
\textbf{(\%)} &
\textbf{(\%)} &
\\
\midrule

Llama-3.2-3B
& 0.47
& 0.36
& 59.81
& 100.00
& 412{,}880 \\

Qwen3-4B
& 2.10
& 0.52
& 87.37
& 100.00
& 532{,}288 \\

Ministral-3-8B
& 1.42
& 0.44
& 90.17
& 100.00
& 1{,}086{,}240 \\

Qwen3-14B
& 1.35
& 0.51
& 88.79
& 100.00
& 1{,}935{,}344 \\

\bottomrule
\end{tabular}
\end{adjustbox}
\end{table*}

\subsubsection{Tile-Size Ablation}

We vary tile size to test the resolution needed for precision allocation.
Using the unrounded measurements, $64\times64$ gives the lowest prefill and decode NLL in all six model-phase comparisons (Table~\ref{tab:tile_size_ablation}).
The finer $32\times32$ partition uses four times as many scales and precision labels, while $128\times128$ increases NLL in the tested settings.
The selected geometry therefore preserves useful local resolution with substantially less metadata than the finer partition.

\begin{table*}[t]
\centering
\scriptsize
\setlength{\tabcolsep}{6.0pt}
\renewcommand{\arraystretch}{0.80}
\caption{
Tile-size ablation across three models.
Prefill and decode quality are reported as held-out negative log-likelihood (NLL).
Bold and underlined values indicate the best and worst results, respectively, according to the unrounded measurements for each model.
}
\label{tab:tile_size_ablation}

\begin{adjustbox}{width=\textwidth}
\begin{tabular}{c cc cc cc}
\toprule
\multirow{2}{*}{\textbf{Tile Size}} &
\multicolumn{2}{c}{\textbf{Llama-3.2-3B}} &
\multicolumn{2}{c}{\textbf{Qwen3-4B}} &
\multicolumn{2}{c}{\textbf{Ministral-3-8B}} \\
\cmidrule(lr){2-3}
\cmidrule(lr){4-5}
\cmidrule(lr){6-7}
&
\textbf{Prefill $\downarrow$} &
\textbf{Decode $\downarrow$} &
\textbf{Prefill $\downarrow$} &
\textbf{Decode $\downarrow$} &
\textbf{Prefill $\downarrow$} &
\textbf{Decode $\downarrow$} \\
\midrule

$32\times32$
& 2.24
& 2.36
& \underline{2.38}
& \underline{2.50}
& 2.02
& 2.13 \\

$\mathbf{64\times64}$
& \textbf{2.24}
& \textbf{2.36}
& \textbf{2.37}
& \textbf{2.49}
& \textbf{2.01}
& \textbf{2.12} \\

$128\times128$
& \underline{2.26}
& \underline{2.39}
& 2.38
& 2.49
& \underline{2.02}
& \underline{2.14} \\

\bottomrule
\end{tabular}
\end{adjustbox}
\end{table*}

\subsubsection{Selected-Weight Storage and W4 Budget}

After choosing the tile geometry, we vary the W4 fraction to examine how the precision budget changes actual storage and runtime behavior.
All configurations in Table~\ref{tab:w4_budget_sweep} use the final \modelname runtime.
Table~\ref{tab:system_llama3b} reports the method-comparison reference, while the uniform endpoints here are configurations of the \modelname runtime.
The default row links the quality result in Table~\ref{tab:general_llama3b} (12.61 perplexity) to its measured storage and throughput.
Packed storage follows the integer-weight relation $8-4\rho$ closely, with reconstruction scales and indexing metadata included in the reported values.
Across the mixed configurations from 40\% to 80\% W4, packed storage decreases from 6.41 to 4.81 bits per weight and resident GPU memory decreases from 2.98 to 2.45\,GiB.
This progression agrees with the selected-representation accounting in \eqref{eq:app_packed_bytes}: each additional W4 tile replaces its stored W8 values with a payload of half the size.

Execution changes much less across these mixed configurations.
Throughput remains between 130.51 and 134.83 tokens/s, with the default 60\% configuration reaching 132.28 tokens/s at 5.61 packed bits per weight.
The representation therefore exposes a useful storage control while retaining a similar generation rate over the tested mixed budgets.
In the runtime described in Appendix~\ref{app:execution_details}, both formats supply INT8 operands, and each tile retains the scale of its selected candidate.
Changing the W4 fraction changes which numerical values are stored and how many bytes they occupy, while the arithmetic organization remains shared.

This distinction explains how the allocation and execution components fit together.
The budget and sensitivity ranking determine the selected numerical model.
Packing realizes that choice in bytes, and operand preparation presents those same selected values to the integer kernels.
The sweep shows how one execution design supports several compact storage levels.

\begin{table*}[t]
\centering
\scriptsize
\setlength{\tabcolsep}{3.0pt}
\renewcommand{\arraystretch}{0.80}
\caption{
Effect of the W4 allocation budget on storage and end-to-end inference for \textbf{Llama-3.2-3B-Instruct} on an NVIDIA A100 with a 128-token prompt and 512 generated tokens.
All configurations use the final \modelname runtime.
The 60\% W4 setting is the default operating point used in the main experiments.
Bold and underlined values indicate the best and worst results, respectively, across the evaluated W4 fractions.
}
\label{tab:w4_budget_sweep}

\begin{adjustbox}{width=\textwidth}
\begin{tabular}{c c c c c c c c}
\toprule
\textbf{W4 Fraction} &
\textbf{Packed BPW $\downarrow$} &
\textbf{TTFT $\downarrow$} &
\textbf{TPOT $\downarrow$} &
\textbf{Throughput $\uparrow$} &
\textbf{Resident $\downarrow$} &
\textbf{Peak $\downarrow$} &
\textbf{Energy $\downarrow$} \\
&
&
\textbf{(ms)} &
\textbf{(ms/token)} &
\textbf{(tok/s)} &
\textbf{(GiB)} &
\textbf{(GiB)} &
\textbf{(J/1K tokens)} \\
\midrule

0\%
& \underline{8.01}
& \textbf{18.39}
& \textbf{7.03}
& \textbf{141.72}
& \underline{3.48}
& \underline{3.48}
& 1177.30 \\

40\%
& 6.41
& 19.33
& 7.39
& 134.83
& 2.98
& 2.98
& \underline{1288.47} \\

50\%
& 6.01
& 19.36
& 7.48
& 133.26
& 2.84
& 2.84
& 1255.14 \\

\textbf{60\%}
& 5.61
& 19.67
& 7.54
& 132.28
& 2.72
& 2.72
& 1267.92 \\

70\%
& 5.21
& 19.80
& 7.58
& 131.47
& 2.58
& 2.58
& 1266.19 \\

80\%
& 4.81
& \underline{19.84}
& \underline{7.64}
& \underline{130.51}
& 2.45
& 2.45
& 1263.40 \\

100\%
& \textbf{4.01}
& 19.60
& 7.17
& 138.96
& \textbf{2.17}
& \textbf{2.17}
& \textbf{1145.29} \\

\bottomrule
\end{tabular}
\end{adjustbox}
\end{table*}

\subsubsection{Calibration-Data Sensitivity}

\begin{table*}[t]
\centering
\scriptsize
\setlength{\tabcolsep}{3.5pt}
\renewcommand{\arraystretch}{0.80}
\caption{
Sensitivity of downstream quality to calibration-set size and sampling seed across three models.
Each calibration size uses three seeds.
General Avg is reported as the mean across seeds; SD is the sample standard deviation, and Min/Max give the observed range.
Bold and underlined General Avg values indicate the highest and lowest mean, respectively, for each model.
}
\label{tab:calibration_robustness}

\begin{adjustbox}{width=\textwidth}
\begin{tabular}{c cccc cccc cccc}
\toprule
\multirow{2}{*}{\textbf{Set Size}} &
\multicolumn{4}{c}{\textbf{Llama-3.2-3B}} &
\multicolumn{4}{c}{\textbf{Qwen3-4B}} &
\multicolumn{4}{c}{\textbf{Ministral-3-8B}} \\
\cmidrule(lr){2-5}
\cmidrule(lr){6-9}
\cmidrule(lr){10-13}
&
\textbf{General Avg $\uparrow$} &
\textbf{SD} &
\textbf{Min} &
\textbf{Max} &
\textbf{General Avg $\uparrow$} &
\textbf{SD} &
\textbf{Min} &
\textbf{Max} &
\textbf{General Avg $\uparrow$} &
\textbf{SD} &
\textbf{Min} &
\textbf{Max} \\
\midrule

8
& \textbf{65.27}
& 0.04
& 65.22
& 65.30
& \textbf{69.27}
& 0.04
& 69.24
& 69.31
& 75.38
& 0.06
& 75.33
& 75.44 \\

16
& \underline{64.97}
& 0.11
& 64.87
& 65.08
& \underline{69.02}
& 0.10
& 68.94
& 69.13
& \textbf{75.54}
& 0.06
& 75.47
& 75.57 \\

24
& 65.02
& 0.03
& 65.00
& 65.06
& 69.10
& 0.06
& 69.05
& 69.16
& 75.47
& 0.07
& 75.43
& 75.55 \\

48
& 64.98
& 0.04
& 64.95
& 65.02
& 69.12
& 0.07
& 69.06
& 69.19
& \underline{75.37}
& 0.03
& 75.34
& 75.40 \\

\bottomrule
\end{tabular}
\end{adjustbox}
\end{table*}

We vary calibration-set size and sampling seed to assess how the data used for allocation affects downstream quality.
Table~\ref{tab:calibration_robustness} covers 8, 16, 24, and 48 documents, with three seeds at each size.
Within this sweep, the reported task-quality scores remain in narrow ranges for each model.
The sample standard deviation across seeds ranges from 0.03 to 0.11 percentage points, and similar quality is retained across a sixfold range of document counts.
Thus, the selected maps continue to support comparable downstream performance under the tested changes in calibration data.

These measurements evaluate the sampling dependence of the procedure described in Appendix~\ref{app:calibration_algorithm}.
Each sampled record contributes prompt-processing and token-generation observations, which provide inputs and loss gradients for the same candidate comparisons.
Phase normalization expresses their accumulated scores as relative tile importance, and the common 4-bit target fixes the number of selected substitutions.
The experiment consequently varies the evidence guiding placement while preserving the representation and storage objective.
The resulting quality stability supports obtaining useful local precision decisions from compact calibration collections.

\subsection{Inference Efficiency}
\label{app:system_results}

\subsubsection{Workload Scaling}

\begin{figure*}[t]
    \centering

    \includegraphics[width=0.274\linewidth]{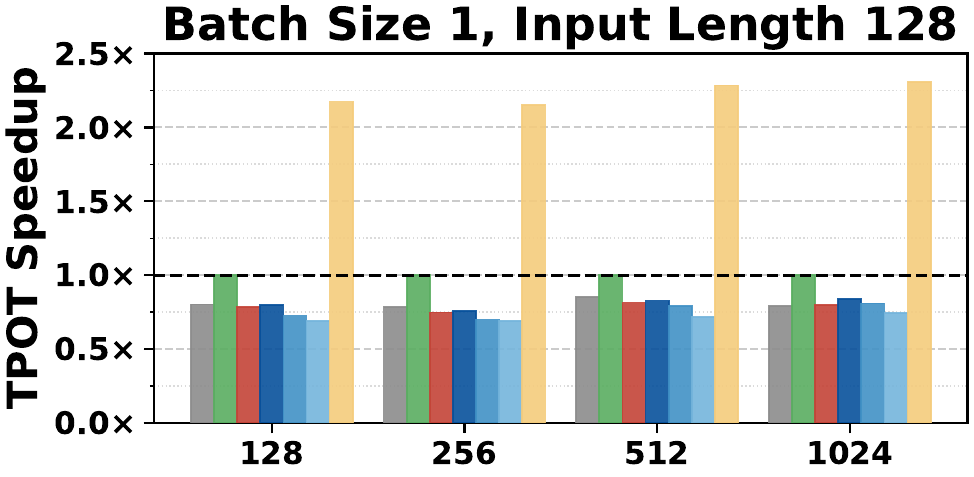}
    \hfill
    \includegraphics[width=0.234\linewidth]{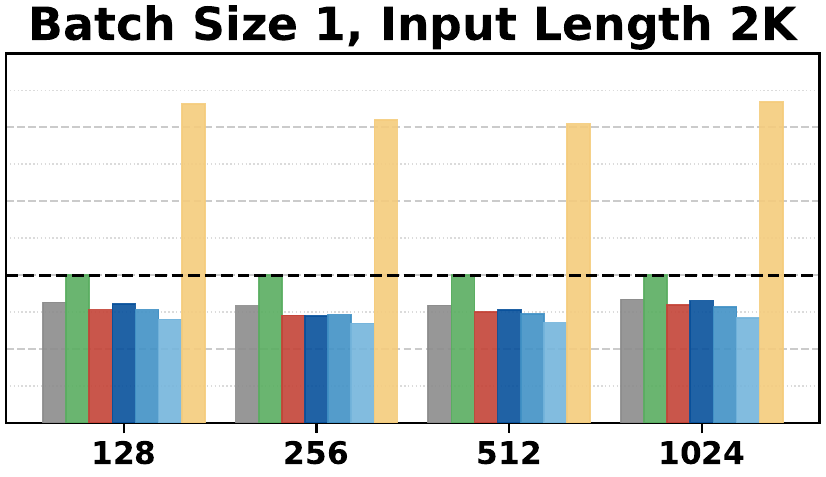}
    \hfill
    \includegraphics[width=0.234\linewidth]{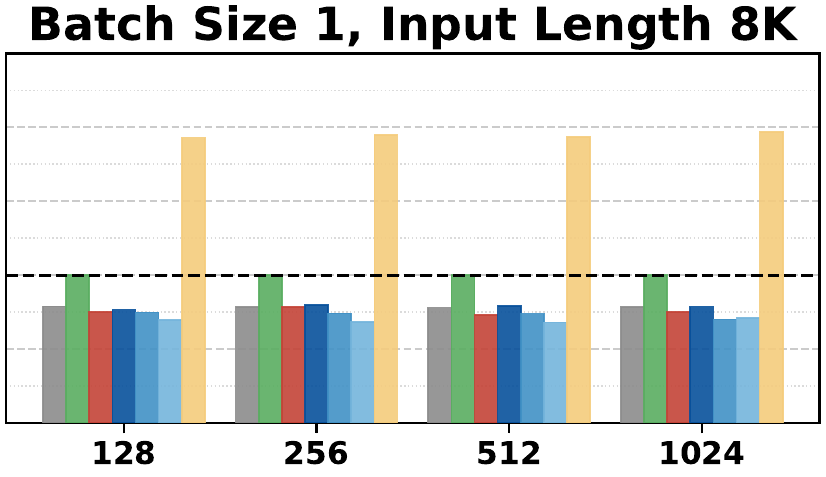}
    \hfill
    \includegraphics[width=0.234\linewidth]{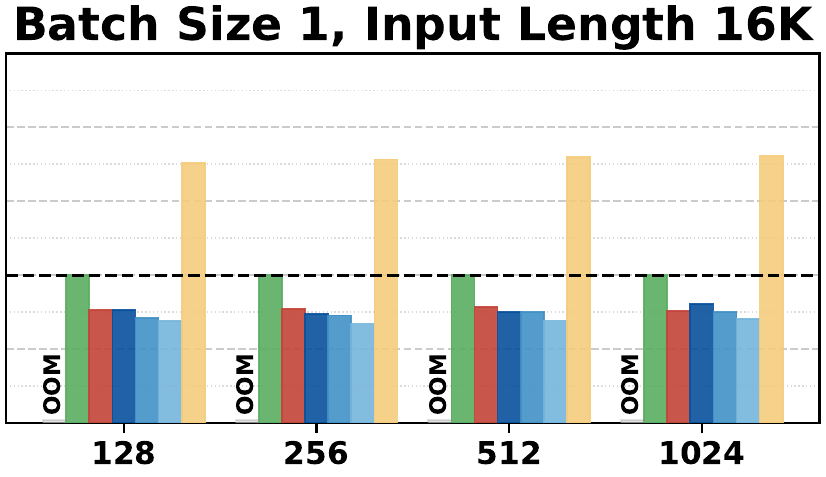}

    \includegraphics[width=0.274\linewidth]{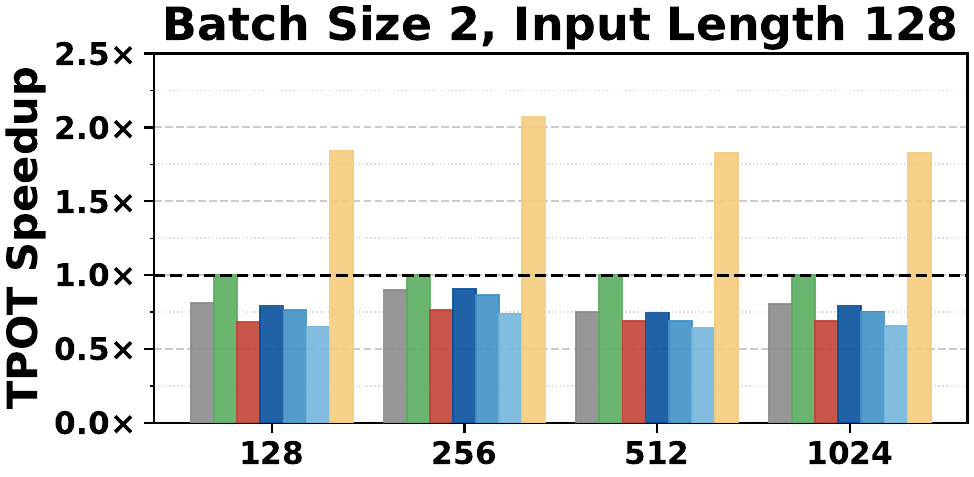}
    \hfill
    \includegraphics[width=0.234\linewidth]{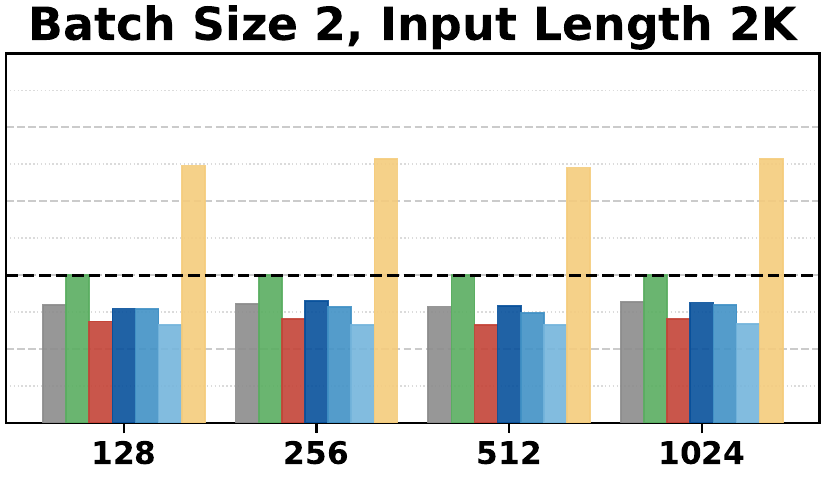}
    \hfill
    \includegraphics[width=0.234\linewidth]{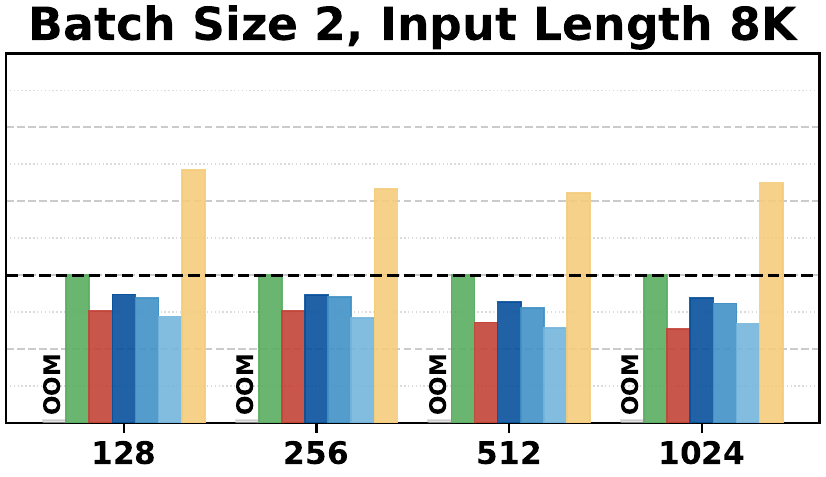}
    \hfill
    \includegraphics[width=0.234\linewidth]{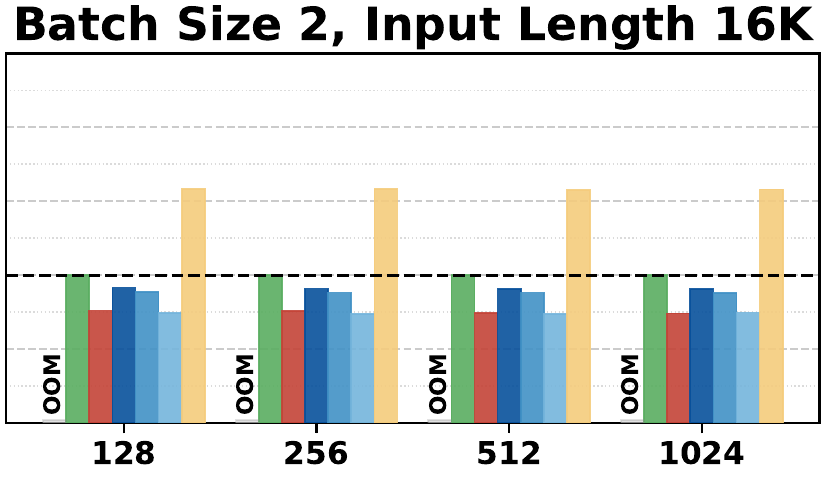}

    \includegraphics[width=0.274\linewidth]{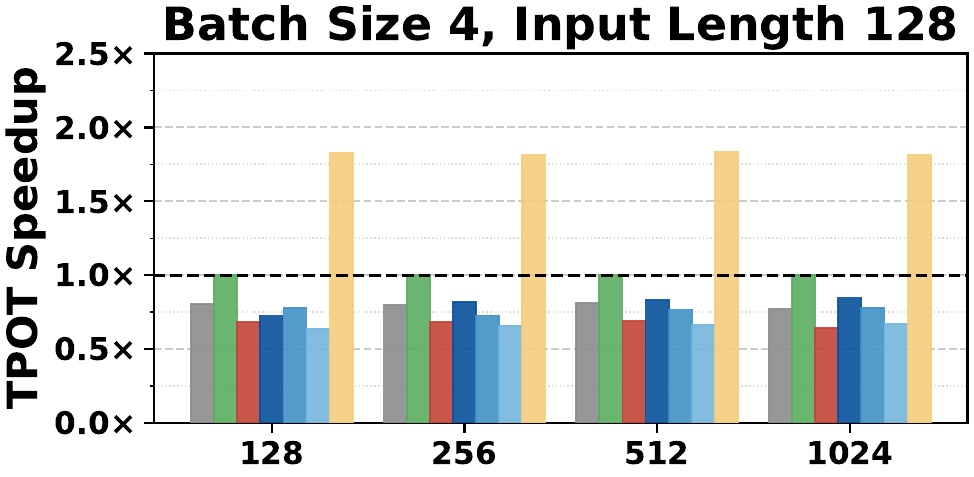}
    \hfill
    \includegraphics[width=0.234\linewidth]{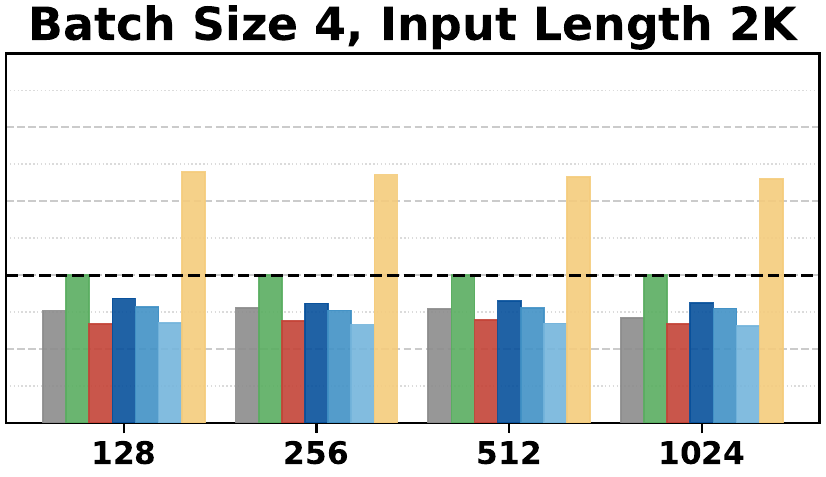}
    \hfill
    \includegraphics[width=0.234\linewidth]{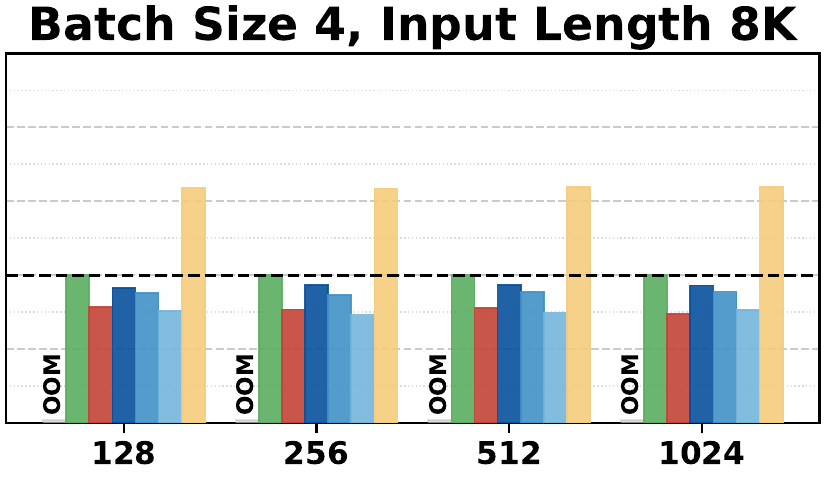}
    \hfill
    \includegraphics[width=0.234\linewidth]{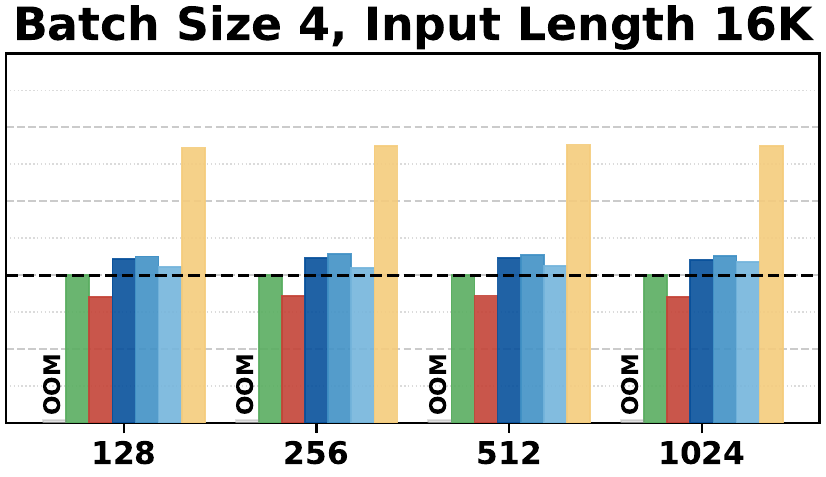}

    \includegraphics[width=0.274\linewidth]{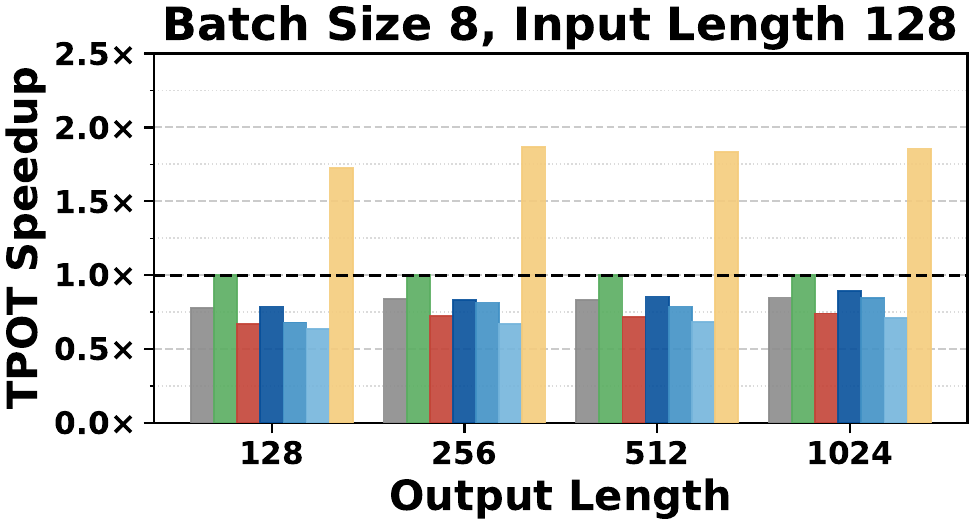}
    \hfill
    \includegraphics[width=0.234\linewidth]{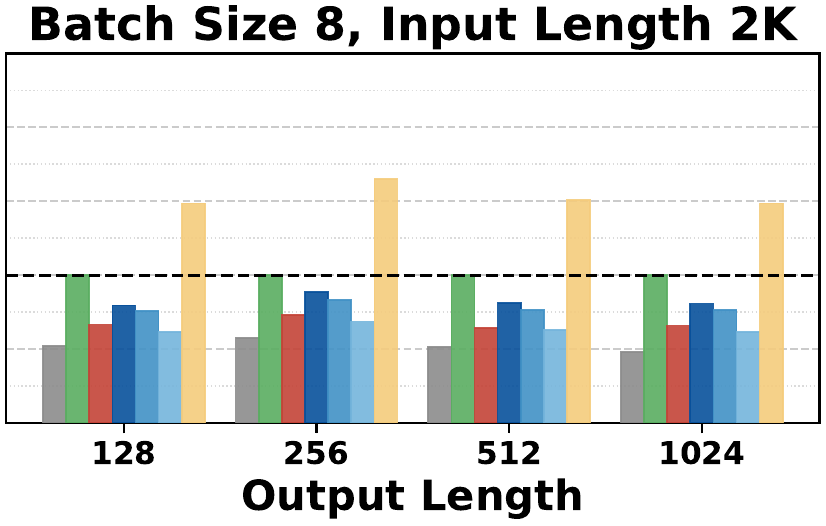}
    \hfill
    \includegraphics[width=0.234\linewidth]{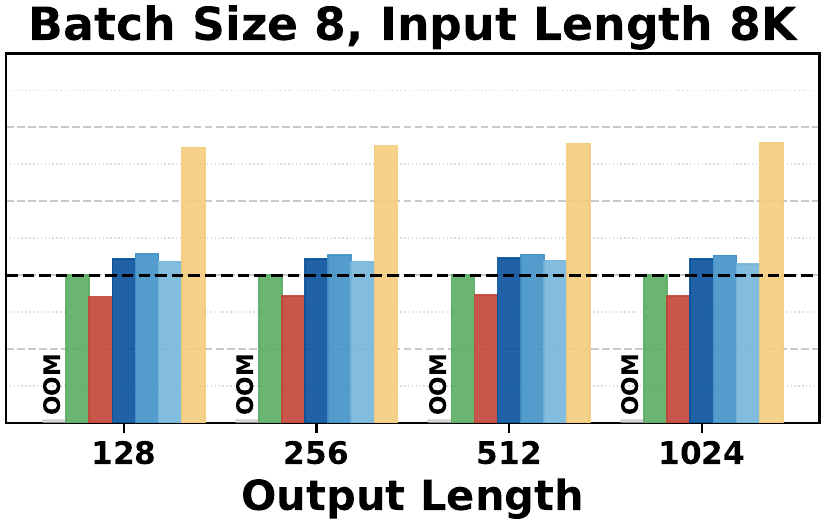}
    \hfill
    \includegraphics[width=0.234\linewidth]{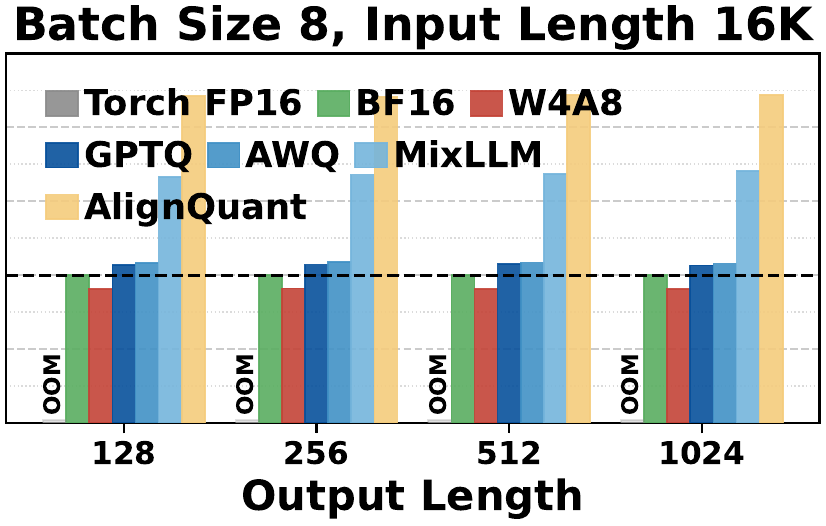}

    \caption{
    TPOT speedup relative to BF16 on Llama-3.2-3B.
    Rows vary batch size over 1, 2, 4, and 8; columns vary prompt length over 128, 2K, 8K, and 16K tokens.
    Each panel varies output length; higher values indicate faster generation.
    }
    \label{fig:tpot_grid}
\end{figure*}

\begin{figure*}[t]
    \centering

    \includegraphics[width=0.274\linewidth]{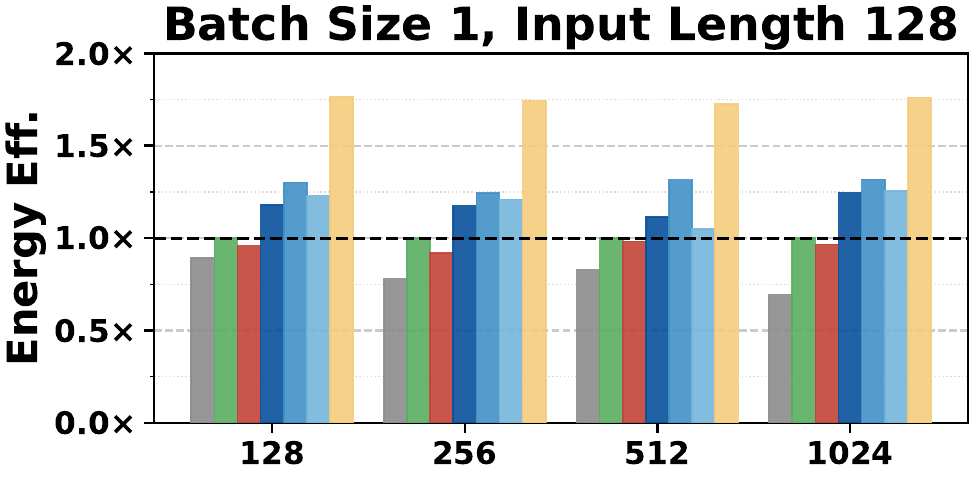}
    \hfill
    \includegraphics[width=0.234\linewidth]{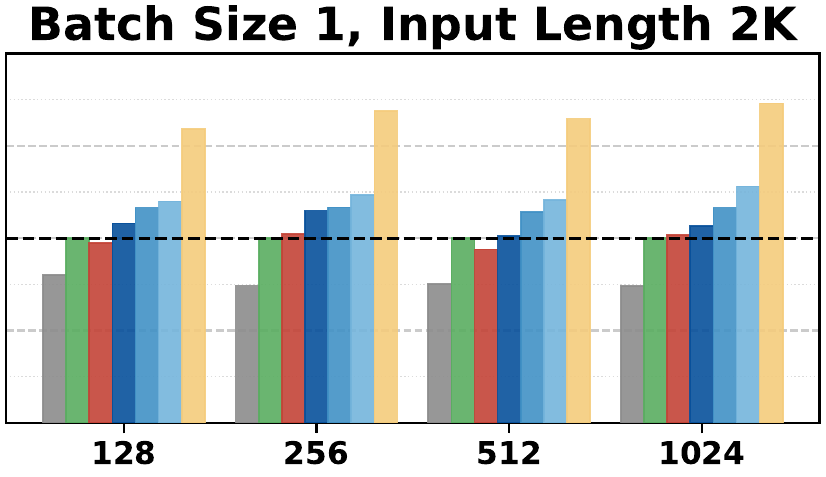}
    \hfill
    \includegraphics[width=0.234\linewidth]{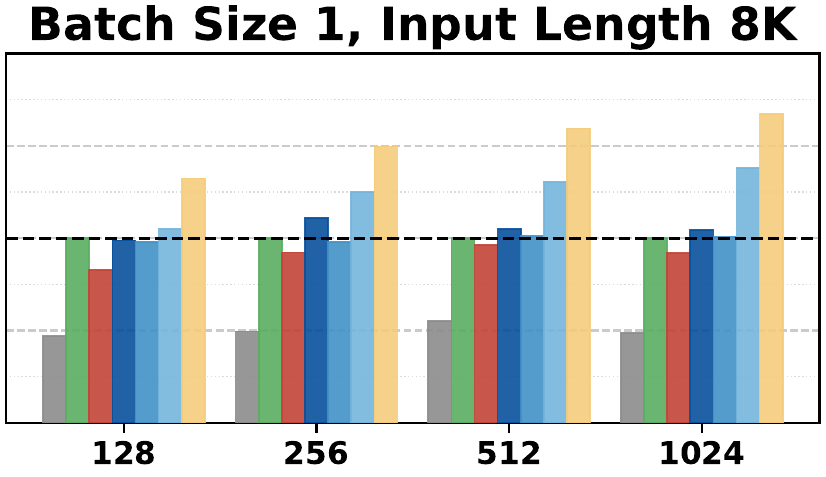}
    \hfill
    \includegraphics[width=0.234\linewidth]{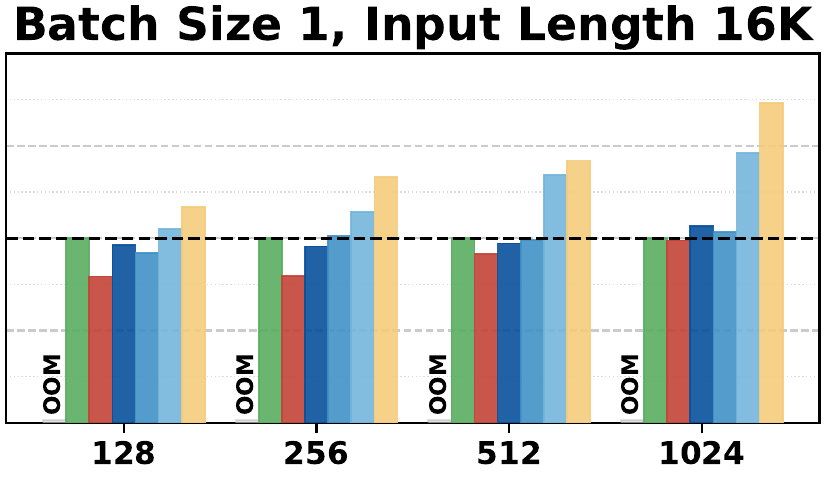}

    \includegraphics[width=0.274\linewidth]{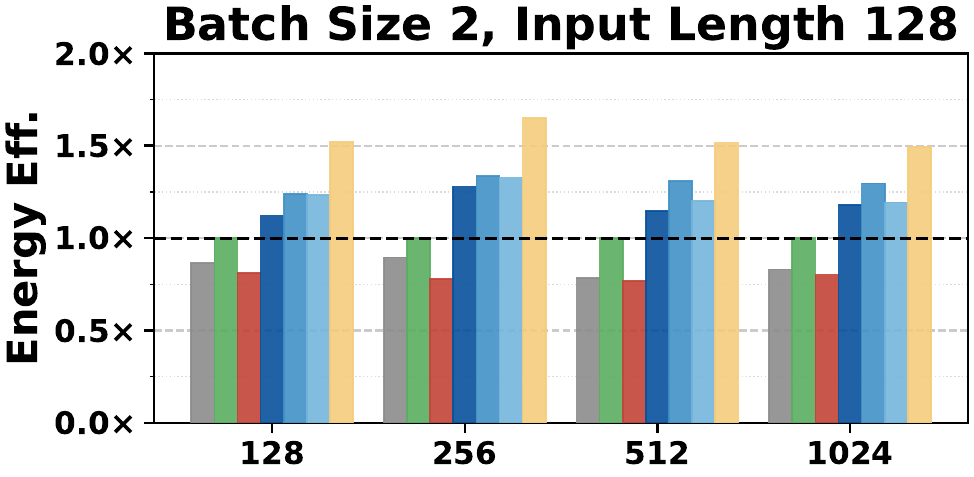}
    \hfill
    \includegraphics[width=0.234\linewidth]{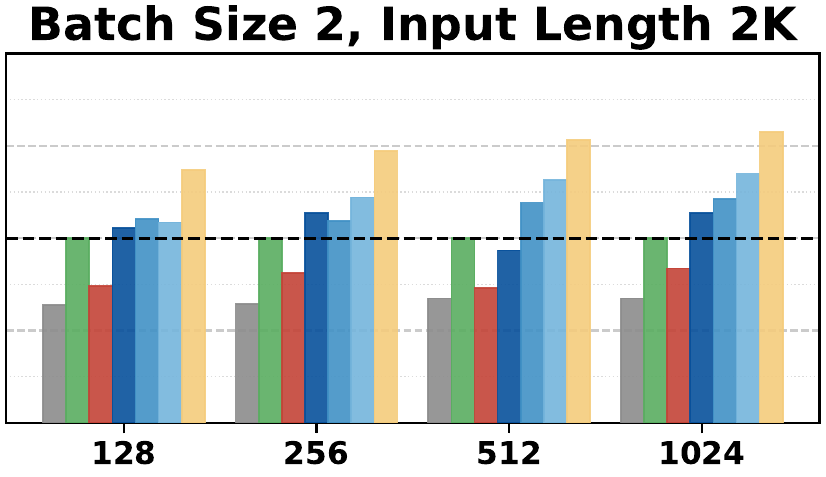}
    \hfill
    \includegraphics[width=0.234\linewidth]{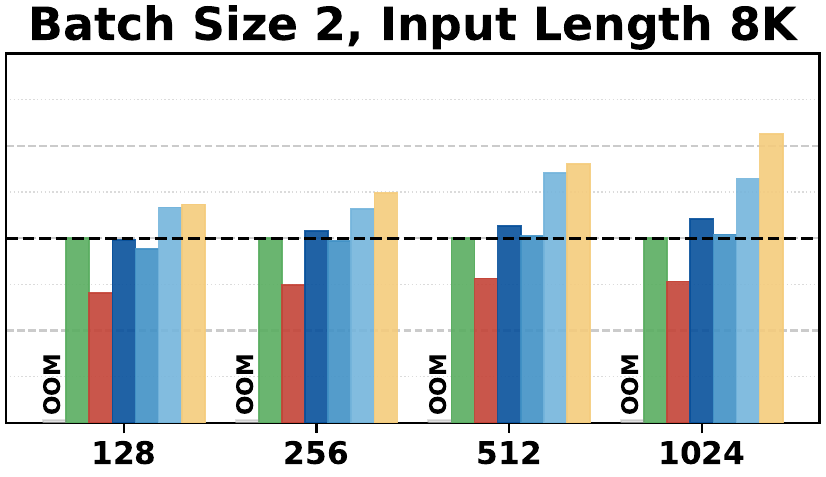}
    \hfill
    \includegraphics[width=0.234\linewidth]{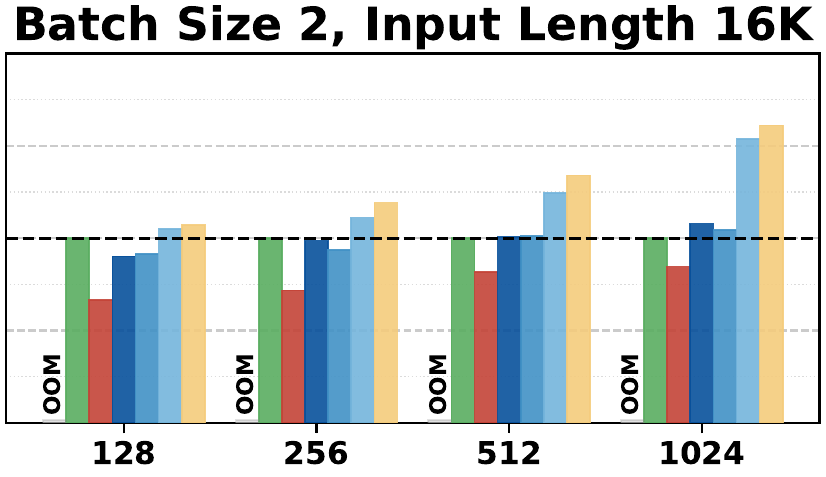}

    \includegraphics[width=0.274\linewidth]{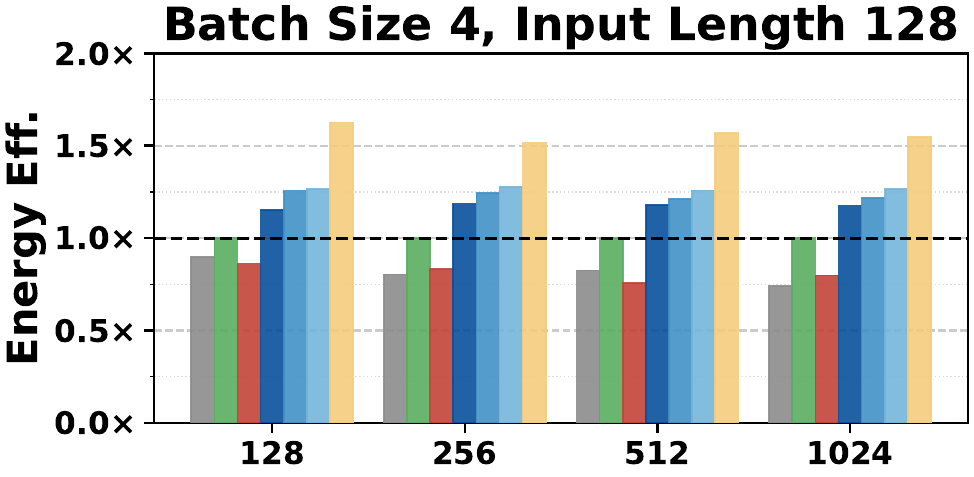}
    \hfill
    \includegraphics[width=0.234\linewidth]{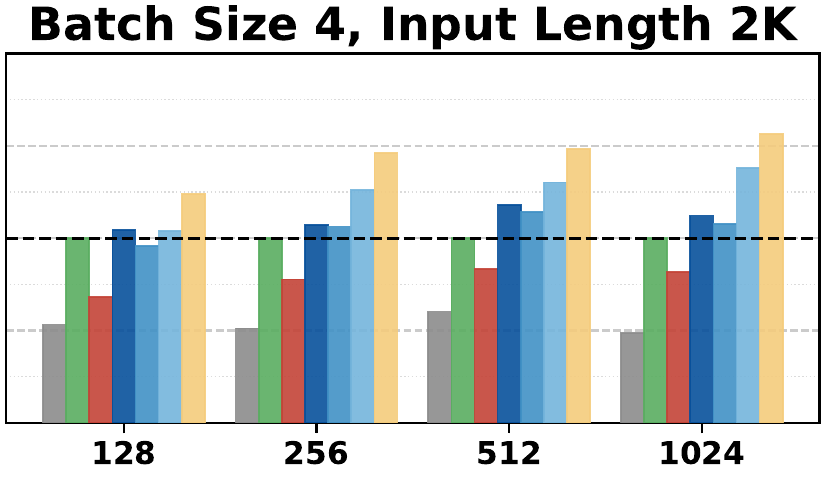}
    \hfill
    \includegraphics[width=0.234\linewidth]{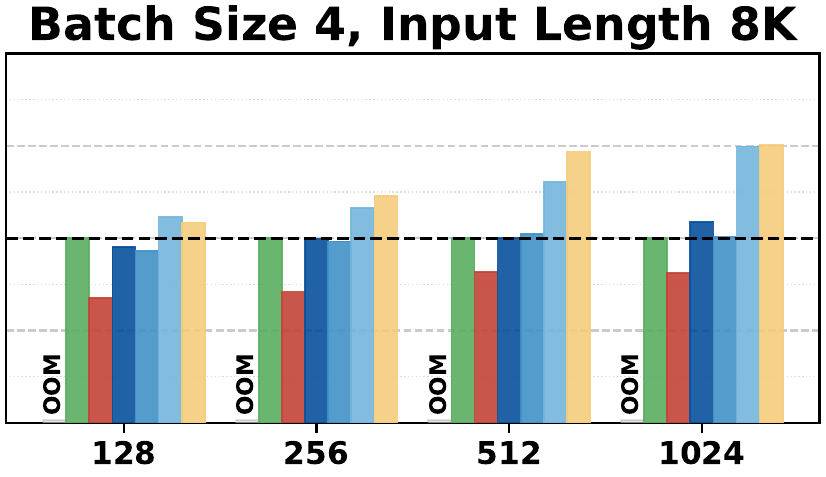}
    \hfill
    \includegraphics[width=0.234\linewidth]{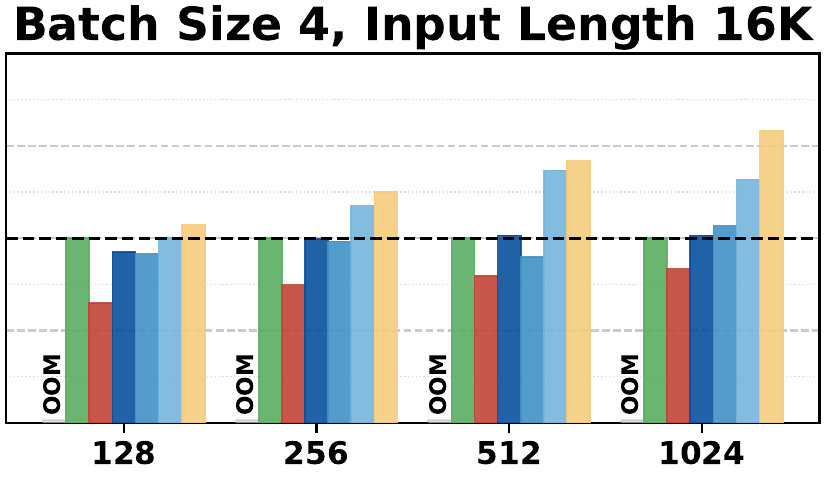}

    \includegraphics[width=0.274\linewidth]{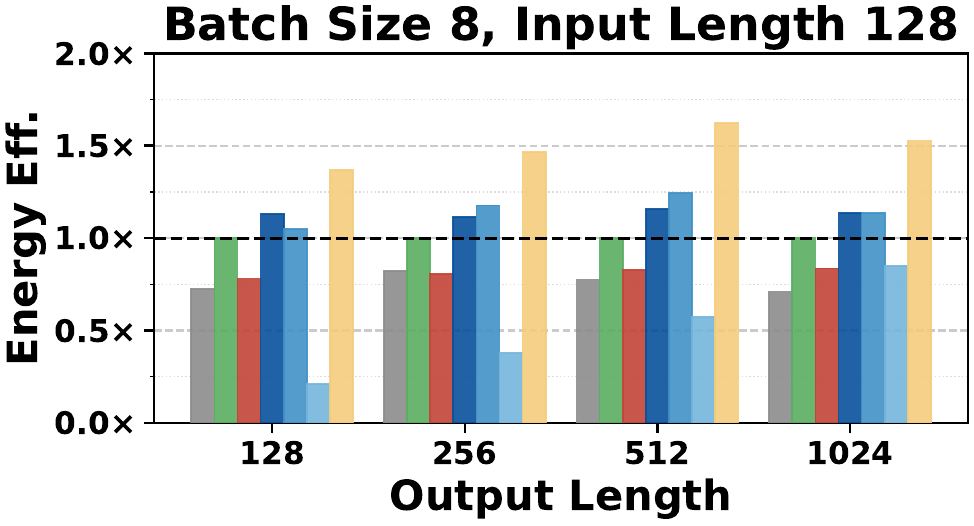}
    \hfill
    \includegraphics[width=0.234\linewidth]{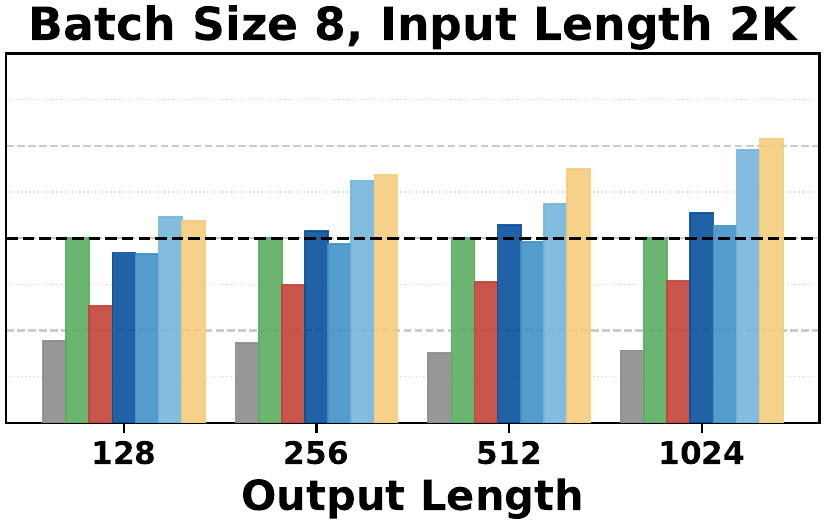}
    \hfill
    \includegraphics[width=0.234\linewidth]{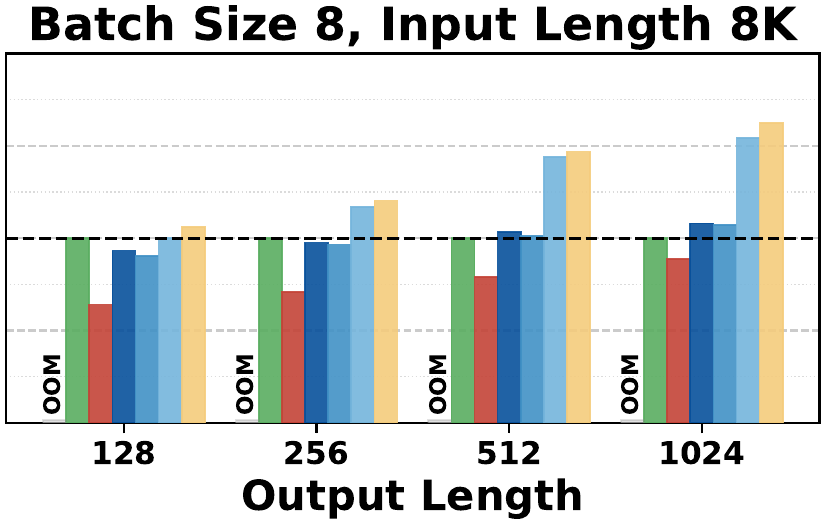}
    \hfill
    \includegraphics[width=0.234\linewidth]{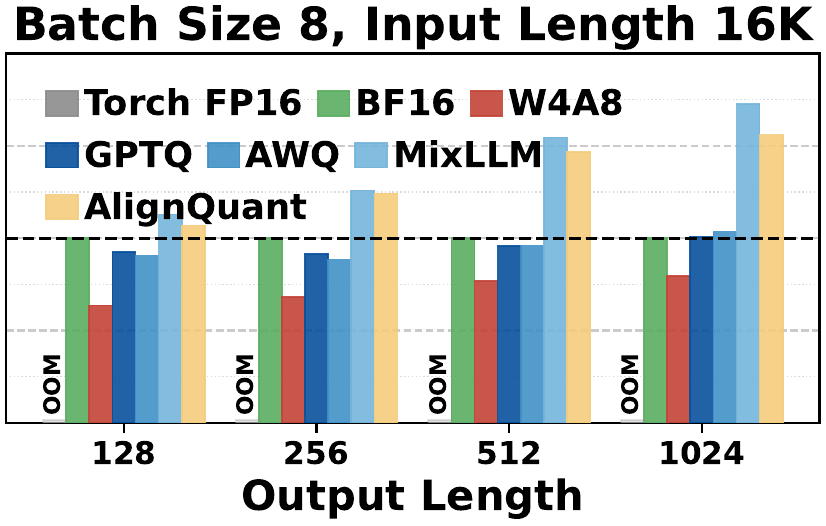}

    \caption{
    Generation energy efficiency relative to BF16 on Llama-3.2-3B.
    Rows vary batch size over 1, 2, 4, and 8; columns vary prompt length over 128, 2K, 8K, and 16K tokens.
    Each panel varies output length; higher values indicate greater energy efficiency.
    }
    \label{fig:energy_grid}
\end{figure*}

The throughput results in Figure~\ref{fig:generation_throughput_grid} show that the execution advantage extends across prompt lengths, output lengths, and batch sizes.
Figures~\ref{fig:tpot_grid} and~\ref{fig:energy_grid} examine the same workload grid from the complementary perspectives of per-token latency and energy efficiency.

\textbf{Per-token generation latency.}
Figure~\ref{fig:tpot_grid} reports TPOT speedup relative to BF16.
\modelname provides the largest speedup among the displayed configurations, exceeding $2\times$ for several batch-one workloads.
The advantage also extends to batched long-input generation, including the 16K-input settings at batch sizes 4 and 8.
These results support adapting integer execution to the available parallelism while reusing the same selected weight representation.

\textbf{Energy efficiency.}
Figure~\ref{fig:energy_grid} shows lower energy per generated token than BF16 across the displayed workloads.
For 8K and 16K prompts, the energy-efficiency gain generally increases with continuation length.
Longer continuations spread prompt processing and weight preparation over more generated tokens, while generation reuses the compact selected weights.
The results connect the shared representation and phase-specific execution to energy savings across request shapes.

Together with Figure~\ref{fig:generation_throughput_grid}, these measurements show that the same selected W4/W8 model improves three related aspects of generation: aggregate throughput, per-token latency, and energy efficiency.
The magnitude of each gain changes with workload composition, while the advantage persists across the tested request shapes.

\subsubsection{Cross-Hardware and Cross-Model Results}

\begin{table*}[t]
\centering
\scriptsize
\setlength{\tabcolsep}{3.0pt}
\renewcommand{\arraystretch}{0.75}
\caption{
End-to-end inference efficiency for \textbf{Llama-3.2-3B-Instruct} on an \textbf{NVIDIA RTX 4090}.
Cell colors are relative to BF16 under the corresponding metric direction: green for better than or equal to BF16, blue for worse than BF16, and gray for BF16.
Bold and underlined values indicate the best and worst results, respectively, among the compared methods excluding the reference columns.
}
\label{tab:system_llama3b_4090}

\begin{adjustbox}{width=\textwidth}
\begin{tabular}{l l c c c c c c c}
\toprule
\multirow{2}{*}{\textbf{Metric}} &
\multirow{2}{*}{\textbf{Unit}} &
\multicolumn{1}{c}{\textbf{Upper Ref.}} &
\multicolumn{1}{c}{\textbf{Lower Ref.}} &
\multicolumn{1}{c}{\textbf{Ours}} &
\multicolumn{4}{c}{\textbf{Others}} \\
\cmidrule(lr){3-3}
\cmidrule(lr){4-4}
\cmidrule(lr){5-5}
\cmidrule(lr){6-9}
& &
\textbf{BF16} &
\textbf{W4A8} &
\textbf{\modelname} &
\textbf{GPTQ} &
\textbf{AWQ} &
\textbf{MixLLM} &
\textbf{SFMP} \\
\midrule

TTFT $\downarrow$
& ms
\setbase{11.79}
& \cellcolor{heatneutral}\base
& \spcell{17.13}
& \spcell{16.54}
& \spcell{11.71}
& \spcell{10.97}
& \textbf{\spcell{8.35}}
& \underline{\spcell{57.55}} \\

TPOT $\downarrow$
& ms/token
\setbase{9.25}
& \cellcolor{heatneutral}\base
& \spcell{7.60}
& \textbf{\spcell{5.51}}
& \spcell{8.13}
& \spcell{8.68}
& \spcell{7.32}
& \underline{\spcell{53.12}} \\

Throughput $\uparrow$
& tok/s
\setbase{108.05}
& \cellcolor{heatneutral}\base
& \sgcell{131.31}
& \textbf{\sgcell{180.64}}
& \sgcell{122.88}
& \sgcell{115.16}
& \sgcell{136.54}
& \underline{\sgcell{18.82}} \\

Resident GPU Mem. $\downarrow$
& GiB
\setbase{5.99}
& \cellcolor{heatneutral}\base
& \spcell{2.08}
& \spcell{2.71}
& \textbf{\spcell{2.11}}
& \spcell{2.12}
& \underline{\spcell{2.87}}
& \spcell{2.16} \\

Peak VRAM $\downarrow$
& GiB
\setbase{6.09}
& \cellcolor{heatneutral}\base
& \spcell{2.16}
& \spcell{2.71}
& \textbf{\spcell{2.20}}
& \spcell{2.21}
& \underline{\spcell{2.92}}
& \spcell{2.37} \\

Energy $\downarrow$
& J/1K tokens
\setbase{2537.20}
& \cellcolor{heatneutral}\base
& \spcell{1536.05}
& \textbf{\spcell{1430.25}}
& \spcell{1520.65}
& \spcell{1549.37}
& \spcell{1690.31}
& \underline{\spcell{21103.45}} \\

\bottomrule
\end{tabular}
\end{adjustbox}
\end{table*}

\begin{table*}[t]
\centering
\scriptsize
\setlength{\tabcolsep}{3.0pt}
\renewcommand{\arraystretch}{0.75}
\caption{
End-to-end inference efficiency for \textbf{Llama-3.2-3B-Instruct} on an \textbf{NVIDIA H100 NVL}.
Cell colors are relative to BF16 under the corresponding metric direction: green for better than or equal to BF16, blue for worse than BF16, and gray for BF16.
Bold and underlined values indicate the best and worst results, respectively, among the compared methods excluding the reference columns.
}
\label{tab:system_llama3b_h100}

\begin{adjustbox}{width=\textwidth}
\begin{tabular}{l l c c c c c c c}
\toprule
\multirow{2}{*}{\textbf{Metric}} &
\multirow{2}{*}{\textbf{Unit}} &
\multicolumn{1}{c}{\textbf{Upper Ref.}} &
\multicolumn{1}{c}{\textbf{Lower Ref.}} &
\multicolumn{1}{c}{\textbf{Ours}} &
\multicolumn{4}{c}{\textbf{Others}} \\
\cmidrule(lr){3-3}
\cmidrule(lr){4-4}
\cmidrule(lr){5-5}
\cmidrule(lr){6-9}
& &
\textbf{BF16} &
\textbf{W4A8} &
\textbf{\modelname} &
\textbf{GPTQ} &
\textbf{AWQ} &
\textbf{MixLLM} &
\textbf{SFMP} \\
\midrule

TTFT $\downarrow$
& ms
\setbase{13.19}
& \cellcolor{heatneutral}\base
& \spcell{17.93}
& \spcell{15.86}
& \spcell{17.53}
& \spcell{18.11}
& \textbf{\spcell{15.79}}
& \underline{\spcell{59.92}} \\

TPOT $\downarrow$
& ms/token
\setbase{12.26}
& \cellcolor{heatneutral}\base
& \spcell{15.31}
& \textbf{\spcell{5.95}}
& \spcell{16.52}
& \spcell{16.91}
& \spcell{14.06}
& \underline{\spcell{56.84}} \\

Throughput $\uparrow$
& tok/s
\setbase{81.54}
& \cellcolor{heatneutral}\base
& \sgcell{65.29}
& \textbf{\sgcell{167.49}}
& \sgcell{60.52}
& \sgcell{59.13}
& \sgcell{71.12}
& \underline{\sgcell{17.59}} \\

Resident GPU Mem. $\downarrow$
& GiB
\setbase{6.02}
& \cellcolor{heatneutral}\base
& \spcell{2.10}
& \spcell{2.76}
& \textbf{\spcell{2.13}}
& \spcell{2.14}
& \underline{\spcell{2.89}}
& \spcell{2.18} \\

Peak VRAM $\downarrow$
& GiB
\setbase{6.11}
& \cellcolor{heatneutral}\base
& \spcell{2.19}
& \spcell{2.76}
& \textbf{\spcell{2.22}}
& \spcell{2.23}
& \underline{\spcell{2.94}}
& \spcell{2.40} \\

Energy $\downarrow$
& J/1K tokens
\setbase{1829.26}
& \cellcolor{heatneutral}\base
& \spcell{1866.95}
& \textbf{\spcell{1133.84}}
& \spcell{1899.70}
& \spcell{1945.35}
& \spcell{1866.01}
& \underline{\spcell{17611.37}} \\

\bottomrule
\end{tabular}
\end{adjustbox}
\end{table*}

\begin{table*}[t]
\centering
\scriptsize
\setlength{\tabcolsep}{3.0pt}
\renewcommand{\arraystretch}{0.75}
\caption{
Cross-model decode efficiency on an \textbf{NVIDIA A100 40GB}.
All measurements use batch size 1, a 128-token prompt, and 512 generated tokens.
TPOT is reported in ms/token, throughput in tokens/s, and energy in J/1K generated tokens.
Cell colors are relative to BF16 under the corresponding metric direction: green for better than BF16, blue for worse than BF16, and gray for BF16.
Bold and underlined values indicate the best and worst results, respectively, among the compared methods excluding the reference columns.
}
\label{tab:cross_model_decode_efficiency}

\begin{adjustbox}{width=\textwidth}
\begin{tabular}{l l l c c c c c c c}
\toprule
\multirow{2}{*}{\textbf{Model}} &
\multirow{2}{*}{\textbf{Metric}} &
\multirow{2}{*}{\textbf{Unit}} &
\multicolumn{1}{c}{\textbf{Upper Ref.}} &
\multicolumn{1}{c}{\textbf{Lower Ref.}} &
\multicolumn{1}{c}{\textbf{Ours}} &
\multicolumn{4}{c}{\textbf{Others}} \\
\cmidrule(lr){4-4}
\cmidrule(lr){5-5}
\cmidrule(lr){6-6}
\cmidrule(lr){7-10}
& & &
\textbf{BF16} &
\textbf{W4A8} &
\textbf{\modelname} &
\textbf{GPTQ} &
\textbf{AWQ} &
\textbf{MixLLM} &
\textbf{SFMP} \\
\midrule

\multirow{3}{*}{Llama-3.2-3B}
& TPOT $\downarrow$
& ms/token
\setbase{17.19}
& \cellcolor{heatneutral}\base
& \spcell{21.12}
& \textbf{\spcell{7.54}}
& \spcell{20.77}
& \spcell{21.78}
& \spcell{24.05}
& \underline{\spcell{82.33}} \\

& Throughput $\uparrow$
& tok/s
\setbase{58.18}
& \cellcolor{heatneutral}\base
& \sgcell{47.33}
& \textbf{\sgcell{132.28}}
& \sgcell{48.13}
& \sgcell{45.91}
& \sgcell{41.58}
& \underline{\sgcell{12.15}} \\

& Energy $\downarrow$
& J/1K tokens
\setbase{2280.68}
& \cellcolor{heatneutral}\base
& \spcell{2322.34}
& \textbf{\spcell{1267.92}}
& \spcell{2047.43}
& \spcell{1732.24}
& \spcell{2172.71}
& \underline{\spcell{22535.65}} \\
\midrule

\multirow{3}{*}{Qwen3-4B}
& TPOT $\downarrow$
& ms/token
\setbase{27.18}
& \cellcolor{heatneutral}\base
& \spcell{33.33}
& \textbf{\spcell{10.88}}
& \spcell{33.09}
& \spcell{34.47}
& \spcell{32.49}
& \underline{\spcell{109.06}} \\

& Throughput $\uparrow$
& tok/s
\setbase{36.78}
& \cellcolor{heatneutral}\base
& \sgcell{30.00}
& \textbf{\sgcell{91.62}}
& \sgcell{30.22}
& \sgcell{29.00}
& \sgcell{30.77}
& \underline{\sgcell{9.17}} \\

& Energy $\downarrow$
& J/1K tokens
\setbase{3540.00}
& \cellcolor{heatneutral}\base
& \spcell{3503.56}
& \textbf{\spcell{1735.21}}
& \spcell{2665.43}
& \spcell{2586.02}
& \spcell{2724.03}
& \underline{\spcell{29045.23}} \\
\midrule

\multirow{3}{*}{Ministral-3-8B}
& TPOT $\downarrow$
& ms/token
\setbase{22.47}
& \cellcolor{heatneutral}\base
& \spcell{28.14}
& \textbf{\spcell{13.15}}
& \spcell{28.27}
& \spcell{29.23}
& \spcell{33.76}
& \underline{\spcell{204.22}} \\

& Throughput $\uparrow$
& tok/s
\setbase{44.50}
& \cellcolor{heatneutral}\base
& \sgcell{35.50}
& \textbf{\sgcell{75.72}}
& \sgcell{35.38}
& \sgcell{34.21}
& \sgcell{29.62}
& \underline{\sgcell{4.90}} \\

& Energy $\downarrow$
& J/1K tokens
\setbase{4285.26}
& \cellcolor{heatneutral}\base
& \spcell{3589.44}
& \textbf{\spcell{2652.38}}
& \spcell{3248.43}
& \spcell{3132.54}
& \spcell{3219.88}
& \underline{\spcell{60742.08}} \\
\midrule

\multirow{3}{*}{Qwen3-14B}
& TPOT $\downarrow$
& ms/token
\setbase{30.29}
& \cellcolor{heatneutral}\base
& \spcell{36.24}
& \textbf{\spcell{19.85}}
& \spcell{36.45}
& \spcell{37.55}
& \spcell{35.17}
& \underline{\spcell{350.49}} \\

& Throughput $\uparrow$
& tok/s
\setbase{32.99}
& \cellcolor{heatneutral}\base
& \sgcell{27.55}
& \textbf{\sgcell{50.16}}
& \sgcell{27.43}
& \sgcell{26.63}
& \sgcell{28.42}
& \underline{\sgcell{2.85}} \\

& Energy $\downarrow$
& J/1K tokens
\setbase{6972.05}
& \cellcolor{heatneutral}\base
& \spcell{5364.27}
& \textbf{\spcell{4354.57}}
& \spcell{4747.57}
& \spcell{4848.51}
& \spcell{5805.63}
& \underline{\spcell{111668.07}} \\

\bottomrule
\end{tabular}
\end{adjustbox}
\end{table*}

\begin{table*}[!htb]
\centering
\scriptsize
\setlength{\tabcolsep}{3.0pt}
\renewcommand{\arraystretch}{0.75}
\caption{
Peak VRAM for \textbf{Llama-3.2-3B-Instruct} across prompt lengths.
Peak VRAM is reported in GiB.
Cell colors are relative to BF16: green for lower memory usage, blue for higher memory usage, and gray for BF16.
Bold and underlined values indicate the best and worst results, respectively, among the compared methods excluding the reference columns.
}
\label{tab:peak_vram_scaling_llama3b}

\begin{adjustbox}{width=\textwidth}
\begin{tabular}{l c c c c c c c}
\toprule
\multirow{2}{*}{\textbf{Length}} &
\multicolumn{1}{c}{\textbf{Upper Ref.}} &
\multicolumn{1}{c}{\textbf{Lower Ref.}} &
\multicolumn{1}{c}{\textbf{Ours}} &
\multicolumn{4}{c}{\textbf{Others}} \\
\cmidrule(lr){2-2}
\cmidrule(lr){3-3}
\cmidrule(lr){4-4}
\cmidrule(lr){5-8}
&
\textbf{BF16} &
\textbf{W4A8} &
\textbf{\modelname} &
\textbf{GPTQ} &
\textbf{AWQ} &
\textbf{MixLLM} &
\textbf{SFMP} \\
\midrule

128
\setbase{6.09}
& \cellcolor{heatneutral}\base
& \spcell{2.16}
& \spcell{2.72}
& \textbf{\spcell{2.20}}
& \spcell{2.21}
& \underline{\spcell{2.92}}
& \spcell{2.37} \\

512
\setbase{6.10}
& \cellcolor{heatneutral}\base
& \spcell{2.20}
& \spcell{2.80}
& \textbf{\spcell{2.22}}
& \spcell{2.23}
& \underline{\spcell{2.99}}
& \spcell{2.41} \\

2K
\setbase{6.35}
& \cellcolor{heatneutral}\base
& \spcell{2.54}
& \spcell{2.93}
& \textbf{\spcell{2.47}}
& \spcell{2.48}
& \underline{\spcell{3.23}}
& \spcell{2.62} \\

4K
\setbase{6.71}
& \cellcolor{heatneutral}\base
& \spcell{2.99}
& \spcell{3.13}
& \textbf{\spcell{2.83}}
& \spcell{2.84}
& \underline{\spcell{3.62}}
& \spcell{2.95} \\

8K
\setbase{7.43}
& \cellcolor{heatneutral}\base
& \spcell{3.90}
& \spcell{3.60}
& \textbf{\spcell{3.55}}
& \spcell{3.56}
& \underline{\spcell{4.40}}
& \spcell{3.69} \\

16K
\setbase{8.88}
& \cellcolor{heatneutral}\base
& \spcell{5.72}
& \textbf{\spcell{4.44}}
& \spcell{4.99}
& \spcell{5.00}
& \underline{\spcell{5.94}}
& \spcell{5.18} \\

\bottomrule
\end{tabular}
\end{adjustbox}
\end{table*}

The cross-model system experiment tests whether the quality-preserving allocations also deliver efficient generation as model dimensions and parameter counts change.
All four models use one A100 GPU with batch size 1, a 128-token prompt, and 512 generated tokens.
Table~\ref{tab:cross_model_decode_efficiency} shows lower TPOT and higher throughput for \modelname on every model, with TPOT speedups of $1.53\times$ to $2.50\times$ over BF16.
On Qwen3-14B, TPOT decreases from 30.29 to 19.85\,ms, and energy per 1,000 generated tokens decreases from 6972.05 to 4354.57\,J.
The largest model therefore combines the task-quality preservation reported earlier with lower time and energy costs for generation.

The common tile representation provides the connection across these model sizes.
Each projection stores its own selected values and scales, while the execution organization in Appendix~\ref{app:execution_details} distributes work over output tiles and input-column groups.
This lets the allocation follow each model's sensitivity distribution while reusing the same forms of integer computation.
The cross-model measurements show that this combination remains effective across the evaluated projection geometries.
Quality and efficiency are thus two properties of the same selected models.

To examine whether the execution design carries across GPUs, we repeat the Llama-3.2-3B workload on H100 NVL and RTX 4090.
\modelname achieves the highest throughput and lowest TPOT among the compared configurations on both devices (Tables~\ref{tab:system_llama3b_h100} and~\ref{tab:system_llama3b_4090}).
Throughput reaches 167.49 tokens/s on H100 NVL and 180.64 tokens/s on RTX 4090, compared with their respective BF16 results of 81.54 and 108.05 tokens/s.
Both measurements also report lower energy per generated token and a resident footprint of approximately 2.7\,GiB.

The representation and operand preparation have complementary roles in these results.
The selected W4 values provide compact storage, and their expansion preserves the integer values and reconstruction scales used by the common INT8 arithmetic.
The same numerical representation can therefore be used by the phase-specific kernels on each evaluated platform.
The measurements support carrying the local precision choices into efficient execution across these devices.

Persistent weight storage gives the model's base footprint, while longer requests add activations, key/value states, and temporary execution buffers.
We therefore measure peak GPU allocation across prompt lengths to assess the representation within a complete workload.
From 128 to 16K prompt tokens, \modelname's peak increases from 2.72 to 4.44\,GiB, while BF16 increases from 6.09 to 8.88\,GiB (Table~\ref{tab:peak_vram_scaling_llama3b}).
At 16K, \modelname uses half the BF16 peak memory and records the lowest peak among the compared configurations.
The compact model thus retains a substantial memory benefit as context-dependent state grows.

The distinction between persistent and temporary storage is specified in Appendices~\ref{app:storage_accounting} and~\ref{app:execution_details}.
For long prefill, a projection prepares selected weights in a temporary INT8 buffer for reuse across activation-row groups.
The packed model remains the persistent representation, and the prepared operands belong to the current projection call.
Peak-memory measurement includes these workload-dependent lifetimes alongside the model and context state.
The observed footprint therefore evaluates compact storage together with the execution strategy that uses it.

\subsection{Runtime-Control Results}
\label{app:runtime_results}

\begin{table*}[!htb]
\centering
\scriptsize
\setlength{\tabcolsep}{5.0pt}
\renewcommand{\arraystretch}{0.80}
\caption{
Runtime optimization ablation for \textbf{Llama-3.2-3B-Instruct} on an NVIDIA A100 40GB with a 128-token prompt and 512 generated tokens.
Bold and underlined values indicate the best and worst results, respectively, across the four runtime configurations.
}
\label{tab:runtime_optimization_ablation}

\begin{adjustbox}{width=\textwidth}
\begin{tabular}{l l c c c c}
\toprule
\multirow{2}{*}{\textbf{Metric}} &
\multirow{2}{*}{\textbf{Unit}} &
\multicolumn{1}{c}{\textbf{Full}} &
\multicolumn{3}{c}{\textbf{Runtime Controls}} \\
\cmidrule(lr){3-3}
\cmidrule(lr){4-6}
& &
\textbf{\modelname} &
\textbf{Fixed decode split} &
\textbf{Separate preparation} &
\textbf{Individual launches} \\
\midrule

TTFT $\downarrow$
& ms
& 19.67
& \textbf{19.62}
& \underline{29.80}
& 22.44 \\

TPOT $\downarrow$
& ms/token
& \textbf{7.54}
& 8.54
& 10.42
& \underline{21.07} \\

Throughput $\uparrow$
& tok/s
& \textbf{132.28}
& 116.79
& 95.66
& \underline{47.45} \\

Peak VRAM $\downarrow$
& GiB
& 2.72
& \textbf{2.71}
& \textbf{2.71}
& \underline{2.73} \\

Energy $\downarrow$
& J/1K tokens
& \textbf{1267.92}
& 1508.95
& 1820.18
& \underline{2234.22} \\

\bottomrule
\end{tabular}
\end{adjustbox}
\end{table*}

Having established the full-model gains, we keep the selected weights fixed and vary the execution configuration to examine where runtime improvements arise.
The CUDA Graph control measures a change in execution submission within \modelname, while the BF16 comparison evaluates the \textbf{complete deployed method}.
Each runtime effect is conditional on the other components active in its comparison.
Adaptive decode splitting lowers TPOT from 8.54 to 7.54\,ms, while TTFT remains approximately 19.6\,ms in both configurations (Table~\ref{tab:runtime_optimization_ablation}).
The change is concentrated in the generation schedule.
As described in Appendix~\ref{app:execution_details}, single-sequence decode has one activation row, so the kernel exposes additional parallel work through output features and input-column groups.
The split choice distributes reduction work, and measured TPOT supports adapting it to projection geometry.

Shared activation preparation has a broader phase effect.
Relative to separate preparation, it reduces TTFT from 29.80 to 19.67\,ms and TPOT from 10.42 to 7.54\,ms.
The implementation in Appendix~\ref{app:shared_activation} prepares a common transformed 8-bit input for compatible query/key/value or gate/up projections.
Each consumer then uses its own selected weights, scales, and output transform.
The reductions in both metrics match this placement of the optimization: input preparation is reused across projections during both prompt processing and generation.
The selected precision map stays attached to each projection while common input work is shared.

CUDA Graph execution addresses the submission of the repeated inference computation.
Compared with individual kernel launches, it lowers TPOT from 21.07 to 7.54\,ms and increases throughput from 47.45 to 132.28 tokens/s (Table~\ref{tab:runtime_optimization_ablation}).
Peak GPU memory is 2.73 and 2.72\,GiB, respectively, so the timing change occurs at a similar runtime footprint.
The comparison holds the selected numerical model fixed and changes how its computation is submitted.
It shows that efficient submission is an important part of converting fast low-bit projection work into full-model generation speed.

The controls identify improvements at three levels: reduction scheduling within a projection, input reuse across projections, and submission of repeated model computation.
These execution choices operate on the values selected by the same calibration and packing procedure.
Combined with the quality ablations, the results explain the roles of the design: local sensitivity guides where numerical precision is retained, compact packing realizes those choices in memory, and the runtime organizes their computation efficiently.
The shared tile partition connects these roles in one deployed model.

\end{document}